\documentclass{article}

\PassOptionsToPackage{numbers, compress}{natbib}
\PassOptionsToPackage{table}{xcolor}

\usepackage{makecell}
\usepackage{wrapfig}
\usepackage{amsthm}
\usepackage[main, final]{neurips_2026}

\usepackage[utf8]{inputenc} 
\usepackage[T1]{fontenc}    
\usepackage{hyperref}       
\usepackage{url}            
\usepackage{booktabs}       
\usepackage{amsfonts}       
\usepackage{nicefrac}       
\usepackage{microtype}      
\usepackage{xcolor}         
\usepackage{graphicx}
\usepackage{amsmath}
\usepackage{amssymb}
\usepackage{algorithm}
\usepackage{algorithmic}
\usepackage{xcolor}
\usepackage{multirow}
\usepackage{pifont} 
\title{SALT: When More Rollouts Don’t Help in Group-Based Policy Optimization and How to Make Them Matter}

\newcommand{\leaderflag}{%
  \begingroup\setlength{\unitlength}{0.65pt}%
  \begin{picture}(7,8)%
    \linethickness{0.35pt}%
    \put(1,0){\line(0,1){8}}%
    \put(1,4){\framebox(5,4){}}%
  \end{picture}%
  \endgroup
}

\newcommand{\authorname}[1]{\makebox[0.29\textwidth][c]{#1}}

\author{%
  \authorname{Powei Chang\thanks{%
    Equal contribution; \textsuperscript{\leaderflag} project leader;
    \textsuperscript{\ding{41}} corresponding author.
  }\kern0.55em\thanks{Work done during an internship at Bilibili Inc.}\kern0.45em} \\
  Bilibili Inc. \\
  Fudan University
  \And
  \authorname{Jinpeng Zhang\textsuperscript{*\,\leaderflag}} \\
  Bilibili Inc.
  \And
  \authorname{Chaoqun Sun\textsuperscript{$\diamond$}} \\
  Bilibili Inc. \\
  Fudan University
  \AND
  \authorname{MiniWell Tsao\textsuperscript{$\diamond$}} \\
  Bilibili Inc. \\
  Fudan University
  \And
  \authorname{Lianrui Li\textsuperscript{$\diamond$}} \\
  Bilibili Inc. \\
  Zhejiang University
  \And
  \authorname{Jianxiang Xiang} \\
  Bilibili Inc.
  \AND
  \authorname{Chenyu Wang} \\
  Bilibili Inc.
  \And
  \authorname{Yukang Gao} \\
  Bilibili Inc.
  \And
  \authorname{Dongying Kong\textsuperscript{\ding{41}}} \\
  Bilibili Inc.
}

\begin{document}

\makeatletter
\let\originalauthorfnsymbol\@fnsymbol
\renewcommand{\@fnsymbol}[1]{%
  \ifnum#1=2\relax\ensuremath{\diamond}\else\originalauthorfnsymbol{#1}\fi
}
\maketitle
\let\@fnsymbol\originalauthorfnsymbol
\makeatother

\begin{abstract}
Reinforcement learning with verifiable rewards (RLVR) often adopts GRPO-style group-relative updates, sampling multiple rollouts per prompt to construct normalized learning signals. However, merely increasing the number of rollouts does not reliably strengthen learning: under GRPO-style group normalization, per-rollout policy-gradient features can concentrate into a low-rank, signed geometry, causing substantial cancellation during aggregation and weakening the effective update. 
We address this failure mode with \textbf{\textit{SALT}}, a \underline{S}ubspace-\underline{A}daptive geometry p\underline{L}ug-in componen\underline{T} that uses sample-wise gradient geometry to reweight the coefficients of group-relative updates.
\textbf{\textit{SALT}} estimates a dominant spectral subspace from the mini-batch Gram geometry, decomposes group-relative coefficients into dominant and spectral-tail channels, and adaptively reweights the tail channel when signed cancellation is severe. Across diverse reasoning-oriented RLVR benchmarks and model scales, \textbf{\textit{SALT}} improves effective update geometry and performance without modifying the reward model or the rollout sampling procedure (\href{https://github.com/Chang-pw/SALT}{Code}).

\end{abstract}

\section{Introduction}

Reinforcement learning with verifiable rewards (RLVR) has become a central recipe for improving the reasoning ability of large language models (LLMs), as it optimizes model-generated solutions using automatically checkable outcome signals rather than subjective preference labels~\citep{grpo,ouyang2022training}.

Building on PPO-style policy optimization~\citep{ppo}, Group Relative Policy Optimization (GRPO) and its variants improve scalability by replacing the critic with group-relative normalization over multiple rollouts, and are now widely used in reasoning-oriented RL pipelines~\citep{deepseekai2025deepseekr1incentivizingreasoningcapability,yu2025dapoopensourcellmreinforcement}.
However, simply increasing the number of rollouts often yields sharply diminishing returns, suggesting a structural inefficiency in how group-relative updates aggregate learning signals~\citep{hou2024doesrlhfscaleexploring,xu2025rolloutsusefuldownsamplingrollouts}.

\newpage
\begin{wrapfigure}[16]{r}{0.46\textwidth}
    \vspace{-0.6em}
    \centering
    \includegraphics[width=0.98\linewidth]{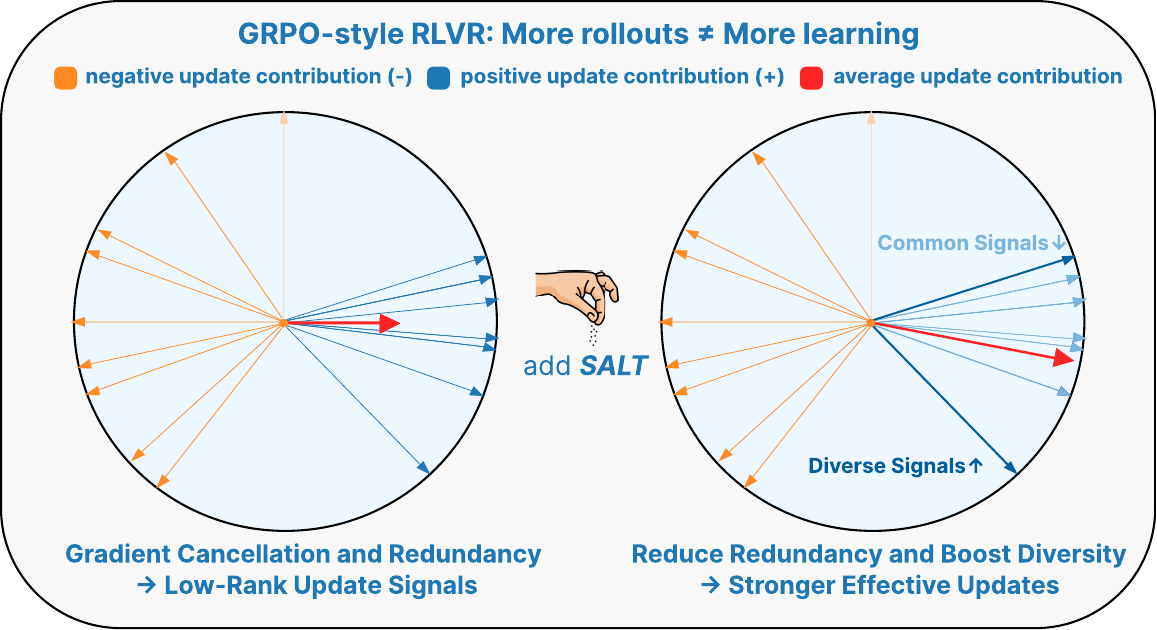}
    \vspace{-0.15em}
    \caption{Illustration of rollout inefficiency in GRPO-style RLVR. Per-rollout gradients exhibit signed, low-rank redundancy, so group-normalized aggregation cancels much of the learning signal. \textbf{\textit{SALT}} reweights dominant and spectral-tail channels to reduce signed cancellation and strengthen effective updates.}
    \label{fig:intro}
\end{wrapfigure}

Prior work attributes this bottleneck to insufficient exploration and limited response diversity within rollout groups~\citep{haarnoja2017reinforcementlearningdeepenergybased}.
Accordingly, existing remedies rely on policy-side signals, such as entropy-based, KL-based, or confidence-based regularization, to mitigate the loss of response diversity and stabilize group-based training~\citep{zhang2025edgegrpoentropydrivengrpoguided,dai2025cdecuriositydrivenexplorationefficient,wang2023reverseklgeneralizingdirect}. This leads to our pivotal research question:

\begin{quote}
Beyond maintaining response diversity, when do additional rollouts \emph{actually translate} into stronger and more informative \emph{group-relative updates}?
\end{quote}

In this work, we take an \textbf{update-space} view of rollout effectiveness.
Even when rollouts remain distinct at the sequence level, per-sample policy-gradient features within each prompt group can be highly redundant~\citep{chang2026spice}.
As illustrated in Figure~\ref{fig:intro}, these features concentrate in a low-rank, signed geometry and, after group-relative advantage weighting, form opposing update directions that cancel during aggregation.
We characterize this geometry by low effective dimensionality ($PR$) for spectral concentration and low effective sample size ($n_{\mathrm{eff}}$)~\citep{Kish1965SurveySampling} for weak net update strength after signed aggregation.
Together, these diagnostics explain why additional rollouts add compute without proportional learning gains.

This saturation arises from the structure of group-relative optimization. GRPO-style objectives center advantages within each prompt group~\citep{grpo}, canceling gradient components shared across rollouts; when per-sample gradient features concentrate into a low-rank, signed geometry, the aggregated update is left to depend on a small \emph{residual} component and can therefore fail to scale with additional rollouts. Thus, rollout inefficiency is not merely a sampling issue, but a failure to aggregate residual, non-redundant signals into effective learning progress.

Motivated by this diagnosis, we propose \textbf{\textit{SALT}}, a \underline{S}ubspace-\underline{A}daptive geometry p\underline{L}ug-in componen\underline{T} for GRPO-style training that uses mini-batch gradient geometry to reweight group-relative coefficients and reduce signed aggregation cancellation. \textit{SALT} estimates a dominant spectral subspace from mini-batch Gram geometry via a lightweight proxy~\citep{Hazelden2025FastNTK}, decomposes group-relative coefficients into dominant and spectral-tail channels, and adaptively reweights the tail channel with a cancellation-aware mixing coefficient. This spectral decomposition is a separately defined geometry-aware surrogate, rather than an identification with the within-prompt mean--residual decomposition above.
By sample-wise geometry-based reweighting, \textit{SALT} improves effective update geometry---spectral spread and signed aggregation strength---rather than relying only on policy-side exploration heuristics. We evaluate \textit{SALT} across reasoning RLVR benchmarks, two model scales, and multiple GRPO-style recipes, assessing both task performance and update-geometry diagnostics. Our contributions are:

\noindent\textbullet \ (\textbf{Phenomenon}) 
We diagnose signed low-rank redundancy in group-relative RLVR: within- and across-group policy-gradient features concentrate in a low-dimensional geometry.


\noindent\textbullet \ (\textbf{Method}) 
We propose \textbf{\textit{SALT}}, a subspace-adaptive plug-in that estimates a dominant spectral subspace and reweights group-relative coefficients through dominant and spectral-tail channels.

\noindent\textbullet \ (\textbf{Effectiveness}) 
We validate \textit{\textbf{SALT}} with extensive experiments and ablations, demonstrating consistent gains and improved effective update directions.

\section{Gradient Redundancy}
\label{section:2}

\subsection{Preliminaries}
\label{sec:pre}
\paragraph{Notation.}
We view an auto-regressive language model with parameters $\theta$ as a policy $\pi_\theta$~\citep{sutton2018reinforcement}. Given a prompt $x$ and response $y=(y_1,\dots,y_{|y|})$,
\begin{equation}
\pi_\theta(y\mid x)=\prod_{t=1}^{|y|}\pi_\theta(y_t\mid x,y_{<t}),
\end{equation}
with reward $r\in[0,1]$ assigned to each prompt--response pair $(x,y)$.

\paragraph{Group Relative Policy Optimization (GRPO). }
We now focus on GRPO~\citep{grpo,deepseekai2025deepseekr1}, which can introduce aggregation biases both within and across groups in our study.
In each iteration, GRPO samples $B$ prompts from $P(Q)$; for each prompt $q$, it then samples a set of $G$ candidate responses
$\{o_1,o_2,\ldots,o_G\}$ from the current policy $\pi_{\theta_{\mathrm{old}}}$. Each response $o_i$ is assigned a scalar reward $r_i$. Rewards within the group are normalized to obtain a response-level advantage~\citep{ding2025multilayergrpoenhancingreasoning}: 
\begin{equation}
\label{eq:advantage}
    \hat{A}_{i}= \frac{r_i-\text{mean}(r_1,r_2,\cdots,r_G)}{\text{std}(r_1,r_2,\cdots,r_G)}.
\end{equation}
It then updates the policy $\pi_\theta$ using a PPO-style clipped objective~\citep{ppo}. Formally, the GRPO objective function is defined as:
\begin{equation}
\label{equation:grpo}
    \begin{aligned}
\mathcal{J}_{\mathrm{GRPO}}(\theta)
&=
\mathbb{E}_{q\sim P(Q),\,\{o_i\}_{i=1}^{G}\sim \pi_{\theta_{\mathrm{old}}}(O\mid q)}
\Bigg[
\frac{1}{G}\sum_{i=1}^{G}\frac{1}{|o_i|}\sum_{t=1}^{|o_i|}
\Bigg(
\min\Bigg(
\frac{\pi_{\theta}(o_{i,t}\mid q,o_{i,<t})}{\pi_{\theta_{\mathrm{old}}}(o_{i,t}\mid q,o_{i,<t})}\,\hat{A}_{i,t},
\\
&
\operatorname{clip}\Bigg(
\frac{\pi_{\theta}(o_{i,t}\mid q,o_{i,<t})}{\pi_{\theta_{\mathrm{old}}}(o_{i,t}\mid q,o_{i,<t})},
1-\epsilon,\;1+\epsilon
\Bigg)\hat{A}_{i,t}
\Bigg) - \beta\,\mathbb{D}_{\mathrm{KL}}\!\left[\pi_{\theta}\,\|\,\pi_{\mathrm{ref}}\right]
\Bigg)
\Bigg]. 
\end{aligned}\end{equation}

More broadly, the rollout-based group optimization paradigm has led to a series of GRPO-style RLVR methods (e.g., DAPO~\citep{yu2025dapoopensourcellmreinforcement}; more formulations in Appendix~\ref{appendix:formulations}). Nevertheless, they are prone to a shared phenomenon, \textbf{within-group and cross-group gradient redundancy}, which we  study and characterize in the following sections.


\subsection{Gradient Redundancy}
We motivate our method from a persistent empirical pattern in GRPO-style RLVR: per-sample gradient features concentrate in a low-dimensional geometry, producing far fewer independent update signals than the raw sample count suggests. We quantify this with two complementary diagnostics: a regularized participation ratio ($PR$) from the batch Gram spectrum for spectral spread, and an effective sample size proxy ($n_{\mathrm{eff}}$) from signed per-sample update contributions for net aggregation strength~\citep{pmlr-v84-yin18a,Kish1965SurveySampling}. We report both within-group and batch-level variants to separate prompt-local redundancy from redundancy persisting across prompts.
\begin{figure}[t]
    \centering
    \includegraphics[width=0.95\linewidth]{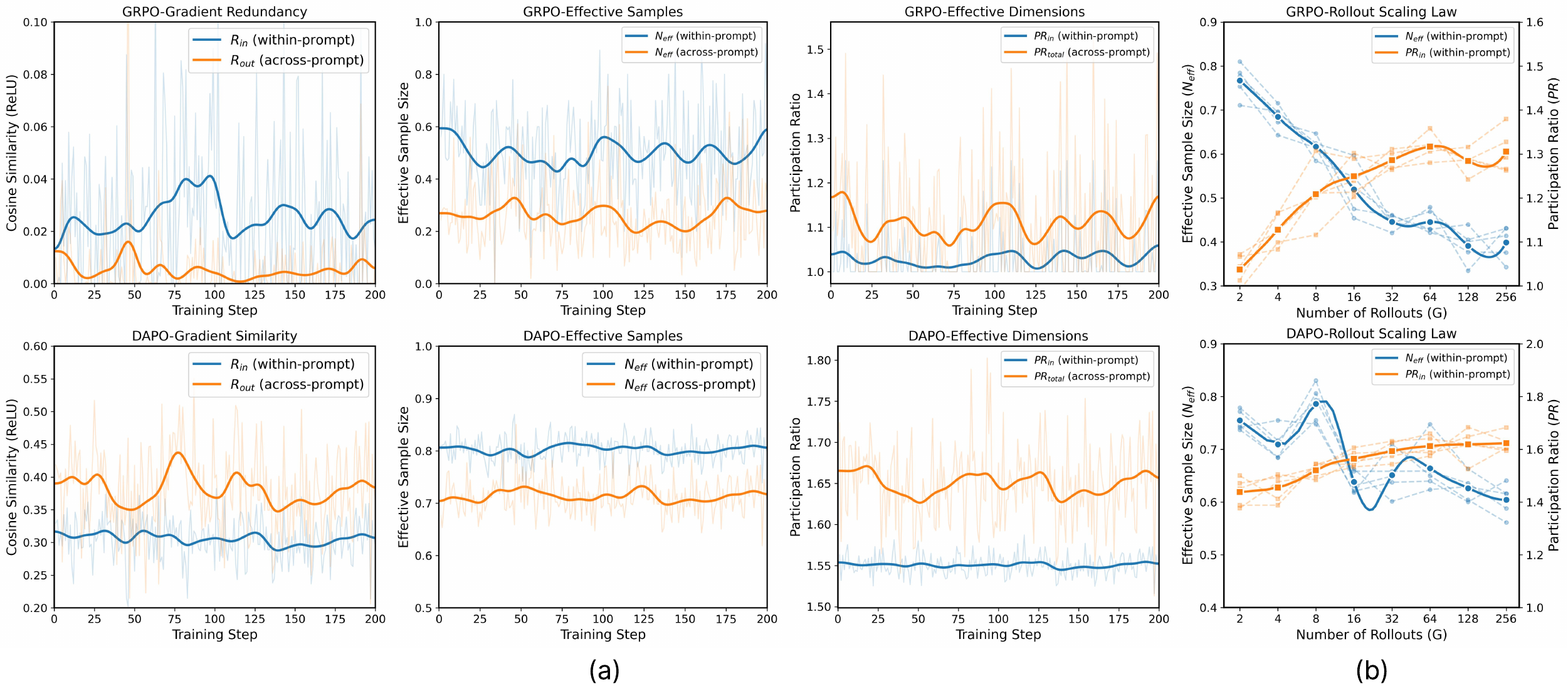}
    \vspace{-1em}
\caption{Signed low-rank gradient redundancy in GRPO-style RLVR.
(a) \textbf{Redundancy dynamics} for GRPO (top) and DAPO (bottom): throughout training, per-sample policy-gradient features concentrate in a low-dimensional subspace, causing strong cancellation and few effective independent update signals despite diverse rollouts.
(b) \textbf{Rollout scaling}: increasing group size makes $PR$ and $n_{\mathrm{eff}}$ saturate rather than scale proportionally, showing that more rollouts need not yield more learning.}
    \label{fig:fact}
\end{figure}

\subsubsection{Diagnostics}
\label{section:features}
\paragraph{Effective dimensionality.}
Let $\{g_i\}_{i=1}^m \subset \mathbb{R}^d$ denote gradient features for a collection of $m$ samples (responses).
We form the normalized Gram matrix
\begin{equation}
K \triangleq \frac{1}{m} G^\top G \in \mathbb{R}^{m\times m},\
G \triangleq [g_1,\ldots,g_m] \in \mathbb{R}^{d\times m}.
\label{eq:k_total}
\end{equation}
Since $K$ is positive semidefinite, it admits eigenvalues $\lambda_1,\ldots,\lambda_m \ge 0$.
We quantify the effective dimensionality of the set ${g_i}{i=1}^m$ via the \textbf{participation ratio} (PR),
defined as the ratio between the squared first spectral moment and the second spectral moment
\begin{equation}
\mathrm{PR}(K) \triangleq \frac{\left(\sum_{i=1}^m \lambda_i\right)^2}{\sum_{i=1}^m \lambda_i^2}.
\label{eq:pr_k}
\end{equation}
Larger $\mathrm{PR}(K)$ indicates a less concentrated spectrum and thus a higher effective dimensionality.
\paragraph{Effective sample size.}
Low $\mathrm{PR}$ indicates that gradient features concentrate in a low-dimensional subspace, i.e., stronger redundancy, but does not distinguish coherent agreement from cancellation. To quantify the \emph{net strength} of the aggregated update, we use an \textbf{effective sample size} proxy~\citep{Kish1965SurveySampling,Yang2006EffectiveSampleSize}.
\begin{equation}
n_{\mathrm{eff}}(\{u_i\}_{i=1}^m) \triangleq
\frac{\big\|\sum_{i=1}^m u_i\big\|_2^2}{\sum_{i=1}^m \|u_i\|_2^2},
\label{eq:neff} 
\end{equation}
where $u_i=\hat{A}_i  \nabla_{\theta}\,\log \pi_{\theta} $ denotes the signed per-sample update.
\paragraph{Alignment.}
To make the ``opposing orientation'' structure explicit, we also track a nonnegative alignment statistic \textbf{positive-part cosine alignment}~\citep{zhestiankin-ponomareva-2021-zhestyatsky} computed between per-sample directions
\begin{equation}
r(\{g_i\}_{i=1}^m)\triangleq \frac{2}{m(m-1)}\sum_{1\le i<j\le m} [\cos(g_i,g_j)]_+.
\end{equation}
where $[x]_+=\max(x,0)$. Higher values indicate stronger aligned modes, which can still cancel after signed aggregation when paired with opposing advantage signs.


\subsubsection{Empirical observations}

\paragraph{Signed low-rank gradient redundancy.}
Group-based RL aggregates rollouts within each prompt group before averaging across prompts. We therefore report both within- and cross-group diagnostics for GRPO and DAPO under the setup in Section~\ref{section:setup}, with implementation details in Appendix~\ref{app:metrics:redundancy}. As shown in Figure~\ref{fig:fact}(a), per-sample gradient features in GRPO-style training exhibit signed low-rank structure: they concentrate in a shared subspace, while signed update directions often oppose the group or batch mean. This reduces spectral spread and causes strong aggregation cancellation, yielding low $PR$ and $n_{\mathrm{eff}}$ even with many rollouts. Appendix~\ref{app:rq1} shows the same pattern on execution-verifiable code RLVR and additional model families.

\paragraph{Rollout scaling.}
Prior work shows that increasing rollout count gives diminishing returns~\citep{shao2024deepseekmathpushinglimitsmathematical,xu2025rolloutsusefuldownsamplingrollouts}. Figure~\ref{fig:fact} (b) revisits this effect in gradient space: as $G$ increases over $\{2,4,8,16,32,64,128,256\}$, neither the spectral-spread diagnostic nor $n_{\mathrm{eff}}$ grows proportionally. Under group-relative objectives, additional rollouts often add directionally redundant gradients, contributing little independent signal after group-centered aggregation. These observations motivate the next subsection, where a simple decomposition explains why group-centered updates are vulnerable to redundancy.

\subsection{Why redundancy hurts GRPO-style training}
\label{sec:theory_redundancy}
Although significant redundancy has been demonstrated, why does it still degrade RL training? For clarity, we analyze only the core policy-gradient term in Eq.~\eqref{equation:grpo}, omitting KL and clipping terms.

\paragraph{Group centering cancels shared gradients.}
For a prompt $q$, GRPO samples $G$ rollouts $\{o_{q,i}\}_{i=1}^G$. Let $\hat A_{q,i}$ be the group-normalized advantage and
$s_{q,i}\triangleq \nabla_\theta \log \pi_\theta(o_{q,i}\mid q)$ be the score gradient. The group-averaged policy-gradient contribution is
\begin{equation}
g_q(\theta)
\triangleq
\frac{1}{G}\sum_{i=1}^G \hat A_{q,i}s_{q,i},
\qquad
\sum_{i=1}^G \hat A_{q,i}=0,
\end{equation}
where the second relation follows from within-group advantage centering~\citep{shao2024deepseekmathpushinglimitsmathematical}. Thus, any perfectly shared component is canceled: if $s_{q,i}\equiv s_q$, then $g_q(\theta)=G^{-1}(\sum_i\hat A_{q,i})s_q=0$.

More generally, write each score gradient as the sum of a common component and a residual~\citep{chen2025grouppolicygradient}:
\begin{equation}
s_{q,i}=\bar s_q + \delta_{q,i},
\quad
\bar s_q \triangleq \frac{1}{G}\sum_{i=1}^G s_{q,i},
\quad
\sum_{i=1}^G \delta_{q,i}=0.
\end{equation}
Substituting this decomposition into the group update gives
\begin{equation}
\begin{aligned}
g_q(\theta)
&=
\frac{1}{G}\Big(\sum_{i=1}^G \hat A_{q,i}\Big)\bar s_q
+
\frac{1}{G}\sum_{i=1}^G \hat A_{q,i}\delta_{q,i}
=
\frac{1}{G}\sum_{i=1}^G \hat A_{q,i}\delta_{q,i}.
\end{aligned}
\end{equation}
Thus, group centering removes the common component and makes the effective update depend on the residual directions, whose magnitude is bounded by
\begin{equation}
\label{equation:bound}
\|g_q(\theta)\|_2
\le
\frac{1}{G}\left(\sum_{i=1}^G \hat A_{q,i}^2\right)^{1/2}
\left(\sum_{i=1}^G \|\delta_{q,i}\|_2^2\right)^{1/2}
\lesssim
\left(\frac{1}{G}\sum_{i=1}^G \|\delta_{q,i}\|_2^2\right)^{1/2},
\end{equation}
where the last step uses $\sum_i \hat A_{q,i}^2 \approx G$ under unit empirical variance within the group 
and the derivation is provided in Appendix~\ref{app:residual_bound}.  

Therefore, more rollouts help only when they increase the residual gradient energy. Redundant score-gradient directions are removed by group centering and do not strengthen the update. This explains the rollout-scaling failure mode: sequence-level diversity may still leave small residual directions that cancel under signed advantages, reducing $n_{\mathrm{eff}}$. Thus, effective rollout scaling requires non-redundant residual directions. This residual-update bottleneck motivates \textit{SALT}, but the method's mini-batch spectral decomposition introduced next is distinct from the within-prompt mean--residual decomposition analyzed here. We empirically validate the residual-bottleneck prediction in Appendix~\ref{app:residual_diagnostic}.


\section{Methodology}
\label{sec:method}
Motivated by the redundancy and signed cancellation identified above, \textit{SALT} does not aim to preserve the original GRPO estimator. Instead, it intentionally constructs a geometry-reweighted surrogate. The procedure is summarized in Algorithm~\ref{alg:salt_full} in Appendix~\ref{app_algo}.


\subsection{Dual-Channel Advantage Reweighting}
\label{sec:pr-neff-reweight}

Let $\mathcal{I}=\{(b,i): b=1,\dots,B,\ i=1,\dots,G\}$ index all responses in the current mini-batch,
and denote $m\triangleq|\mathcal{I}|=BG$. Stack the  group-normalized advantages
into
\begin{equation}
a \triangleq [\hat A_{b,i}]_{(b,i)\in\mathcal{I}} \in \mathbb{R}^{m}.
\end{equation}

\paragraph{Dominant subspace.}
Let the normalized batch Gram matrix $K_{\mathrm{total}}\in\mathbb{R}^{m\times m}$ be as defined in Eq.~\eqref{eq:k_total}.
Consider its eigen-decomposition
\begin{equation}
K_{\mathrm{total}} = V\Lambda V^\top,\quad
\Lambda=\mathrm{diag}(\lambda_1,\dots,\lambda_m),
\end{equation}
where $V=[v_1,\dots,v_m]$ is orthonormal. Let $PR_{\mathrm{total}}\triangleq PR(K_{\mathrm{total}})$
be the participation ratio defined in Eq.~\eqref{eq:pr_k} and~\eqref{eq:pr_total_2}. We set the dominant-subspace dimension by
\begin{equation}
k_t \triangleq \mathrm{clip}\!\left(\big\lfloor  PR_{\mathrm{total}}\big\rceil,\ 1, BG\right),
\end{equation}
and define $V_{k}\triangleq[v_1,\dots,v_{k_t}]\in\mathbb{R}^{m\times k_t}$ together with the orthogonal projectors~\citep{yu2020gradientsurgerymultitasklearning}
\begin{equation}
P_k \triangleq V_k V_k^\top,\qquad P_k^\perp \triangleq I - P_k.
\end{equation}
All quantities above are computed from mini-batch statistics~\citep{delgiudice2021effectivedimensionality,pmlr-v238-giaffar24a}.

\paragraph{Two-channel decomposition and reweighting.}
Motivated by the cancellation diagnosis in Sec.~2.3, we separately define a signed dominant spectral channel together with a one-sided spectral-tail channel:
\begin{equation}
a_{\mathrm{main}} \triangleq P_k a,\qquad
a_{\mathrm{exp}} \triangleq P_k^\perp\,[a]_+,
\end{equation}
where $[a]_+ \triangleq \max(a,0)$ is applied elementwise. This avoids symmetrically reweighting heterogeneous negative tail coefficients that may reintroduce cancellation, while $P_k a$ preserves group-relative feedback. These dominant/tail spectral channels are not identified with the within-prompt common/residual terms in Sec.~2.3. We validate this method choice in the ablation study in Section~\ref{sec:ablation}.

Let $\{u_{b,i}\}_{(b,i)\in\mathcal{I}}$ be the signed per-sample
update contributions, and let
$n_{\mathrm{eff}}(\{u_{b,i}\})$ be the effective sample size proxy already defined in Eq.~\eqref{eq:neff}. We set the adaptive exploration
mixing coefficient by
\begin{equation}
\alpha_t \triangleq \max \Big(0\;,\;1 - n_{\mathrm{eff}}\!\left(\{u_{b,i}\}_{(b,i)\in\mathcal{I}}\right)\Big).
\end{equation}
Finally, we form the reweighted advantages
\begin{equation}
\begin{aligned}
a' 
&\triangleq a_{\mathrm{main}} + \alpha_t a_{\mathrm{exp}} = P_k a + \alpha_t P_k^\perp\,[a]_+.
\end{aligned}
\label{eq:aprime}
\end{equation}
We keep the GRPO-style objective unchanged defined in Eq.~\eqref{equation:grpo} and replace the original coefficients $\hat A_{b,i}$
by the corresponding entries of $a'$.

\subsection{Computational Cost}
\label{sec:computational-cost}
\renewcommand{\arraystretch}{1.05}
\definecolor{lightgray}{gray}{0.9}
\colorlet{lightbrown}{brown!10}

\begin{wraptable}{r}{0.48\textwidth}
\vspace{-2em}
\centering
\small
\caption{
\textbf{LM-head proxy preserves \textit{SALT}-relevant geometry.}
Results are averaged over validation checkpoints and mini-batches.
}
\label{tab:proxy_main}
\vspace{1em}
    \resizebox{0.46\textwidth}{!}{%

\begin{tabular}{lccc}
\toprule
\textbf{Proxy}
& \textbf{Gram} $\rho$ $\uparrow$
& \textbf{PR err.} $\downarrow$
& \textbf{Subspace} $\uparrow$ \\
\midrule
\rowcolor{lightgray}\multicolumn{4}{l}{\textbf{Reference Gradients Baselines}} \\
\midrule
Full gradient
& $0.71$
& $0.09$
& $0.78$ \\
Blockwise sketch
& $0.76$
& $0.07$
& $0.81$ \\
Last block
& $0.79$
& $0.06$
& $0.84$ \\
\midrule
\rowcolor{lightgray}\multicolumn{4}{l}{\textbf{Null Baselines}} \\
\midrule
Permutation null
& $0.02$
& $0.47$
& $0.18$ \\
Noise null
& $0.00$
& $0.56$
& $0.13$ \\
\bottomrule
\end{tabular}
}
\vspace{-0.8em}
\end{wraptable}
\textit{SALT} uses mini-batch gradient geometry, but full-model
per-sample gradients are prohibitive for LLMs. We therefore use the
LM-head/output-projection gradient as a lightweight proxy, a standard
choice for scalable gradient embeddings and influence estimation
~\citep{badge,tracin,trak}. The proxy is not an entry-wise
approximation to the full gradient; \textbf{\textit{SALT} only requires
it to preserve the sample-wise geometry used for reweighting.} Table~\ref{tab:proxy_main} validates this requirement: the proxy
preserves \textit{SALT}-relevant geometry against full-gradient and
blockwise references, while null baselines fail. Details are in Appendix~\ref{app:proxy}.

Let the output logits be $z_{b,i,t}=W_{\mathrm{out}}h_{b,i,t}+b_{\mathrm{out}}$, where $h_{b,i,t}\in\mathbb{R}^{d}$ and $W_{\mathrm{out}}\in\mathbb{R}^{|\mathcal{V}|\times d}$. We use the response-level output-projection gradient as the proxy feature:
\begin{equation}
\label{eq:back_logit}
\nabla_{W_{\mathrm{out}}}\ell_{b,i}
=
\frac{1}{|o_{b,i}|}\sum_{t=1}^{|o_{b,i}|}
\big(\nabla_{z_{b,i,t}}\ell_{b,i,t}\big)h_{b,i,t}^{\top}.
\end{equation}
Thus, the geometry cost scales with the rollout mini-batch size $m=BG$ rather than the total parameter count: forming the $m\times m$ Gram matrix costs $O(m^2)$ memory, and extracting the top-$k_t$ subspace costs $O(m^2k_t)$ time with partial eigensolvers. Runtime results are provided in Appendix~\ref{app:time_cost}.

\subsection{\textit{SALT} as a Geometry-Reweighted Surrogate}
\label{sec:surrogate}

\textit{SALT} is not an unbiased reformulation of the original
GRPO estimator, but a geometry-reweighted surrogate that trades
controlled coefficient distortion for reduced signed cancellation. Let $m=BG$ and $S=[s_1,\ldots,s_m]$. 
Treating $a'$ as a stop-gradient coefficient, the original update,
the \textit{SALT} update, and the corresponding batch surrogate are
\begin{equation}
\label{eq:salt_surrogate}
\begin{aligned}
\hat g_{\mathrm{GRPO}}
&= \frac{1}{m}Sa,
&
\hat g_{\mathrm{SALT}}
&= \frac{1}{m}Sa',
&
\mathcal{J}_{\mathrm{SALT}}(\theta)
=
\frac{1}{m}\sum_{i=1}^{m}
a'_i \log \pi_\theta(o_i \mid q_i).
\end{aligned}
\end{equation}
instead of the original GRPO surrogate with coefficients $a_i$. Its deviation from GRPO satisfies
\begin{equation}
\hat g_{\mathrm{SALT}}-\hat g_{\mathrm{GRPO}}
=
\frac{1}{m}S(a'-a),
\qquad
\|\hat g_{\mathrm{SALT}}-\hat g_{\mathrm{GRPO}}\|_2
\le
\frac{1}{m}\|S\|_{\mathrm{op}}\|a'-a\|_2 .
\end{equation}
Thus, the surrogate deviation is controlled by the reweighting magnitude and mini-batch gradient geometry. When the original GRPO-style update is weakened by signed cancellation, \textit{SALT} adaptively increases the spectral-tail contribution through $n_{\mathrm{eff}}$. This is a biased surrogate: we do not claim true-policy-gradient unbiasedness, global convergence, or monotonic policy improvement. The next experiments section evaluates whether cancellation reduction outweighs coefficient distortion in the measured settings, and Appendix~\ref{app:bias_cancellation} quantifies the corresponding trade-off.

\section{Experiment}
In this section, we conduct extensive experiments to answer the following research questions:

\noindent\textbullet \ (\textbf{RQ1}) 
Does \textit{SALT} consistently improve GRPO-style performance across benchmarks and model scales, and are the gains cost-effective given the added geometry module?

\noindent\textbullet \ (\textbf{RQ2}) 
How does \textit{SALT} affect exploration and optimization dynamics compared with alternative exploration controls?

\noindent\textbullet \ (\textbf{RQ3}) 
Which components of \textit{SALT} are necessary, and how do they contribute to performance and geometric metrics?

\noindent\textbullet \ (\textbf{RQ4}) Does \textit{SALT} make larger rollout groups actually useful under a fixed rollout budget, and why?

\noindent\textbullet \ (\textbf{RQ5}) Does \textit{SALT} generalize across RLVR objectives and task domains?

\subsection{Experimental Setup}
\label{section:setup}
\paragraph{Data and Evaluation. } We evaluate  \textit{SALT} on GPQA-Diamond~\citep{rein2023gpqa} and four math reasoning benchmarks: AIME24~\citep{huggingfaceh4_aime2024}, AIME25~\citep{opencompass_aime2025}, GSM8K~\citep{cobbe2021gsm8k}  and MATH500~\citep{lightman2023lets}.
These datasets span scientific QA and mathematical problems. To ensure the stability of experiments, we run each experiment five times and report the average results with Accuracy and Pass@8.
For the original code setting, we train on MBPP~\citep{mbpp} and evaluate on HumanEval/HumanEval+~\citep{chen2021evaluating,evalplus} in Appendix~\ref{app:rq1}; additional code and logic evaluations are reported in Section~\ref{sec:generalization}.

\renewcommand{\arraystretch}{1.05}
\definecolor{lightgray}{gray}{0.9} 
\colorlet{lightbrown}{blue!10}
\colorlet{lightbrown}{brown!10}

\begin{table}[t]
    \centering
    \caption{Main results on reasoning benchmarks. We compare the GRPO and DAPO algorithms with our method \textit{SALT} across two different training datasets. 
    We report ACC and Pass@8 and results are averaged over $5$ runs. 
        \textbf{Bold} numbers indicate improvements with \textit{SALT}.
        Overall, \textbf{incorporating \textit{SALT} yields consistent improvements} over GRPO-style algorithms validated by significance tests.}
    \small
    \resizebox{0.97\textwidth}{!}{%
\begin{tabular}{clcccccccccc}
\toprule
\multirow{2}{*}{\textbf{Dataset}} & \multirow{2}{*}{\textbf{Method}} & \multicolumn{2}{c}{\textbf{AIME24}} & \multicolumn{2}{c}{\textbf{AIME25}} & \multicolumn{2}{c}{\textbf{GSM8K}} & \multicolumn{2}{c}{\textbf{MATH-500}} & \multicolumn{2}{c}{\textbf{GPQA}} \\
\cmidrule(lr){3-4}\cmidrule(lr){5-6}\cmidrule(lr){7-8}\cmidrule(lr){9-10}\cmidrule(lr){11-12}
  &      &    ACC       & Pass@8      & ACC       & Pass@8      & ACC     & Pass@8     & ACC      & Pass@8      & ACC       & Pass@8          \\ \midrule

\rowcolor{lightgray} \multicolumn{12}{c}{\textbf{Deepseek-Distill-Qwen-1.5B}}     \\ \midrule
$\sim$ & Vanilla&
29.1 & 59.3 & 22.0 & 42.6 & 80.3 & 95.0 & 85.5 & 96.4 & 34.5 & 82.4   \\ \midrule

\multirow{4}{*}{\textbf{MATH-TRAIN}}
& GRPO &
29.3 & 59.3 & 25.6 & 43.3 & 80.7 & 94.8 & 85.6 & 96.7 & 35.0 & 82.2 \\
& \cellcolor{lightbrown}\; + \textit{SALT}
& \cellcolor{lightbrown}\textbf{32.1} & \cellcolor{lightbrown}\textbf{62.9}
& \cellcolor{lightbrown}\textbf{27.1} & \cellcolor{lightbrown}\textbf{47.2}
& \cellcolor{lightbrown}\textbf{83.4} & \cellcolor{lightbrown}\textbf{96.5}
& \cellcolor{lightbrown}\textbf{87.2} & \cellcolor{lightbrown}\textbf{98.4}
& \cellcolor{lightbrown}\textbf{38.4} & \cellcolor{lightbrown}\textbf{85.5} \\
\cmidrule(lr){2-12}
& DAPO
& 31.1 & 59.9
& 23.9 & 44.3
& 81.0 & 96.1
& 86.1 & 97.0
& 37.0 & 81.7 \\
& \cellcolor{lightbrown}\; + \textit{SALT}
& \cellcolor{lightbrown}30.6 & \cellcolor{lightbrown}\textbf{60.6}
& \cellcolor{lightbrown}\textbf{25.6} & \cellcolor{lightbrown}\textbf{46.3}
& \cellcolor{lightbrown}\textbf{82.5} & \cellcolor{lightbrown}\textbf{97.1}
& \cellcolor{lightbrown}\textbf{87.1} & \cellcolor{lightbrown}97.0
& \cellcolor{lightbrown}\textbf{39.6} & \cellcolor{lightbrown}\textbf{83.5} \\
\midrule

\multirow{4}{*}{\textbf{DAPO-MATH}}
& GRPO
& 29.1 & 59.0
& 23.6 & 44.2
& 82.2 & 95.3
& 85.9 & 96.8
& 36.2 & 83.7 \\
& \cellcolor{lightbrown}\; + \textit{SALT}
& \cellcolor{lightbrown}\textbf{32.7} & \cellcolor{lightbrown}\textbf{62.8}
& \cellcolor{lightbrown}\textbf{27.8} & \cellcolor{lightbrown}\textbf{47.8}
& \cellcolor{lightbrown}\textbf{83.0} & \cellcolor{lightbrown}\textbf{96.5}
& \cellcolor{lightbrown}\textbf{88.0} & \cellcolor{lightbrown}\textbf{97.3}
& \cellcolor{lightbrown}\textbf{37.7} & \cellcolor{lightbrown}\textbf{85.0} \\
\cmidrule(lr){2-12}
& DAPO
& 29.0 & 61.3
& 26.0 & 45.0
& 82.2 & 95.0
& 84.8 & 96.8
& 36.0 & 80.9 \\
& \cellcolor{lightbrown}\; + \textit{SALT}
& \cellcolor{lightbrown}\textbf{32.9} & \cellcolor{lightbrown}\textbf{63.8}
& \cellcolor{lightbrown}\textbf{28.0} & \cellcolor{lightbrown}\textbf{47.0}
& \cellcolor{lightbrown}\textbf{84.1} & \cellcolor{lightbrown}\textbf{96.8}
& \cellcolor{lightbrown}\textbf{88.0} & \cellcolor{lightbrown}\textbf{98.0}
& \cellcolor{lightbrown}\textbf{38.5} & \cellcolor{lightbrown}\textbf{82.9} \\
\midrule

\rowcolor{lightgray} \multicolumn{12}{c}{\textbf{Deepseek-Distill-Qwen-7B}}     \\ \midrule
$\sim$ & Vanilla
& 50.7 & 78.0
& 36.7 & 64.0
& 90.3 & 97.4
& 94.9 & 98.8
& 48.9 & 82.4 \\
\midrule

\multirow{4}{*}{\textbf{MATH-TRAIN}}
& GRPO
& 60.7 & 77.3
& 42.0 & 71.3
& 89.9 & 97.7
& 94.9 & 99.0
& 46.8 & 83.8 \\
& \cellcolor{lightbrown}\;\; + \textit{SALT}
& \cellcolor{lightbrown}\textbf{64.7} & \cellcolor{lightbrown}\textbf{80.4}
& \cellcolor{lightbrown}\textbf{45.3} & \cellcolor{lightbrown}\textbf{74.5}
& \cellcolor{lightbrown}\textbf{91.9} & \cellcolor{lightbrown}\textbf{98.7}
& \cellcolor{lightbrown}\textbf{95.2} & \cellcolor{lightbrown}98.9
& \cellcolor{lightbrown}\textbf{51.2} & \cellcolor{lightbrown}\textbf{87.9} \\
\cmidrule(lr){2-12}
& DAPO
& 54.0 & 79.3
& 42.0 & 62.7
& 90.1 & 97.5
& 94.2 & 99.0
& 48.9 & 83.7 \\
& \cellcolor{lightbrown}\;\; + \textit{SALT}
& \cellcolor{lightbrown}\textbf{57.3} & \cellcolor{lightbrown}\textbf{82.2}
& \cellcolor{lightbrown}\textbf{44.8} & \cellcolor{lightbrown}\textbf{66.6}
& \cellcolor{lightbrown}\textbf{91.7} & \cellcolor{lightbrown}\textbf{99.0}
& \cellcolor{lightbrown}\textbf{95.2} & \cellcolor{lightbrown}\textbf{99.1}
& \cellcolor{lightbrown}\textbf{52.2} & \cellcolor{lightbrown}\textbf{88.1} \\
\midrule

\multirow{4}{*}{\textbf{DAPO-MATH}}
& GRPO
& 61.5 & 78.0
& 42.8 & 72.1
& 90.4 & 97.8
& 95.0 & 99.1
& 47.6 & 84.6 \\
& \cellcolor{lightbrown}\;\; + \textit{SALT}
& \cellcolor{lightbrown}\textbf{65.4} & \cellcolor{lightbrown}\textbf{81.2}
& \cellcolor{lightbrown}\textbf{46.5} & \cellcolor{lightbrown}\textbf{75.5}
& \cellcolor{lightbrown}\textbf{92.4} & \cellcolor{lightbrown}\textbf{99.3}
& \cellcolor{lightbrown}\textbf{96.4} & \cellcolor{lightbrown}\textbf{99.2}
& \cellcolor{lightbrown}\textbf{51.3} & \cellcolor{lightbrown}\textbf{89.2} \\
\cmidrule(lr){2-12}
& DAPO
& 54.8 & 80.0
& 43.0 & 64.0
& 90.4 & 97.6
& 94.4 & 99.1
& 49.2 & 84.1 \\
& \cellcolor{lightbrown}\;\; + \textit{SALT}
& \cellcolor{lightbrown}\textbf{56.6} & \cellcolor{lightbrown}\textbf{81.6}
& \cellcolor{lightbrown}\textbf{46.3} & \cellcolor{lightbrown}\textbf{68.3}
& \cellcolor{lightbrown}\textbf{92.1} & \cellcolor{lightbrown}\textbf{99.2}
& \cellcolor{lightbrown}\textbf{95.4} & \cellcolor{lightbrown}\textbf{99.3}
& \cellcolor{lightbrown}\textbf{53.1} & \cellcolor{lightbrown}\textbf{88.7} \\ 
\bottomrule
\end{tabular}
}
\label{tab:main_experiment}
\end{table}

\paragraph{Training Dataset. } We train our policy on two complementary math datasets for $2$ epochs: MATH-Train-7.5k~\citep{hendrycksmath2021} and DAPO-Math-17k~\citep{yu2025dapoopensourcellmreinforcement}, aiming for a broader and more robust evaluation of training effectiveness across diverse math domains.

\paragraph{Settings.}
We use DeepSeek-Distill-Qwen~\citep{deepseekai2025deepseekr1incentivizingreasoningcapability} models at two scales (1.5B and 7B) as the base policy for training.
We adopt a rule-based verifier to provide the reward signal.
For each question we sample $8$ responses with temperature $0.6$ during training; we use the same sampling configuration for validation.
Additional hyperparameters and implementation details are deferred to the Appendix~\ref{app:details}.

\subsection{Main Result}
\label{sec:main_result}
\renewcommand{\arraystretch}{1.05}
\definecolor{lightgray}{gray}{0.9}
\colorlet{lightbrown}{brown!10}

\begin{wraptable}{r}{0.48\textwidth}
    \vspace{-1.8em}
    \centering
    \small
    \caption{Average performance aggregated over reasoning benchmarks under GRPO and DAPO training. \textbf{Bold} denotes better performance.}
    \vspace{1em}
    \setlength{\tabcolsep}{2.5pt}
    \resizebox{0.45\textwidth}{!}{%
    \begin{tabular}{llcc}
    \toprule
\multirow{1}{*}{\textbf{Dataset}}   & \multirow{1}{*}{\textbf{Method}} 
    &  \textbf{Pass@1} & \textbf{Pass@8}  \\
    \midrule

    \multirow{10}{*}{\makecell{\textbf{MATH-TRAIN}}}
    & GRPO & 51.2 & 75.3 \\
    & \; + Entropy Bonus & 51.0 & 75.1 \\
    & \; + Clip Cov & 51.6 & \underline{75.8} \\
    & \; + KL Cov & \underline{51.8} & \underline{75.8} \\
    & \cellcolor{lightbrown}\; + \textit{SALT}
    & \cellcolor{lightbrown}\textbf{53.6}
    & \cellcolor{lightbrown}\textbf{78.1} \\
    \cmidrule(lr){2-4}
    & DAPO & 51.8 & 75.8 \\
    & \; + Entropy Bonus & 51.7 & 75.9 \\
    & \; + Clip Cov & 51.9 & 76.0 \\
    & \; + KL Cov & \underline{52.0} & \underline{76.1} \\
    & \cellcolor{lightbrown}\; + \textit{SALT}
    & \cellcolor{lightbrown}\textbf{53.1}
    & \cellcolor{lightbrown}\textbf{76.9} \\

    \midrule

    \multirow{10}{*}{\makecell{\textbf{DAPO-MATH}}}    & GRPO & 51.4 & 75.8 \\
    & \; + Entropy Bonus & 51.6 & 76.0 \\
    & \; + Clip Cov & 52.1 & 75.6 \\
    & \; + KL Cov & \underline{52.2} & \underline{76.4} \\
    & \cellcolor{lightbrown}\; + \textit{SALT}
    & \cellcolor{lightbrown}\textbf{53.8}
    & \cellcolor{lightbrown}\textbf{77.9} \\
    \cmidrule(lr){2-4}
    & DAPO & 51.6 & 75.8 \\
    & \; + Entropy Bonus & 51.8 & 76.1 \\
    & \; + Clip Cov & 52.3 & 75.8 \\
    & \; + KL Cov & \underline{52.5} & \underline{76.4} \\
    & \cellcolor{lightbrown}\; + \textit{SALT}
    & \cellcolor{lightbrown}\textbf{54.3}
    & \cellcolor{lightbrown}\textbf{77.7} \\

    \bottomrule
    \end{tabular}%
    }
    \label{tab:rq2}
    \vspace{-3em}
\end{wraptable}

\paragraph{(For RQ1-1) \textit{SALT} consistently improves GRPO-style training.}
Table~\ref{tab:main_experiment} compares GRPO/DAPO with \textit{SALT} under the same training and sampling setup. Averaged over tasks, \textit{SALT} improves the corresponding backbone by about +2.41 percentage points per setting, with paired 95\% CIs in Table~\ref{tab:sign}. Gains are largest on AIME24 and GPQA, with no consistent degradation. These results show that \textit{SALT} robustly improves rollout aggregation, increasing both accuracy and pass@8 without extra exploration budget. Beyond the main setting, Appendix~\ref{app:rq1} shows cross-model gains of +3 acc on average across additional model families, including Qwen3-8B~\citep{qwen3technicalreport} and LLaMA-3.1-8B~\citep{llama3modelcard}, and further demonstrates that \textit{SALT} improves acc on HumanEval~\citep{chen2021evaluating} in execution-verifiable code RLVR.

\paragraph{(For RQ1-2) \textit{SALT} provides favorable cost--benefit.}
Since \textit{SALT} adds a geometry module, we further evaluate whether its gains justify the extra wall-clock cost. Appendix~\ref{app:time_cost} reports end-to-end training time. Across models and training sets, \textit{SALT} adds about 7.7\% average overhead in training time while improving both geometric effective metrics and accuracy by an average of 2.41 points. Thus, the controlled departure from GRPO in Section~\ref{sec:surrogate} is practically worthwhile, yielding measurable gains with modest extra cost.
\begin{figure*}[t]
    \centering
    \includegraphics[width=0.95\linewidth]{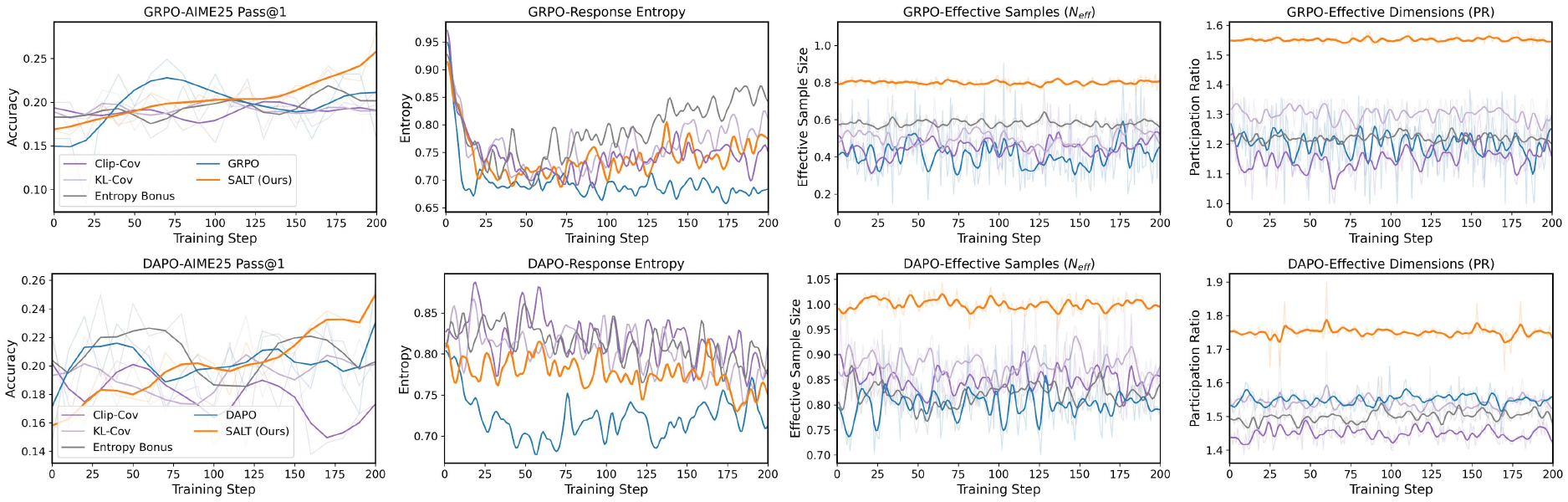}     
    \vspace{-0.3em}
    \caption{Training dynamics on MATH-TRAIN with GRPO (top) and DAPO (bottom), showing AIME25 accuracy, response entropy, $PR$, and $n_{\mathrm{eff}}$. Entropy-based baselines increase response entropy but yield limited gains in geometry metrics and accuracy. In contrast, \textit{SALT} improves both $PR$ and $n_{\mathrm{eff}}$, \textbf{achieving stronger performance via update exploration, not higher entropy.}}
    \label{fig:rq2}
\end{figure*}

\paragraph{(For RQ2) \textit{SALT} promotes effective exploration beyond entropy-increasing controls.}
Figure~\ref{fig:rq2} tracks AIME25 test performance together with response entropy, effective dimensionality $PR$ (Eq.~\eqref{eq:pr_k}), and effective sample size $n_{\mathrm{eff}}$ (Eq.~\eqref{eq:neff}) during GRPO training on both MATH-TRAIN and DAPO-MATH.
Entropy-oriented baselines (entropy bonus, clip-cov, KL-cov~\citep{cui2025entropymechanismreinforcementlearning,shen2025entropycontrolllmrlalgorithms,mnih2016asynchronousmethodsdeepreinforcement}) increase entropy as intended, but their improvements in $PR$ and $n_{\mathrm{eff}}$ remain limited and do not consistently translate into better AIME25 performance.
In contrast, \textit{SALT} achieves the strongest AIME25 results while simultaneously yielding higher $PR$ and $n_{\mathrm{eff}}$, indicating that \textit{SALT} turns rollouts into more diverse and less-canceling update signals rather than merely amplifying sampling entropy.

Table~\ref{tab:rq2} further confirms this trend across diverse benchmarks, where \textit{SALT} consistently outperforms standard GRPO and entropy-promoting controls. 
In particular, improvement is observed on both the 1.5B and 7B scales  
(see Appendix~\ref{app:rq2} for full results and additional analysis).

\subsection{Ablation Study}
\label{sec:ablation}

\begin{figure}[t]
    \centering
    \includegraphics[width=0.95\linewidth]{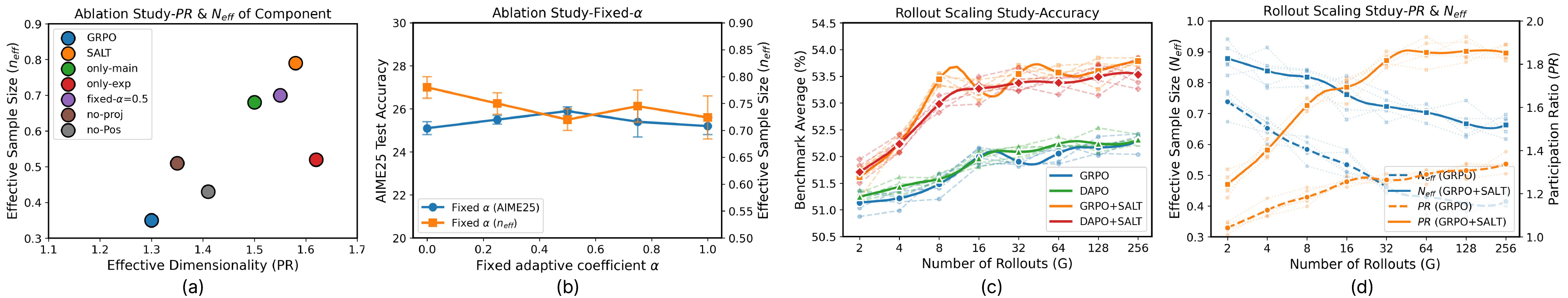}

\caption{\textbf{Ablation and rollout-scaling analysis of \textit{SALT}.}
(a) \textit{SALT} achieves the best PR--$n_{\mathrm{eff}}$ trade-off; ablations degrade update geometry.
(b) Fixed-$\alpha$ exposes a brittle exploration--cancellation trade-off avoided by adaptive mixing.
(c--d) With a fixed rollout budget, \textit{SALT} better uses larger groups by raising $PR$ while preserving $n_{\mathrm{eff}}$, yielding consistent gains over GRPO/DAPO.}
    \label{fig:rq3}

    \vspace{-1em}
\end{figure}

\paragraph{(For RQ3) Both the dual-channel structure and cancellation-aware mixing are necessary.}
Figure~\ref{fig:rq3}(a) compares \textit{SALT} with its ablations in the PR--$n_{\mathrm{eff}}$ plane.
Removing either channel hurts update geometry: \textit{only-main} reduces PR, while \textit{only-exp} increases cancellation and lowers $n_{\mathrm{eff}}$.
Other controls show similar degradation: \textit{no-proj}/\textit{rand-proj} move toward the baseline, \textit{no-Pos} reduces $n_{\mathrm{eff}}$, and alternative mixing rules fail to reach the high-PR/high-$n_{\mathrm{eff}}$ regime.
In contrast, full \textit{SALT} achieves the best geometry and benchmark performance, indicating more diverse and less-canceling updates.
Details are provided in Appendix~\ref{app:ablation}.

Figure~\ref{fig:rq3} (b) sweeps a fixed mixing coefficient $\alpha\in\{0,0.25,0.5,0.75,1.0\}$ and exposes a clear trade-off:
larger $\alpha$ boosts exploration but lowers $n_{\mathrm{eff}}$ via stronger gradient cancellation, while smaller $\alpha$ preserves $n_{\mathrm{eff}}$ but under-explores and hurts accuracy.
This sensitivity makes fixed-$\alpha$ tuning brittle and \textit{SALT} avoids it by adapting the mixing strength online. It further shows the realized adaptive $\alpha_t$ over training, illustrating that \textit{SALT} replaces this brittle knob with an online-adjusted mixing signal.

Beyond these component ablations, Appendix~\ref{app:matched_intervention} further introduces a matched-coefficient subspace intervention, which preserves the coefficient distribution of \textit{SALT} while breaking its alignment with the true mini-batch gradient geometry. The degradation under these matched null variants, together with consistently low clipping fractions, suggests that \textbf{\textit{SALT}'s gains come from geometry-aligned coefficient assignment rather than coefficient redistribution or PPO clipping artifacts.}




\subsection{Rollout Scaling Study}
\label{sec:rollout}
\paragraph{(For RQ4-1) \textit{SALT} makes larger rollout groups useful under a fixed budget.} We study whether \textit{SALT} makes larger rollout groups $G$ useful under a fixed rollout budget. Although increasing $G$ provides more rollouts, it can also amplify gradient cancellation, so additional samples may not translate into stronger updates. As shown in Figure~\ref{fig:rq3} (c), when $G$ increases from 2 to 256, GRPO and DAPO exhibit diminishing returns, whereas \textit{SALT} yields consistent gains across group sizes, with larger improvements at larger $G$. This indicates that \textit{SALT} better utilizes additional rollouts under the same rollout budget, rather than merely benefiting from more compute.

\vspace{-0.3em}

\paragraph{(For RQ4-2) Why: \textit{SALT} increases diversity while mitigating cancellation via adaptive mixing.}
We also track participation ratio ($PR$) for diversity and effective sample size $n_{\mathrm{eff}}$ for gradient alignment in Figure~\ref{fig:rq3} (d). As $G$ increases, $PR$ grows for both methods, but baseline $n_{\mathrm{eff}}$ decreases, revealing stronger cancellation. SALT maintains higher $n_{\mathrm{eff}}$ while further improving $PR$, allowing larger $G$ to translate into effective updates. 
Overall, \textit{SALT} enables large-$G$ scaling under a fixed budget by jointly increasing $PR$ and maintaining $n_{\mathrm{eff}}$ through adaptive main/exploration mixing. Additional mechanism evidence is provided in Appendix~\ref{app:rq4}.

\vspace{-0.3em}

\subsection{Generalization across Objectives and Domains}
\label{sec:generalization}
\renewcommand{\arraystretch}{1.05}
\colorlet{lightbrown}{brown!10}

\begin{table}[t]
    \centering
    \caption{\textbf{Generalization across RLVR objectives and task domains.} (a) Average Pass@1/Pass@8 over AIME24, AIME25, GSM8K, MATH-500, and GPQA using DeepSeek-Distill-Qwen-7B trained on MATH-TRAIN. Each objective is evaluated with and without \textit{SALT} under a matched training budget. (b) Pass@1 on code and logic benchmarks; \textbf{Canc.} denotes the mean cancellation severity measured during training. Results are averaged over five seeds. 
    }
    \label{tab:rq5}
    \fontsize{8.2}{9.4}\selectfont
    \begin{minipage}[t]{0.485\textwidth}
        \centering
        \vspace{0pt}
        \textbf{(a) Across RLVR objectives}
        \vspace{0.25em}

        \setlength{\tabcolsep}{2.0pt}
        \begin{tabular*}{\linewidth}{@{\extracolsep{\fill}}lccc@{}}
            \toprule
            \textbf{Objective} & \textbf{Base} & \textbf{+\textit{SALT}} & \textbf{$\Delta$ P@1} \\
            \midrule
            GRPO  & 66.9/85.8 & \cellcolor{lightbrown}\textbf{69.7/88.1} & \textbf{+2.8} \\
            DAPO  & 65.8/84.4 & \cellcolor{lightbrown}\textbf{68.2/87.0} & \textbf{+2.4} \\
            CISPO & 67.4/86.2 & \cellcolor{lightbrown}\textbf{69.9/88.4} & \textbf{+2.5} \\
            GSPO  & 67.8/86.5 & \cellcolor{lightbrown}\textbf{70.2/88.6} & \textbf{+2.4} \\
            DPPO  & 68.1/86.9 & \cellcolor{lightbrown}\textbf{70.5/88.9} & \textbf{+2.4} \\
            \bottomrule
        \end{tabular*}
    \end{minipage}
    \hfill
    \begin{minipage}[t]{0.485\textwidth}
        \centering
        \vspace{0pt}
        \textbf{(b) Across task domains}
        \vspace{0.25em}

        \setlength{\tabcolsep}{1.5pt}
        \begin{tabular*}{\linewidth}{@{\extracolsep{\fill}}lcccc@{}}
            \toprule
            \textbf{Benchmark} & \textbf{GRPO} & \textbf{+\textit{SALT}} & \textbf{$\Delta$} & \textbf{Canc.} \\
            \midrule
            HumanEval        & 78.0 & \cellcolor{lightbrown}\textbf{79.3} & \textbf{+1.3} & 0.21 \\
            HumanEval+       & 72.4 & \cellcolor{lightbrown}\textbf{73.5} & \textbf{+1.1} & 0.21 \\
            LiveCodeBench-v5 & 24.6 & \cellcolor{lightbrown}\textbf{27.0} & \textbf{+2.4} & 0.38 \\
            ZebraLogic       & 31.2 & \cellcolor{lightbrown}\textbf{34.6} & \textbf{+3.4} & 0.52 \\
            K\&K held-out    & 58.4 & \cellcolor{lightbrown}\textbf{62.1} & \textbf{+3.7} & 0.55 \\
            \bottomrule
        \end{tabular*}
    \end{minipage}
\end{table}

\paragraph{(For RQ5-1) \textit{SALT} generalizes across RLVR objectives.}
Table~\ref{tab:rq5}(a) compares matched-budget training with GRPO, DAPO, CISPO~\citep{minimax2025cispo}, GSPO~\citep{zheng2025gspo}, and DPPO~\citep{qi2026dppo}. Adding \textit{SALT} improves every objective by 2.4--2.8 Pass@1 points, indicating that its sample-geometry reweighting is complementary to different update constraints rather than specific to one clipping formulation.

\paragraph{(For RQ5-2) \textit{SALT} generalizes beyond mathematical reasoning.}
Table~\ref{tab:rq5}(b) extends the evaluation to code and logic. The gain is modest when measured cancellation is mild on HumanEval/HumanEval+~\citep{chen2021evaluating,evalplus}, increases on LiveCodeBench-v5~\citep{jain2025livecodebench}, and is largest on ZebraLogic~\citep{lin2025zebralogic} and K\&K~\citep{xie2024knightsknaves}, where cancellation is strongest. Thus, the cross-domain gains track the diagnosed cancellation severity rather than requiring a uniform intervention across tasks.

\vspace{-0.3em}

\section{Related Work}
\vspace{-0.3em}

\paragraph{Reinforcement Learning in Reasoning.}
Recent progress in language model reasoning has increasingly relied on reinforcement learning (RL) to improve multi-step reasoning by directly optimizing task-level objectives on model-generated solutions~\citep{ding2025multilayergrpoenhancingreasoning}. 
A representative and widely used PPO-style~\citep{ppo} approach in this setting is \emph{Group Relative Policy Optimization}, which constructs group-relative advantages from multiple completions per prompt without an explicit critic~\citep{grpo,deepseekai2025deepseekr1}.

\vspace{-0.3em}

\paragraph{Optimization of GRPO.}
Recent GRPO variants improve training stability by modifying the optimization objective: DAPO introduces decoupled clipping and dynamic sampling~\citep{yu2025dapoopensourcellmreinforcement}, while GSPO uses sequence-level ratios to stabilize RL for large Mixture-of-Experts models~\citep{zheng2025gspo}. Other works modify update constraints, reweight advantages, or add entropy-aware advantages, self-correction, and multi-stage refinement to improve reasoning~\citep{minimax2025cispo,qi2026dppo,plyusov2026fgrpo,zhang2025grpolead,zhang2025edgegrpoentropydrivengrpoguided,yao2025diversity_neurips,li2025scrpoerrorsinsights,feng2025groupingrouppolicyoptimizationllm,zhou2025evolvinglanguagemodelslabels}. However, these methods mainly reshape gradient signals rather than \textbf{explicitly modeling gradient features}, leaving an important gap.

\vspace{-0.3em}

\section{Conclusion}


\vspace{-0.3em}
\paragraph{Limitations.}
\textit{SALT} uses an LM-head gradient proxy and extra Gram computation, so its efficiency and geometry estimates may vary across architectures and training regimes. It is also a biased geometry-reweighted surrogate, whose gains depend on reducing signed cancellation enough to offset coefficient distortion; we do not claim true-policy-gradient unbiasedness, global convergence, or monotonic policy improvement.
\vspace{-0.3em}
\paragraph{Future work.}
Beyond the math and code RLVR settings studied here, future work will explore noisier learned rewards, multi-modal policies, agentic or search-based tasks, and other forms of group-based policy optimization.
\emph{Less redundancy, more progress.}

\section*{Acknowledgments and Disclosure of Funding}
This work was conducted at Bilibili Inc. and was supported by Bilibili Inc. through personnel time, computing resources, and research infrastructure. The authors declare no competing interests beyond the affiliations and support disclosed in this paper.

\bibliographystyle{plainnat}
\bibliography{references}

@article{grpo,
  title   = {DeepSeekMath: Pushing the Limits of Mathematical Reasoning in Open Language Models},
  author  = {Shao, Zhihong and Wang, Peiyi and Zhu, Qihao and Xu, Runxin and Song, Junxiao and Bi, Xiao and Zhang, Haowei and Zhang, Mingchuan and Li, Y. K. and Wu, Yu and Guo, Daya},
  journal = {arXiv preprint arXiv:2402.03300},
  year    = {2024},
  doi     = {10.48550/arXiv.2402.03300},
  url     = {https://arxiv.org/abs/2402.03300}
}

@article{chang2026spice,
  title={SPICE: Submodular Penalized Information-Conflict Selection for Efficient Large Language Model Training},
  author={Chang, Powei and Zhang, Jinpeng and Chen, Bowen and Wang, Chenyu and Guo, Chenlu and Zhang, Yixing and Gao, Yukang and Xiang, JianXiang and Gao, Yue and Sun, Chaoqun and others},
  journal={arXiv preprint arXiv:2601.23155},
  year={2026}
}

@article{ppo,
  title   = {Proximal Policy Optimization Algorithms},
  author  = {Schulman, John and Wolski, Filip and Dhariwal, Prafulla and Radford, Alec and Klimov, Oleg},
  journal = {arXiv preprint arXiv:1707.06347},
  year    = {2017},
  doi     = {10.48550/arXiv.1707.06347},
  url     = {https://arxiv.org/abs/1707.06347}
}

@article{deepseekai2025deepseekr1,
  title   = {DeepSeek-R1: Incentivizing Reasoning Capability in LLMs via Reinforcement Learning},
  author  = {DeepSeek-AI and Guo, Daya and Yang, Dejian and Zhang, Haowei and Song, Junxiao and Zhang, Ruoyu and Xu, Runxin and Zhu, Qihao and Ma, Shirong and Wang, Peiyi and others},
  journal = {arXiv preprint arXiv:2501.12948},
  year    = {2025},
  doi     = {10.48550/arXiv.2501.12948},
  url     = {https://arxiv.org/abs/2501.12948}
}

@misc{yu2025dapoopensourcellmreinforcement,
  title         = {DAPO: An Open-Source LLM Reinforcement Learning System at Scale},
  author        = {Yu, Qiying and Zhang, Zheng and Zhu, Ruofei and Yuan, Yufeng and Zuo, Xiaochen and Yue, Yu and Dai, Weinan and Fan, Tiantian and Liu, Gaohong and Liu, Lingjun and Liu, Xin and Lin, Haibin and Lin, Zhiqi and Ma, Bole and Sheng, Guangming and Tong, Yuxuan and Zhang, Chi and Zhang, Mofan and Zhang, Wang and Zhu, Hang and Zhu, Jinhua and Chen, Jiaze and Chen, Jiangjie and Wang, Chengyi and Yu, Hongli and Song, Yuxuan and Wei, Xiangpeng and Zhou, Hao and Liu, Jingjing and Ma, Wei-Ying and Zhang, Ya-Qin and Wang, Mingxuan and Yan, Lin and Qiao, Mu and Wu, Yonghui},
  year          = {2025},
  eprint        = {2503.14476},
  archivePrefix = {arXiv},
  primaryClass  = {cs.LG},
  url           = {https://arxiv.org/abs/2503.14476}
}

@article{zheng2025gspo,
  title   = {Group Sequence Policy Optimization},
  author  = {Zheng, Chujie and Liu, Shixuan and Li, Mingze and Chen, Xiong-Hui and Yu, Bowen and Gao, Chang and Dang, Kai and Liu, Yuqiong and Men, Rui and Yang, An and Zhou, Jingren and Lin, Junyang},
  journal = {arXiv preprint arXiv:2507.18071},
  year    = {2025},
  doi     = {10.48550/arXiv.2507.18071},
  url     = {https://arxiv.org/abs/2507.18071}
}

@article{minimax2025cispo,
  title   = {MiniMax-M1: Scaling Test-Time Compute Efficiently with Lightning Attention},
  author  = {{MiniMax} and Chen, Aili and Li, Aonian and Gong, Bangwei and Jiang, Binyang and Fei, Bo and others},
  journal = {arXiv preprint arXiv:2506.13585},
  year    = {2025},
  doi     = {10.48550/arXiv.2506.13585},
  url     = {https://arxiv.org/abs/2506.13585}
}

@article{qi2026dppo,
  title   = {Rethinking the Trust Region in LLM Reinforcement Learning},
  author  = {Qi, Penghui and Zhou, Xiangxin and Liu, Zichen and Pang, Tianyu and Du, Chao and Lin, Min and Lee, Wee Sun},
  journal = {arXiv preprint arXiv:2602.04879},
  year    = {2026},
  doi     = {10.48550/arXiv.2602.04879},
  url     = {https://arxiv.org/abs/2602.04879}
}

@article{plyusov2026fgrpo,
  title   = {F-GRPO: Don't Let Your Policy Learn the Obvious and Forget the Rare},
  author  = {Plyusov, Daniil and Gorbatovski, Alexey and Shaposhnikov, Boris and Sinii, Viacheslav and Malakhov, Alexey and Korotyshova, Daria and Gavrilov, Daniil},
  journal = {arXiv preprint arXiv:2602.06717},
  year    = {2026},
  doi     = {10.48550/arXiv.2602.06717},
  url     = {https://arxiv.org/abs/2602.06717}
}

@inproceedings{zhang2025grpolead,
  title     = {{GRPO-LEAD}: A Difficulty-Aware Reinforcement Learning Approach for Concise Mathematical Reasoning in Language Models},
  author    = {Zhang, Jixiao and Zuo, Chunsheng},
  booktitle = {Proceedings of the 2025 Conference on Empirical Methods in Natural Language Processing},
  pages     = {5642--5654},
  year      = {2025},
  url       = {https://aclanthology.org/2025.emnlp-main.287/}
}

@misc{ding2025multilayergrpoenhancingreasoning,
      title={Multi-Layer GRPO: Enhancing Reasoning and Self-Correction in Large Language Models}, 
      author={Fei Ding and Baiqiao Wang and Zijian Zeng and Youwei Wang},
      year={2025},
      eprint={2506.04746},
      archivePrefix={arXiv},
      primaryClass={cs.LG},
      url={https://arxiv.org/abs/2506.04746}, 
}

@misc{zhang2025edgegrpoentropydrivengrpoguided,
      title={EDGE-GRPO: Entropy-Driven GRPO with Guided Error Correction for Advantage Diversity}, 
      author={Xingjian Zhang and Siwei Wen and Wenjun Wu and Lei Huang},
      year={2025},
      eprint={2507.21848},
      archivePrefix={arXiv},
      primaryClass={cs.AI},
      url={https://arxiv.org/abs/2507.21848}, 
}

@misc{li2025scrpoerrorsinsights,
      title={ScRPO: From Errors to Insights}, 
      author={Lianrui Li and Dakuan Lu and Jiawei Shao and Chi Zhang and Xuelong Li},
      year={2025},
      eprint={2511.06065},
      archivePrefix={arXiv},
      primaryClass={cs.AI},
      url={https://arxiv.org/abs/2511.06065}, 
}

@misc{feng2025groupingrouppolicyoptimizationllm,
      title={Group-in-Group Policy Optimization for LLM Agent Training}, 
      author={Lang Feng and Zhenghai Xue and Tingcong Liu and Bo An},
      year={2025},
      eprint={2505.10978},
      archivePrefix={arXiv},
      primaryClass={cs.LG},
      url={https://arxiv.org/abs/2505.10978}, 
}

@misc{zhou2025evolvinglanguagemodelslabels,
      title={Evolving Language Models without Labels: Majority Drives Selection, Novelty Promotes Variation}, 
      author={Yujun Zhou and Zhenwen Liang and Haolin Liu and Wenhao Yu and Kishan Panaganti and Linfeng Song and Dian Yu and Xiangliang Zhang and Haitao Mi and Dong Yu},
      year={2025},
      eprint={2509.15194},
      archivePrefix={arXiv},
      primaryClass={cs.LG},
      url={https://arxiv.org/abs/2509.15194}, 
}

@book{sutton2018reinforcement,
  title     = {Reinforcement Learning: An Introduction},
  author    = {Sutton, Richard S. and Barto, Andrew G.},
  edition   = {2},
  year      = {2018},
  publisher = {MIT Press},
  address   = {Cambridge, MA}
}

@inproceedings{yao2025diversity_neurips,
  title     = {Diversity-Aware Policy Optimization for Large Language Model Reasoning},
  author    = {Yao, Jian and Cheng, Ran and Wu, Xingyu and Wu, Jibin and Tan, Kay Chen},
  booktitle = {Advances in Neural Information Processing Systems (NeurIPS)},
  year      = {2025},
  note      = {Spotlight},
  url       = {https://openreview.net/forum?id=5eZ0iykpDU}
}

@inproceedings{pmlr-v84-yin18a,
  title     = {Gradient Diversity: a Key Ingredient for Scalable Distributed Learning},
  author    = {Yin, Dong and Pananjady, Ashwin and Lam, Max and Papailiopoulos, Dimitris and Ramchandran, Kannan and Bartlett, Peter},
  booktitle = {Proceedings of the Twenty-First International Conference on Artificial Intelligence and Statistics},
  pages     = {1998--2007},
  year      = {2018},
  volume    = {84},
  series    = {Proceedings of Machine Learning Research},
  publisher = {PMLR},
  url       = {https://proceedings.mlr.press/v84/yin18a.html}
}

@article{Recanatesi2022ScaleDependent,
  title        = {A Scale-Dependent Measure of System Dimensionality},
  author       = {Stefano Recanatesi and Serena Bradde and Vijay Balasubramanian and Nicholas A. Steinmetz and Eric Shea-Brown},
  journal      = {Patterns},
  volume       = {3},
  number       = {8},
  pages        = {100555},
  year         = {2022},
  publisher    = {Elsevier},
  doi          = {10.1016/j.patter.2022.100555},
  url          = {https://www.sciencedirect.com/science/article/pii/S266638992200160X},
}

@article{Hazelden2025FastNTK,
  title        = {Fast Neural Tangent Kernel Alignment, Norm and Effective Rank via Trace Estimation},
  author       = {James Hazelden},
  journal      = {arXiv:2511.10796},
  year         = {2025},
  archivePrefix= {arXiv},
  eprint       = {2511.10796},
  doi          = {10.48550/arXiv.2511.10796},
  url          = {https://www.arxiv.org/abs/2511.10796},
}

@book{Kish1965SurveySampling,
  author    = {Leslie Kish},
  title     = {Survey Sampling},
  publisher = {John Wiley \& Sons},
  year      = {1965},
}

@article{Yang2006EffectiveSampleSize,
  title        = {Effective Sample Size: Quick Estimation of the Effect of Related Samples in Genetic Case-Control Association Analyses},
  author       = {Yaning Yang and Elaine F. Remmers and Chukwuma B. Ogunwole and Daniel L. Kastner and Peter K. Gregersen and Wentian Li},
  journal      = {arXiv: q-bio/0611093},
  year         = {2006},
  archivePrefix= {arXiv},
  eprint       = {q-bio/0611093},
  url          = {https://arxiv.org/abs/q-bio/0611093},
}

@inproceedings{zhestiankin-ponomareva-2021-zhestyatsky,
  title = "Zhestyatsky at {S}em{E}val-2021 Task 2: {R}e{LU} over Cosine Similarity for {BERT} Fine-tuning",
  author = "Zhestiankin, Boris and Ponomareva, Maria",
  editor = "Palmer, Alexis and Schneider, Nathan and Schluter, Natalie and Emerson, Guy and Herbelot, Aurelie and Zhu, Xiaodan",
  booktitle = "Proceedings of the 15th International Workshop on Semantic Evaluation (SemEval-2021)",
  month = aug,
  year = "2021",
  address = "Online",
  publisher = "Association for Computational Linguistics",
  url = "https://aclanthology.org/2021.semeval-1.17/",
  doi = "10.18653/v1/2021.semeval-1.17",
  pages = "163--168"
}

@misc{shao2024deepseekmathpushinglimitsmathematical,
      title={DeepSeekMath: Pushing the Limits of Mathematical Reasoning in Open Language Models}, 
      author={Zhihong Shao and Peiyi Wang and Qihao Zhu and Runxin Xu and Junxiao Song and Xiao Bi and Haowei Zhang and Mingchuan Zhang and Y. K. Li and Y. Wu and Daya Guo},
      year={2024},
      eprint={2402.03300},
      archivePrefix={arXiv},
      primaryClass={cs.CL},
      url={https://arxiv.org/abs/2402.03300}, 
}

@misc{chen2025grouppolicygradient,
      title={Group Policy Gradient}, 
      author={Junhua Chen and Zixi Zhang and Hantao Zhong and Rika Antonova},
      year={2025},
      eprint={2510.03679},
      archivePrefix={arXiv},
      primaryClass={cs.LG},
      url={https://arxiv.org/abs/2510.03679}, 
}

@article{delgiudice2021effectivedimensionality,
  title        = {Effective Dimensionality: A Tutorial},
  author       = {Del Giudice, Marco},
  journal      = {Multivariate Behavioral Research},
  year         = {2021},
  volume       = {56},
  number       = {3},
  pages        = {527--542},
  doi          = {10.1080/00273171.2020.1743631},
  url          = {https://doi.org/10.1080/00273171.2020.1743631}
}

@InProceedings{pmlr-v238-giaffar24a,
  title        = {The Effective Number of Shared Dimensions Between Paired Datasets},
  author       = {Giaffar, Hamza and Rull\'{a}n Bux\'{o}, Camille and Aoi, Mikio},
  booktitle    = {Proceedings of The 27th International Conference on Artificial Intelligence and Statistics},
  pages        = {4249--4257},
  year         = {2024},
  editor       = {Dasgupta, Sanjoy and Mandt, Stephan and Li, Yingzhen},
  volume       = {238},
  series       = {Proceedings of Machine Learning Research},
  month        = {02--04 May},
  publisher    = {PMLR},
  url          = {https://proceedings.mlr.press/v238/giaffar24a.html}
}

@misc{yu2020gradientsurgerymultitasklearning,
      title={Gradient Surgery for Multi-Task Learning}, 
      author={Tianhe Yu and Saurabh Kumar and Abhishek Gupta and Sergey Levine and Karol Hausman and Chelsea Finn},
      year={2020},
      eprint={2001.06782},
      archivePrefix={arXiv},
      primaryClass={cs.LG},
      url={https://arxiv.org/abs/2001.06782}, 
}

@inproceedings{badge,
  title     = {Deep Batch Active Learning by Diverse, Uncertain Gradient Lower Bounds},
  author    = {Ash, Jordan T. and Zhang, Chicheng and Krishnamurthy, Akshay and Langford, John and Agarwal, Alekh},
  booktitle = {International Conference on Learning Representations (ICLR)},
  year      = {2020},
  url       = {https://openreview.net/forum?id=ryghZJBKPS}
}

@inproceedings{tracin,
  title     = {Estimating Training Data Influence by Tracing Gradient Descent},
  author    = {Pruthi, Garima and Liu, Frederick and Kale, Satyen and Sundararajan, Mukund},
  booktitle = {Advances in Neural Information Processing Systems (NeurIPS)},
  year      = {2020},
  url       = {https://arxiv.org/abs/2002.08484}
}

@inproceedings{trak,
  title     = {{TRAK}: Attributing Model Behavior at Scale},
  author    = {Park, Sung Min and Georgiev, Kristian and Ilyas, Andrew and Leclerc, Guillaume and Madry, Aleksander},
  booktitle = {Proceedings of the 40th International Conference on Machine Learning (ICML)},
  year      = {2023},
  series    = {Proceedings of Machine Learning Research},
  volume    = {202},
  url       = {https://proceedings.mlr.press/v202/park23c.html}
}

@misc{ouyang2022training,
  title         = {Training language models to follow instructions with human feedback},
  author        = {Ouyang, Long and Wu, Jeff and Jiang, Xu and Almeida, Diogo and Wainwright, Carroll L. and Mishkin, Pamela and Zhang, Chong and Agarwal, Sandhini and Slama, Katarina and Ray, Alex and Schulman, John and Hilton, Jacob and Kelton, Fraser and Miller, Luke and Simens, Maddie and Askell, Amanda and Welinder, Peter and Christiano, Paul and Leike, Jan and Lowe, Ryan},
  year          = {2022},
  eprint        = {2203.02155},
  archivePrefix = {arXiv},
  primaryClass  = {cs.CL},
  url           = {https://arxiv.org/abs/2203.02155}
}

@misc{schulman_kl_approx,
  author       = {Schulman, John},
  title        = {Approximating KL Divergence},
  howpublished = {Blog post},
  year         = {2016},
  url          = {http://joschu.net/blog/kl-approx.html},
  note         = {Accessed 2026-01-16. Some secondary sources cite the post as 2020.}
}

@misc{schulman2016gae,
  title         = {High-Dimensional Continuous Control Using Generalized Advantage Estimation},
  author        = {Schulman, John and Moritz, Philipp and Levine, Sergey and Jordan, Michael I. and Abbeel, Pieter},
  year          = {2016},
  eprint        = {1506.02438},
  archivePrefix = {arXiv},
  primaryClass  = {cs.LG},
  url           = {https://arxiv.org/abs/1506.02438}
}

@misc{rein2023gpqa,
  title         = {GPQA: A Graduate-Level Google-Proof Q\&A Benchmark},
  author        = {Rein, David and Hou, Betty Li and Stickland, Asa Cooper and Petty, Jackson and Pang, Richard Yuanzhe and Dirani, Julien and Michael, Julian and Bowman, Samuel R.},
  year          = {2023},
  eprint        = {2311.12022},
  archivePrefix = {arXiv},
  primaryClass  = {cs.CL}
}

@misc{huggingfaceh4_aime2024,
  title        = {HuggingFaceH4/aime\_2024},
  author       = {{Hugging Face H4}},
  year         = {2024},
  howpublished = {\url{https://huggingface.co/datasets/HuggingFaceH4/aime_2024}},
  note         = {Accessed: 2026-01-16}
}

@misc{opencompass_aime2025,
  title        = {opencompass/AIME2025},
  author       = {{OpenCompass}},
  year         = {2025},
  howpublished = {\url{https://huggingface.co/datasets/opencompass/AIME2025}},
  note         = {Accessed: 2026-01-16}
}

@article{hendrycksmath2021,
  title={Measuring Mathematical Problem Solving With the MATH Dataset},
  author={Dan Hendrycks and Collin Burns and Saurav Kadavath and Akul Arora and Steven Basart and Eric Tang and Dawn Song and Jacob Steinhardt},
  journal={NeurIPS},
  year={2021}
}

@misc{cobbe2021gsm8k,
  title         = {Training Verifiers to Solve Math Word Problems},
  author        = {Cobbe, Karl and Kosaraju, Vineet and Bavarian, Mohammad and Chen, Mark and Jun, Heewoo and Kaiser, Lukasz and Plappert, Matthias and Tworek, Jerry and Hilton, Jacob and Nakano, Reiichiro and Hesse, Christopher and Schulman, John},
  year          = {2021},
  eprint        = {2110.14168},
  archivePrefix = {arXiv},
  primaryClass  = {cs.LG},
  note          = {Introduces GSM8K}
}

@article{lightman2023lets,
      title={Let's Verify Step by Step}, 
      author={Lightman, Hunter and Kosaraju, Vineet and Burda, Yura and Edwards, Harri and Baker, Bowen and Lee, Teddy and Leike, Jan and Schulman, John and Sutskever, Ilya and Cobbe, Karl},
      journal={arXiv preprint arXiv:2305.20050},
      year={2023}
}

@misc{deepseekai2025deepseekr1incentivizingreasoningcapability,
      title={DeepSeek-R1: Incentivizing Reasoning Capability in LLMs via Reinforcement Learning}, 
      author={DeepSeek-AI},
      year={2025},
      eprint={2501.12948},
      archivePrefix={arXiv},
      primaryClass={cs.CL},
      url={https://arxiv.org/abs/2501.12948}, 
}

@article{sheng2024hybridflow,
  title   = {HybridFlow: A Flexible and Efficient RLHF Framework},
  author  = {Guangming Sheng and Chi Zhang and Zilingfeng Ye and Xibin Wu and Wang Zhang and Ru Zhang and Yanghua Peng and Haibin Lin and Chuan Wu},
  year    = {2024},
  journal = {arXiv preprint arXiv: 2409.19256}
}

@misc{kydlicek2026mathverify,
  author       = {Kydl{\'i}{\v{c}}ek, Hynek and Hugging Face},
  title        = {Math-Verify},
  year         = {2026},
  url          = {https://github.com/huggingface/Math-Verify},
  note         = {GitHub repository. Accessed 2026-01-16.}
}

@misc{xu2025rolloutsusefuldownsamplingrollouts,
      title={Not All Rollouts are Useful: Down-Sampling Rollouts in LLM Reinforcement Learning}, 
      author={Yixuan Even Xu and Yash Savani and Fei Fang and J. Zico Kolter},
      year={2025},
      eprint={2504.13818},
      archivePrefix={arXiv},
      primaryClass={cs.LG},
      url={https://arxiv.org/abs/2504.13818}, 
}

@misc{hou2024doesrlhfscaleexploring,
      title={Does RLHF Scale? Exploring the Impacts From Data, Model, and Method}, 
      author={Zhenyu Hou and Pengfan Du and Yilin Niu and Zhengxiao Du and Aohan Zeng and Xiao Liu and Minlie Huang and Hongning Wang and Jie Tang and Yuxiao Dong},
      year={2024},
      eprint={2412.06000},
      archivePrefix={arXiv},
      primaryClass={cs.CL},
      url={https://arxiv.org/abs/2412.06000}, 
}

@misc{dai2025cdecuriositydrivenexplorationefficient,
      title={CDE: Curiosity-Driven Exploration for Efficient Reinforcement Learning in Large Language Models}, 
      author={Runpeng Dai and Linfeng Song and Haolin Liu and Zhenwen Liang and Dian Yu and Haitao Mi and Zhaopeng Tu and Rui Liu and Tong Zheng and Hongtu Zhu and Dong Yu},
      year={2025},
      eprint={2509.09675},
      archivePrefix={arXiv},
      primaryClass={cs.CL},
      url={https://arxiv.org/abs/2509.09675}, 
}

@misc{wang2023reverseklgeneralizingdirect,
      title={Beyond Reverse KL: Generalizing Direct Preference Optimization with Diverse Divergence Constraints}, 
      author={Chaoqi Wang and Yibo Jiang and Chenghao Yang and Han Liu and Yuxin Chen},
      year={2023},
      eprint={2309.16240},
      archivePrefix={arXiv},
      primaryClass={cs.LG},
      url={https://arxiv.org/abs/2309.16240}, 
}

@misc{haarnoja2017reinforcementlearningdeepenergybased,
      title={Reinforcement Learning with Deep Energy-Based Policies}, 
      author={Tuomas Haarnoja and Haoran Tang and Pieter Abbeel and Sergey Levine},
      year={2017},
      eprint={1702.08165},
      archivePrefix={arXiv},
      primaryClass={cs.LG},
      url={https://arxiv.org/abs/1702.08165}, 
}

@misc{cui2025entropymechanismreinforcementlearning,
      title={The Entropy Mechanism of Reinforcement Learning for Reasoning Language Models}, 
      author={Ganqu Cui and Yuchen Zhang and Jiacheng Chen and Lifan Yuan and Zhi Wang and Yuxin Zuo and Haozhan Li and Yuchen Fan and Huayu Chen and Weize Chen and Zhiyuan Liu and Hao Peng and Lei Bai and Wanli Ouyang and Yu Cheng and Bowen Zhou and Ning Ding},
      year={2025},
      eprint={2505.22617},
      archivePrefix={arXiv},
      primaryClass={cs.LG},
      url={https://arxiv.org/abs/2505.22617}, 
}

@misc{shen2025entropycontrolllmrlalgorithms,
      title={On Entropy Control in LLM-RL Algorithms}, 
      author={Han Shen},
      year={2025},
      eprint={2509.03493},
      archivePrefix={arXiv},
      primaryClass={cs.LG},
      url={https://arxiv.org/abs/2509.03493}, 
}

@misc{mnih2016asynchronousmethodsdeepreinforcement,
      title={Asynchronous Methods for Deep Reinforcement Learning}, 
      author={Volodymyr Mnih and Adrià Puigdomènech Badia and Mehdi Mirza and Alex Graves and Timothy P. Lillicrap and Tim Harley and David Silver and Koray Kavukcuoglu},
      year={2016},
      eprint={1602.01783},
      archivePrefix={arXiv},
      primaryClass={cs.LG},
      url={https://arxiv.org/abs/1602.01783}, 
}

@misc{llama3modelcard,
    title={Llama 3 Model Card},
    author={AI@Meta},
    year={2024},
    url = {https://github.com/meta-llama/llama3/blob/main/MODEL_CARD.md}
}

@misc{qwen3technicalreport,
      title={Qwen3 Technical Report}, 
      author={Qwen Team},
      year={2025},
      eprint={2505.09388},
      archivePrefix={arXiv},
      primaryClass={cs.CL},
      url={https://arxiv.org/abs/2505.09388}, 
}

@article{yang2024qwen25mathtechnicalreportmathematical,
  title={Qwen2.5-Math Technical Report: Toward Mathematical Expert Model via Self-Improvement}, 
  author={An Yang and Beichen Zhang and Binyuan Hui and Bofei Gao and Bowen Yu and Chengpeng Li and Dayiheng Liu and Jianhong Tu and Jingren Zhou and Junyang Lin and Keming Lu and Mingfeng Xue and Runji Lin and Tianyu Liu and Xingzhang Ren and Zhenru Zhang},
  journal={arXiv preprint arXiv:2409.12122
        
        
        
        },
  year={2024}
}

@article{mbpp,
  title={Program Synthesis with Large Language Models},
  author={Austin, Jacob and Odena, Augustus and Nye, Maxwell and Bosma, Maarten and Michalewski, Henryk and Dohan, David and Jiang, Ellen and Cai, Carrie and Terry, Michael and Le, Quoc and others},
  journal={arXiv preprint arXiv:2108.07732
        
        
        
        },
  year={2021}
}

@misc{chen2021evaluating,
      title={Evaluating Large Language Models Trained on Code},
      author={Mark Chen and Jerry Tworek and Heewoo Jun and Qiming Yuan and Henrique Ponde de Oliveira Pinto and Jared Kaplan and Harri Edwards and Yuri Burda and Nicholas Joseph and Greg Brockman and Alex Ray and Raul Puri and Gretchen Krueger and Michael Petrov and Heidy Khlaaf and Girish Sastry and Pamela Mishkin and Brooke Chan and Scott Gray and Nick Ryder and Mikhail Pavlov and Alethea Power and Lukasz Kaiser and Mohammad Bavarian and Clemens Winter and Philippe Tillet and Felipe Petroski Such and Dave Cummings and Matthias Plappert and Fotios Chantzis and Elizabeth Barnes and Ariel Herbert-Voss and William Hebgen Guss and Alex Nichol and Alex Paino and Nikolas Tezak and Jie Tang and Igor Babuschkin and Suchir Balaji and Shantanu Jain and William Saunders and Christopher Hesse and Andrew N. Carr and Jan Leike and Josh Achiam and Vedant Misra and Evan Morikawa and Alec Radford and Matthew Knight and Miles Brundage and Mira Murati and Katie Mayer and Peter Welinder and Bob McGrew and Dario Amodei and Sam McCandlish and Ilya Sutskever and Wojciech Zaremba},
      year={2021},
      eprint={2107.03374},
      archivePrefix={arXiv},
      primaryClass={cs.LG}
}

@inproceedings{evalplus,
  title = {Is Your Code Generated by Chat{GPT} Really Correct? Rigorous Evaluation of Large Language Models for Code Generation},
  author = {Liu, Jiawei and Xia, Chunqiu Steven and Wang, Yuyao and Zhang, Lingming},
  booktitle = {Thirty-seventh Conference on Neural Information Processing Systems},
  year = {2023},
  url = {https://openreview.net/forum?id=1qvx610Cu7},
}

@inproceedings{jain2025livecodebench,
  title     = {{LiveCodeBench}: Holistic and Contamination Free Evaluation of Large Language Models for Code},
  author    = {Jain, Naman and Han, King and Gu, Alex and Li, Wen-Ding and Yan, Fanjia and Zhang, Tianjun and Wang, Sida and Solar-Lezama, Armando and Sen, Koushik and Stoica, Ion},
  booktitle = {International Conference on Learning Representations},
  year      = {2025},
  url       = {https://openreview.net/forum?id=chfJJYC3iL}
}

@article{lin2025zebralogic,
  title   = {{ZebraLogic}: On the Scaling Limits of LLMs for Logical Reasoning},
  author  = {Lin, Bill Yuchen and Le Bras, Ronan and Richardson, Kyle and Sabharwal, Ashish and Poovendran, Radha and Clark, Peter and Choi, Yejin},
  journal = {arXiv preprint arXiv:2502.01100},
  year    = {2025},
  doi     = {10.48550/arXiv.2502.01100},
  url     = {https://arxiv.org/abs/2502.01100}
}

@article{xie2024knightsknaves,
  title   = {On Memorization of Large Language Models in Logical Reasoning},
  author  = {Xie, Chulin and Huang, Yangsibo and Zhang, Chiyuan and Yu, Da and Chen, Xinyun and Lin, Bill Yuchen and Li, Bo and Ghazi, Badih and Kumar, Ravi},
  journal = {arXiv preprint arXiv:2410.23123},
  year    = {2024},
  doi     = {10.48550/arXiv.2410.23123},
  url     = {https://arxiv.org/abs/2410.23123}
}

\newpage
\appendix
\section*{Appendix}
\section{Proof of the residual-gradient norm bound}
\label{app:residual_bound}

This appendix provides a detailed derivation of the norm bound (Eq.~\eqref{equation:bound}) used in
Section~\ref{sec:theory_redundancy}. The argument is deterministic, which holds for each prompt group
without invoking expectations, and relies only on standard Euclidean inequalities.

\paragraph{Setup.}
Fix a prompt $q$. GRPO-style training samples $G$ rollouts $\{o_{q,i}\}_{i=1}^G$ and forms the
group-averaged score-gradient update
\begin{equation}
g_q(\theta) \;\triangleq\; \frac{1}{G}\sum_{i=1}^G \hat A_{q,i}\, s_{q,i},\ \
s_{q,i} \triangleq \nabla_\theta \log \pi_\theta(o_{q,i}\mid q).
\end{equation}
In standard GRPO-style implementations, the advantage is centered within each prompt group~\citep{shao2024deepseekmathpushinglimitsmathematical}, hence
\begin{equation}
\sum_{i=1}^G \hat A_{q,i} \;=\; 0.
\label{eq:group_centering}
\end{equation}
Define the within-group mean score gradient and residuals:
\begin{equation}
\bar s_q \;\triangleq\; \frac{1}{G}\sum_{i=1}^G s_{q,i},
\qquad
\delta_{q,i} \;\triangleq\; s_{q,i}-\bar s_q.
\label{eq:def_residual}
\end{equation}
By construction, the residuals sum to zero:
\begin{equation}
\sum_{i=1}^G \delta_{q,i}
=
\sum_{i=1}^G s_{q,i} - G\bar s_q
=
G\bar s_q - G\bar s_q
=
0.
\label{eq:residuals_sum_zero}
\end{equation}

\paragraph{Residual-only form of the group update.}
Substituting $s_{q,i}=\bar s_q+\delta_{q,i}$ into $g_q(\theta)$ yields
\begin{align}
g_q(\theta)
&=
\frac{1}{G}\sum_{i=1}^G \hat A_{q,i}(\bar s_q+\delta_{q,i})
\nonumber\\
&=
\frac{1}{G}\Big(\sum_{i=1}^G \hat A_{q,i}\Big)\bar s_q
+
\frac{1}{G}\sum_{i=1}^G \hat A_{q,i}\delta_{q,i}.
\label{eq:gq_decompose}
\end{align}
Using the group-centering property (Eq.~\eqref{eq:group_centering}), the mean-component term vanishes,
giving the residual-only identity
\begin{equation}
g_q(\theta)
=
\frac{1}{G}\sum_{i=1}^G \hat A_{q,i}\delta_{q,i}.
\label{eq:gq_residual_only}
\end{equation}

\paragraph{Norm bound via Cauchy--Schwarz.}
We now prove the bound (Eq.~\eqref{equation:bound})
\begin{equation}
\|g_q(\theta)\|_2
\le
\frac{1}{G}
\Big(\sum_{i=1}^G \hat A_{q,i}^2\Big)^{1/2}
\Big(\sum_{i=1}^G \|\delta_{q,i}\|_2^2\Big)^{1/2}.
\label{eq:main_bound}
\end{equation}

\begin{proof}
    Define
\begin{equation}
\label{eq:def_u}
u \;\triangleq\; \sum_{i=1}^G \hat A_{q,i}\delta_{q,i}\in\mathbb{R}^d,
\end{equation}
so that
\begin{equation}
g_q(\theta)=\frac{1}{G}u.
\end{equation}
Hence $\|g_q(\theta)\|_2=\frac{1}{G}\|u\|_2$. We bound $\|u\|_2$ by duality. For any vector $v\in\mathbb{R}^d$
with $\|v\|_2=1$,
\begin{align}
v^\top u
&=
v^\top\sum_{i=1}^G \hat A_{q,i}\delta_{q,i}
=
\sum_{i=1}^G \hat A_{q,i}\, v^\top \delta_{q,i}.
\label{eq:vt_u_expand}
\end{align}
Applying the scalar Cauchy--Schwarz inequality to the sequences
$\{\hat A_{q,i}\}_{i=1}^G$ and $\{v^\top \delta_{q,i}\}_{i=1}^G$ gives
\begin{equation}
|v^\top u|
\le
\Big(\sum_{i=1}^G \hat A_{q,i}^2\Big)^{1/2}
\Big(\sum_{i=1}^G (v^\top \delta_{q,i})^2\Big)^{1/2}.
\label{eq:cs_scalar}
\end{equation}
Next, since $\|v\|_2=1$, we have for each $i$,
\begin{equation}
(v^\top \delta_{q,i})^2
\le
\|v\|_2^2 \,\|\delta_{q,i}\|_2^2
=
\|\delta_{q,i}\|_2^2,
\label{eq:proj_bound}
\end{equation}
which implies
\begin{equation}
\sum_{i=1}^G (v^\top \delta_{q,i})^2
\le
\sum_{i=1}^G \|\delta_{q,i}\|_2^2.
\label{eq:proj_bound_sum}
\end{equation}
Combining Eq.~\eqref{eq:cs_scalar} and~\eqref{eq:proj_bound_sum} yields, for all $\|v\|_2=1$,
\begin{equation}
|v^\top u|
\le
\Big(\sum_{i=1}^G \hat A_{q,i}^2\Big)^{1/2}
\Big(\sum_{i=1}^G \|\delta_{q,i}\|_2^2\Big)^{1/2}.
\label{eq:vt_u_bound}
\end{equation}
Finally, using the dual characterization of the Euclidean norm,
$\|u\|_2 = \sup_{\|v\|_2=1} v^\top u$ (equivalently, $\sup_{\|v\|_2=1}|v^\top u|$), we obtain
\begin{equation}
\|u\|_2
\le
\Big(\sum_{i=1}^G \hat A_{q,i}^2\Big)^{1/2}
\Big(\sum_{i=1}^G \|\delta_{q,i}\|_2^2\Big)^{1/2}.
\label{eq:u_bound}
\end{equation}
Multiplying by $1/G$ and recalling Eq.~\eqref{eq:def_u} proves the bound (Eq.~\eqref{eq:main_bound} and~\eqref{equation:bound}). 
\end{proof}

\section{Residual-gradient diagnostic}
\label{app:residual_diagnostic}

Section~\ref{sec:theory_redundancy} gives a within-prompt mean--residual analysis of why group-centered updates can be weak. We validate this diagnostic mechanism across rollout group sizes. This diagnostic is distinct from \textit{SALT}'s mini-batch dominant/spectral-tail decomposition, which is defined in Section~\ref{sec:pr-neff-reweight}.

\paragraph{Metrics.} For each prompt group, let $s_{q,i}$ denote the per-rollout score-gradient feature and decompose it as
$$
s_{q,i} = \bar{s}_q + \delta_{q,i},
\qquad
\bar{s}_q = \frac{1}{G}\sum_{i=1}^{G}s_{q,i}.
$$
We compute the total, shared, and residual gradient energies as
$$
E_{\mathrm{tot}}(q)=\sum_{i=1}^{G}\|s_{q,i}\|_2^2, \quad E_{\mathrm{sh}}(q)=G\|\bar{s}_q\|_2^2, \quad E_{\delta}(q)=\sum_{i=1}^{G}\|s_{q,i}-\bar{s}_q\|_2^2.
$$
The residual-energy ratio is then defined as
$$
\rho_{\delta}(q)=\frac{E_{\delta}(q)}{E_{\mathrm{tot}}(q)}.
$$
We also measure the realized group-update norm
$$
\|g_q\|_2 =
\left\|
\frac{1}{G}\sum_{i=1}^{G}\hat{A}_{q,i}s_{q,i}
\right\|_2
$$
and compare it with the residual-energy bound
$$
B_{\delta}(q)=
\left(
\frac{1}{G}\sum_{i=1}^{G}\|\delta_{q,i}\|_2^2
\right)^{1/2}.
$$

In practice, we compute these diagnostic quantities from the same LM-head gradient-proxy features used by \textit{SALT}, avoiding the need to materialize full-model per-sample gradients. They explain the group-centered cancellation mechanism but are not the projectors used by \textit{SALT}. Equivalently, given the group-wise Gram matrix $K_q=S_q^\top S_q$, we compute

$$
E_{\delta}(q)
=
\mathrm{tr}(K_q)
-
\frac{1}{G}\mathbf{1}^{\top}K_q\mathbf{1}, \quad
\|g_q\|_2^2
=
\frac{1}{G^2}
\hat{A}_q^\top K_q \hat{A}_q.
$$
\begin{figure}[t]
    \centering
    \includegraphics[width=0.9\linewidth]{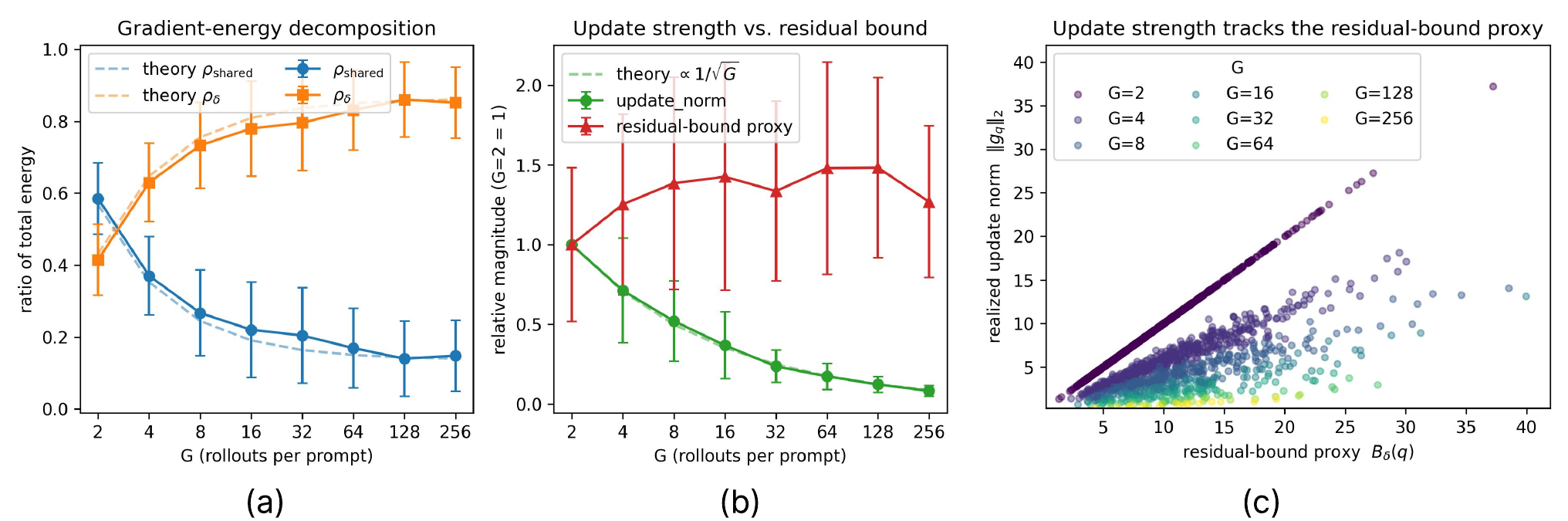}     
\caption{
\textbf{Residual-gradient bottleneck under rollout scaling.}
(a) As the rollout group size $G$ increases, the shared-energy ratio decreases while the residual-energy ratio increases and saturates.
(b) The residual-bound proxy remains bounded, whereas the realized update norm decreases with the group-averaging scale.
(c) Across prompt groups, the realized update norm follows the residual-bound proxy, supporting that effective GRPO-style rollout scaling depends on non-redundant residual directions after group centering rather than on rollout count alone.
}
    \label{fig:residual_diagnostic}
\end{figure}

\paragraph{Validation Results.} Figure~\ref{fig:residual_diagnostic} validates the residual-bottleneck prediction. As $G$ increases, the residual-energy ratio increases but saturates, while the residual-bound proxy remains bounded and the realized update norm decreases. Moreover, the update norm correlates with the residual-bound proxy both globally, with Pearson $0.56$ and Spearman $0.52$, and within each fixed $G$, where correlations remain above $0.95$. Thus, larger rollout groups strengthen GRPO-style training only when they introduce non-redundant residual directions that survive group-centered aggregation.

\section{Implementation Details}
\label{app:details}

In this section, we will provide more details on the implementation of our experiments in Section~\ref{section:setup}

\subsection{Hyperparameter Configuration}
For reproducibility, we implement our method in the VeRL~\citep{sheng2024hybridflow} framework as a modular component, enabling plug-and-play integration with various GRPO-style algorithms.
We conduct experiments using DeepSeek-Distill-Qwen models at two scales (1.5B and 7B) and all experiments are conducted on a single server equipped with 8x NVIDIA H20 GPUs.
Table~\ref{tab:hyperparameter} summarizes the hyperparameter settings used in our experiments (including GRPO and DAPO).
Additionally, in Section~\ref{sec:main_result} we compare methods proposed by \citet{cui2025entropymechanismreinforcementlearning} for increasing policy entropy. By constraining updates to tokens with high covariance, these methods prevent entropy collapse (the specific formulas are provided in Appendix~\ref{appendix:formulations}).
\renewcommand{\arraystretch}{1.05}
    \definecolor{lightgray}{gray}{0.9} 

\begin{table}[!htb]
    \centering
    \caption{The hyperparameter configuration for training.}
    \small  
    \begin{tabular} {lc}
    \toprule
Hyperparameter              & Value (default setting) \\
\midrule
\rowcolor{lightgray} \textit{Training Setting} & \\
Rollout samples per prompt & 8     \\ Rollout temperature    & 0.6   \\
Learning rate               & 1e-6  \\ Validation temperature & 0.6   \\
Entropy coefficient         & 0     \\ KL loss coefficient    & 0.001 \\
Prompt length               & 1024  \\ Response length        & 8192  \\
Train batch size            & 64    \\ Mini-Batch Size        & 32   \\
Training epochs & 2 \\ \midrule
\rowcolor{lightgray} \textit{Additional Setting for DAPO} & \\
Generation batch size            & 128   \\
Filter groups enable             & True  \\
Max numbers of generation batches & 10    \\
Clip ratio (Low)                 & 0.2   \\
Clip ratio (High)                & 0.28 \\
\rowcolor{lightgray} \textit{KL-Covariance} & \\
KL covariance ratio & 0.002     \\ PPO KL coefficient    & 1   \\
Clip ratio (Low)                 & 0.2   \\
Clip ratio (High)                & 0.2 \\ 
Training epochs & 2 \\ \midrule
\rowcolor{lightgray} \textit{Clip-Covariance} & \\
Clip    covariance ratio         & 0.0002   \\
Clip    covariance lb             & 1.0  \\
Clip    covariance ub  & 5.0    \\
Clip ratio (Low)                 & 1   \\
Clip ratio (High)                & 1 \\
     \bottomrule
    \end{tabular}
    \label{tab:hyperparameter}
\end{table}

\subsection{System Prompt and Reward}
\paragraph{System prompt.}
For all experiments, we applied a unified system prompt to constrain the model’s output format, requiring explicit step-by-step reasoning and a clearly delineated final answer
\begin{quote}
    \small
    Please reason step by step, and put your final answer within \textbackslash boxed\{\}.
\end{quote}
which is the default prompt in VeRL~\citep{sheng2024hybridflow}.

\paragraph{Reward function.}
To better align the training signal with the reward for solving problems correctly, we use the Math-Verify library~\citep{kydlicek2026mathverify} to extract the final answer (see the reference code for details). For GRPO, the reward is set to $1$ for a correct answer and $0$ otherwise; for DAPO, the reward is $1$ for a correct answer and $-1$ otherwise.
\section{Formal definitions of metrics}
\label{appendix:metrics}

\subsection{Metrics in redundancy observation}
\label{app:metrics:redundancy}
In Section~\ref{section:features}, we introduce the metrics used to measure gradient redundancy. This section presents a detailed description of these metrics, which are divided into \textbf{in} and \textbf{out} categories.

In each iteration, we sample a mini-batch of prompts $\{q_b\}_{b=1}^B$. For each prompt $q_b$, we sample a group of
$G$ responses and obtain a set of gradient features $\{g_{b,i}\}_{i=1}^G \subset \mathbb{R}^d$.
We use $g_{b,i}$ as the geometric object for measuring alignment and effective dimensionality. Throughout, we define
the flattened index set of all sampled responses as $\mathcal{I}=\{(b,i): b=1,\ldots,B,\ i=1,\ldots,G\}$ with
$|\mathcal{I}|=BG$.

To quantify cancellation in the actual update, we additionally define the signed per-sample update contribution
$u_{b,i}\in\mathbb{R}^{|\theta|}$ as the parameter-gradient contribution multiplied by the scalar
coefficient used by the objective. Concretely, we instantiate
\begin{equation}
u_{b,i} \triangleq \hat A_{b,i}\, v_{b,i},
\end{equation}
where $v_{b,i}$ is the response-level parameter-gradient contribution and
$\hat A_{b,i}$ is the normalized advantage.

\paragraph{(1) Positive cosine similarity.}
For any two vectors $x,y\in\mathbb{R}^d$, define the cosine similarity
$\cos(x,y)\triangleq \langle x,y\rangle /(\|x\|_2\|y\|_2)$ and its positive version $[\cos(x,y)]_+\triangleq\max\{0,\cos(x,y)\}$.
For a set of $m$ vectors $\{g_i\}_{i=1}^m$, we define the mean positive cosine similarity~\citep{zhestiankin-ponomareva-2021-zhestyatsky}
\begin{equation}
r(\{g_i\}_{i=1}^m)\triangleq \frac{2}{m(m-1)}\sum_{1\le i<j\le m} [\cos(g_i,g_j)]_+.
\end{equation}
We report within-group and batch-level variants:
\begin{equation}
r_{\mathrm{in}}(q_b) \triangleq r(\{g_{b,i}\}_{i=1}^G),
\end{equation}
\begin{equation}
r_{\mathrm{in,mean}} \triangleq \frac{1}{B}\sum_{b=1}^B r_{\mathrm{in}}(q_b),
\end{equation}
\begin{equation}
r_{\mathrm{total}} \triangleq r(\{g_{b,i}\}_{(b,i)\in\mathcal{I}}).
\end{equation}

\paragraph{(2) Participation ratio (PR).}
Given a set of $m$ gradient features $\{g_i\}_{i=1}^m$, let $G=[g_1,\ldots,g_m]\in\mathbb{R}^{d\times m}$ and form the
normalized Gram matrix
\begin{equation}
K \triangleq \frac{1}{m}G^\top G \in \mathbb{R}^{m\times m}.
\end{equation}
Let $\{\lambda_i\}_{i=1}^m$ denote the eigenvalues of $K$ (computed by a symmetric eigensolver); in implementation we
clamp $\lambda_i\leftarrow \max\{0,\lambda_i\}$ for numerical stability. We define the participation ratio~\citep{Recanatesi2022ScaleDependent,Hazelden2025FastNTK}
\begin{equation}
\mathrm{PR}(K)\triangleq \frac{\left(\sum_{i=1}^m \lambda_i\right)^2}{\sum_{i=1}^m \lambda_i^2 }.
\end{equation}
Larger $\mathrm{PR}(K)$ indicates a less concentrated spectrum and thus a higher effective dimensionality of the set
$\{g_i\}_{i=1}^m$.

We report within-group and batch-level variants:
\begin{equation}
\mathrm{PR}_{\mathrm{in}}(q_b)\triangleq \mathrm{PR}(K_{\mathrm{in}}(q_b)),
\end{equation}
\begin{equation}
K_{\mathrm{in}}(q_b) \triangleq \frac{1}{G}G_b^\top G_b,\ \ G_b\triangleq[g_{b,1},\ldots,g_{b,G}],
\end{equation}
\begin{equation}
\mathrm{PR}_{\mathrm{in,mean}} \triangleq \frac{1}{B}\sum_{b=1}^B \mathrm{PR}_{\mathrm{in}}(q_b),
\end{equation}
\begin{equation}
\mathrm{PR}_{\mathrm{total}} \triangleq \mathrm{PR}(K_{\mathrm{total}}),
\label{eq:pr_total_2}
\end{equation}
\begin{equation}
K_{\mathrm{total}}\triangleq \frac{1}{BG}G_{\mathrm{all}}^\top G_{\mathrm{all}},
\end{equation}
where $G_{\mathrm{all}}\triangleq [g_{b,i}]_{(b,i)\in\mathcal{I}}$ stacks all gradient features in the mini-batch.

\paragraph{(3) Effective sample size proxy.}
For a set of signed update vectors $\{u_i\}_{i=1}^m$, we define the effective sample size proxy
\begin{equation}
n_{\mathrm{eff}}(\{u_i\}_{i=1}^m)\triangleq \frac{\left\|\sum_{i=1}^m u_i\right\|_2^2}{\sum_{i=1}^m \|u_i\|_2^2}.
\end{equation}
This ratio increases when updates add coherently and decreases under cancellation~\citep{Kish1965SurveySampling,Yang2006EffectiveSampleSize}.

We report within- and cross-group variants:
\begin{equation}
n_{\mathrm{eff}}^{\mathrm{in}}(q_b)\triangleq n_{\mathrm{eff}}(\{u_{b,i}\}_{i=1}^G),
\end{equation}
\begin{equation}
n_{\mathrm{eff,in,mean}}\triangleq \frac{1}{B}\sum_{b=1}^B n_{\mathrm{eff}}^{\mathrm{in}}(q_b),
\end{equation}
and for cross-group aggregation we first compute prompt-level mean updates
\begin{equation}
\bar u_b \triangleq \frac{1}{G}\sum_{i=1}^G u_{b,i},
\end{equation}
then define
\begin{equation}
n_{\mathrm{eff}}^{\mathrm{out}} \triangleq n_{\mathrm{eff}}(\{\bar u_b\}_{b=1}^B).
\end{equation}

\subsection{Metrics in evaluation}
In Section~\ref{section:setup}, we evaluate the benchmarks with the following metrics:

\paragraph{Pass@k.}
\textit{Pass@k} is a standard metric for math-generation benchmarks that measures the probability that at least one of \(k\) randomly selected candidate solutions solves the problem.
Concretely, for each instance, we draw \(n\) candidate outputs (typically with stochastic decoding) and mark \(c\) of them as correct according to the evaluator (e.g., unit tests or an exact-match checker).
The estimated pass@k is given by
\begin{equation}
\mathrm{pass@}k
=1-\frac{\binom{n-c}{k}}{\binom{n}{k}},
\qquad 1\le k\le n,
\end{equation}
where $\binom{n-c}{k}=0$ when $n-c<k$. This estimates the probability that a uniformly chosen subset of \(k\) samples contains at least one correct solution. Under the evaluation setting $n=8$, Pass@1 equals $c/8$ and Pass@8 equals $\mathbf{1}\{c>0\}$.
We report \(\mathrm{pass@}k\) averaged over all instances and, to ensure the stability of the experiments, we will run validation five times and report the average results.

\section{Proxy Fidelity and Blockwise Whole-Model Geometry}
\label{app:proxy}

\textit{SALT} relies on the geometry of per-sample policy-gradient
features, including sample-wise similarity, signed alignment,
spectral concentration, and the dominant subspace used for the
dominant/spectral-tail decomposition. Directly materializing full
per-sample gradients over all model parameters is prohibitively
expensive for LLM training. Therefore, in our practical
implementation, we \textbf{use the LM-head/output-projection gradient
as a lightweight proxy} for estimating the geometry required by
\textit{SALT}.

We emphasize that the proxy is not intended to match the
full-model gradient entry-wise. \textbf{\textit{SALT} only uses the proxy through
low-dimensional sample-wise geometry}: the cosine-Gram structure,
the signed alignment pattern relevant to cancellation, the
participation-ratio spectrum, and the top eigenspace used to
construct the dominant and spectral-tail channels. This appendix
therefore evaluates \textbf{whether the LM-head proxy preserves these
\textit{SALT}-relevant geometric quantities against stronger references},
including full-gradient diagnostics on smaller models and
blockwise whole-model sketches on larger models.

\subsection{References and Protocol}
\label{app:proxy_protocol}

For each validation checkpoint, we use the same rollouts,
rewards, and group-normalized advantages as in training. We
compare the LM-head proxy against three types of reference
gradients.
\renewcommand{\arraystretch}{1.05}
\definecolor{lightgray}{gray}{0.9}
\colorlet{lightbrown}{brown!10}
\begin{table}[t]
\centering
\small
\caption{
Proxy fidelity against full-gradient and blockwise references.
Results are averaged over validation checkpoints and mini-batches.
\textbf{The LM-head proxy preserves the geometry used by \textit{SALT} substantially
better than permutation and noise nulls.}
}
\label{tab:proxy_fidelity}
\resizebox{\textwidth}{!}{
\begin{tabular}{lcccccc}
\toprule
\textbf{Baselines}& \textbf{Gram} $\rho$ $\uparrow$
& \textbf{Sign agree.} $\uparrow$
& \textbf{PR rel. err.} $\downarrow$
& \textbf{Subspace overlap} $\uparrow$
& $k_t$ \textbf{match} $\uparrow$
& $a'$ \textbf{cosine} $\uparrow$ \\
\midrule
\rowcolor{lightgray}\multicolumn{7}{l}{\textbf{Reference Gradients Baselines}} \\
\midrule

Full gradient, small model
& $0.71 \pm 0.06$
& $0.82 \pm 0.03$
& $0.09 \pm 0.04$
& $0.78 \pm 0.05$
& $0.74 \pm 0.07$
& $0.86 \pm 0.04$ \\

Blockwise aggregate
& $0.76 \pm 0.05$
& $0.84 \pm 0.03$
& $0.07 \pm 0.03$
& $0.81 \pm 0.04$
& $0.79 \pm 0.06$
& $0.88 \pm 0.03$ \\

Early block
& $0.58 \pm 0.07$
& $0.75 \pm 0.04$
& $0.16 \pm 0.06$
& $0.64 \pm 0.06$
& $0.61 \pm 0.08$
& $0.77 \pm 0.05$ \\

Middle block
& $0.64 \pm 0.06$
& $0.78 \pm 0.04$
& $0.13 \pm 0.05$
& $0.69 \pm 0.05$
& $0.66 \pm 0.07$
& $0.80 \pm 0.05$ \\

Late block
& $0.73 \pm 0.05$
& $0.83 \pm 0.03$
& $0.08 \pm 0.04$
& $0.79 \pm 0.05$
& $0.77 \pm 0.06$
& $0.87 \pm 0.04$ \\

Last block, disjoint
& $0.79 \pm 0.04$
& $0.85 \pm 0.03$
& $0.06 \pm 0.03$
& $0.84 \pm 0.04$
& $0.82 \pm 0.05$
& $0.90 \pm 0.03$ \\

Random-block average
& $0.68 \pm 0.06$
& $0.80 \pm 0.04$
& $0.10 \pm 0.04$
& $0.73 \pm 0.05$
& $0.71 \pm 0.07$
& $0.84 \pm 0.04$ \\

\midrule
\rowcolor{lightgray}\multicolumn{7}{l}{\textbf{Null Baselines}} \\
\midrule

Permutation null
& $0.02 \pm 0.04$
& $0.51 \pm 0.02$
& $0.47 \pm 0.15$
& $0.18 \pm 0.06$
& $0.19 \pm 0.08$
& $0.08 \pm 0.05$ \\

Noise null
& $0.00 \pm 0.03$
& $0.50 \pm 0.02$
& $0.56 \pm 0.18$
& $0.13 \pm 0.05$
& $0.12 \pm 0.06$
& $0.03 \pm 0.04$ \\
\bottomrule
\end{tabular}
}
\end{table}

\paragraph{Small-model full-gradient reference.} First, on smaller models where diagnostic computation is
tractable, we compute full trainable-parameter per-sample
gradients. This directly tests whether the proxy reflects
whole-model gradient geometry. These full gradients are used only
for validation and are never materialized during \textit{SALT} training.

\paragraph{Blockwise whole-model sketch.}
 Second, for larger models, we compute blockwise per-sample
gradients from representative Transformer blocks, including
early, middle, late, and randomly sampled blocks. We also report
an aggregated blockwise reference by combining the sampled-block
Gram statistics. This blockwise sketch tests whether the LM-head
proxy captures geometry beyond a single adjacent layer.

\paragraph{Disjoint last-block reference.}
 Third, we retain the original disjoint last-block validation. The
LM-head and last-block parameters have zero parameter overlap,
so agreement cannot be explained by shared parameters.

\paragraph{Null baselines.}
 We include two null baselines: a permutation null, which randomly
permutes sample correspondence between proxy and reference
gradients, and a noise null, which replaces proxy features with
random vectors of matched dimension.

\subsection{Metrics}
\label{app:proxy_metrics}

Let $p_i$ denote the LM-head proxy for sample $i$, and $g_i$ a
reference gradient. We evaluate only the geometric quantities used
by \textit{SALT}.

\paragraph{Gram correlation.}
We compute cosine-Gram matrices
\begin{equation}
    \widetilde{K}^{p}_{ij}
    =
    \frac{\langle p_i,p_j\rangle}{\|p_i\|_2\|p_j\|_2},
    \quad
    \widetilde{K}^{g}_{ij}
    =
    \frac{\langle g_i,g_j\rangle}{\|g_i\|_2\|g_j\|_2},
\end{equation}
and report Spearman correlation over off-diagonal entries.

\paragraph{Signed alignment.}
Since \textit{SALT} targets signed cancellation, we report sign agreement
between $\widetilde{K}^{p}_{ij}$ and $\widetilde{K}^{g}_{ij}$ on
high-magnitude reference pairs.

\paragraph{Spectral consistency.}
We compare the participation ratio
\begin{equation}
    \mathrm{PR}(K)
    =
    \frac{\left(\sum_r \lambda_r\right)^2}
    {\sum_r \lambda_r^2},
\end{equation}
and report PR error as well as the consistency of the selected
dominant dimension $k_t$.

\paragraph{Subspace consistency.}
Let $V_k^p$ and $V_k^g$ be the top-$k$ eigenspaces of the proxy
and reference Gram matrices. We report principal-angle overlap,
computed from the singular values of $(V_k^p)^\top V_k^g$.

\paragraph{Reweighting consistency.}
To measure whether the proxy leads to similar \textit{SALT} decisions, we
compare the reweighted advantages induced by proxy and reference
geometries:
\begin{equation}
    a_{\mathrm{proxy}}'
    =
    P_k^p a
    +
    \alpha_t^p (I-P_k^p) [a]_+,
\end{equation}
and analogously $a_{\mathrm{ref}}'$. We report cosine similarity
and rank correlation between them.

\subsection{Results}
\label{app:proxy_results}

Table~\ref{tab:proxy_fidelity} summarizes proxy fidelity across
checkpoints and mini-batches. The LM-head proxy preserves the
\textit{SALT}-relevant geometry against both full-gradient diagnostics and
blockwise references. It achieves substantially higher Gram
correlation, signed agreement, PR consistency, subspace overlap,
and reweighting consistency than permutation and noise nulls.

\begin{figure}[t]
    \centering
    \includegraphics[width=0.95\linewidth]{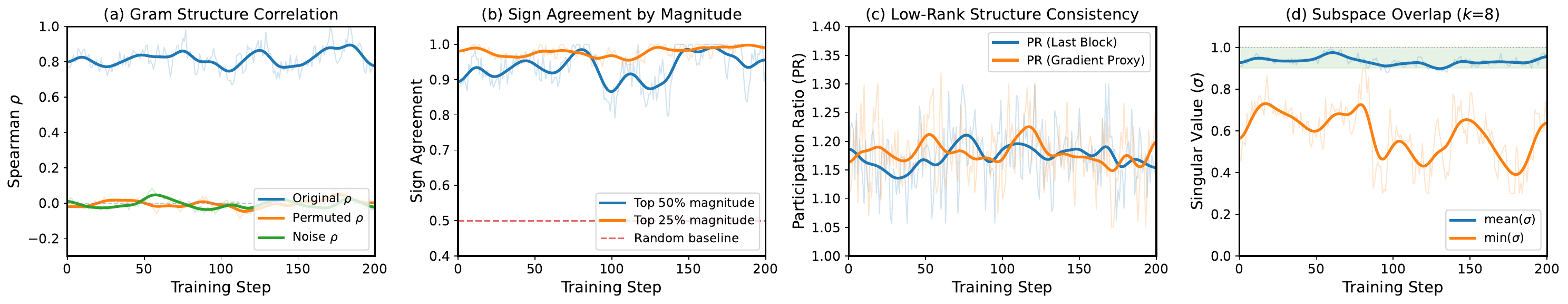}
    \vspace{-0.5em}
    \caption{
\textbf{Disjoint last-block proxy validation.}
We validate the LM-head proxy against per-sample gradients from the last Transformer block on identical rollouts and advantages. Since the two parameter subsets are disjoint, their agreement cannot arise from shared parameters. We compare the geometry used by \textit{SALT}: (a) cosine-Gram correlation is high while permutation/noise nulls are near zero; (b) high-magnitude signed pairs agree far above chance; (c) $PR$ dynamics are closely tracked; and (d) the dominant top-$k$ eigenspace ($k{=}8$) is well preserved, despite less stable tail directions.
}
    \label{fig:proxy}
\end{figure}

Figure~\ref{fig:proxy} shows the original disjoint
last-block validation. Despite zero parameter overlap, the LM-head
proxy tracks the last-block reference in Gram structure, signed
alignment, PR dynamics, and dominant subspace, while the null
baselines remain near chance. Together with the full-gradient and
blockwise results in Table~\ref{tab:proxy_fidelity}, this supports
using the LM-head proxy as a scalable estimator of the geometry
required by \textit{SALT}.

\renewcommand{\arraystretch}{1.05}
\definecolor{lightgray}{gray}{0.9} 
\colorlet{lightbrown}{blue!10}
\colorlet{lightbrown}{brown!10}

\newcommand{\pos}[1]{\textcolor{red}{#1}}

\begin{wraptable}{r}{0.49\textwidth}
    \centering
    \small
    \vspace{-8.2em}
    \caption{Cross-model generalization under matched budgets. Across models using the same RLVR pipeline, GRPO\textit{+SALT} consistently outperforms GRPO.}
    \vspace{1em}
\resizebox{0.48\textwidth}{!}{
\begin{tabular}{llcc}
\toprule
\multirow{2}{*}{\textbf{Model}} &  & \multicolumn{2}{c}{\textbf{Average}}  \\
\cmidrule(lr){3-4}
  &      &    ACC       & Pass@8          \\ \midrule
\rowcolor{lightgray}\multicolumn{4}{c}{\textbf{Deepseek-Distill-Qwen Model}} \\
\midrule

 \multirow{3}{*}{\textbf{Deepseek-Distill-Qwen-1.5B}}
                            & GRPO    & 51.2              & 75.3	 \\
    & \cellcolor{lightbrown}\; + \textit{SALT}    & \cellcolor{lightbrown}53.6&\cellcolor{lightbrown}78.1 \\
    & \cellcolor{lightbrown}\; + $\Delta$    & \cellcolor{lightbrown}\pos{+ 2.4}&\cellcolor{lightbrown}\pos{+ 2.8} \\
 \cmidrule(lr){2-4}
  \multirow{3}{*}{\textbf{Deepseek-Distill-Qwen-7B}}
& GRPO    & 66.9              & 85.8	 \\
    & \cellcolor{lightbrown}\; + \textit{SALT}    & \cellcolor{lightbrown}69.7&\cellcolor{lightbrown}88.1 \\
    & \cellcolor{lightbrown}\; + $\Delta$    & \cellcolor{lightbrown}\pos{+ 2.8}&\cellcolor{lightbrown}\pos{+ 2.3} \\\midrule
\rowcolor{lightgray}\multicolumn{4}{c}{\textbf{Instruct Model}} \\
\midrule
 \multirow{3}{*}{\textbf{LLaMA-3.1-8B}}
                            & GRPO    & 33.5              & 54.8	 \\
    & \cellcolor{lightbrown}\; + \textit{SALT}    & \cellcolor{lightbrown}36.1&\cellcolor{lightbrown}57.0 \\
    & \cellcolor{lightbrown}\; + $\Delta$    & \cellcolor{lightbrown}\pos{+ 2.6} &\cellcolor{lightbrown}\pos{+ 2.2}\\
 \cmidrule(lr){2-4}
  \multirow{3}{*}{\textbf{Qwen-3-8B}}
& GRPO    & 80.1              & 89.0	 \\
    & \cellcolor{lightbrown}\; + \textit{SALT}    & \cellcolor{lightbrown}82.8&\cellcolor{lightbrown}92.3 \\
    & \cellcolor{lightbrown}\; + $\Delta$    & \cellcolor{lightbrown}\pos{+ 2.7}&\cellcolor{lightbrown}\pos{+ 3.3} \\\midrule

\rowcolor{lightgray}\multicolumn{4}{c}{\textbf{Math Model}} \\
\midrule
 \multirow{3}{*}{\textbf{Qwen-2.5-Math-1.5B}}
                            & GRPO    & 43.7              & 62.5  \\

    & \cellcolor{lightbrown}\; + \textit{SALT}    & \cellcolor{lightbrown}46.9&\cellcolor{lightbrown}65.1 \\
    & \cellcolor{lightbrown}\; + $\Delta$    & \cellcolor{lightbrown}\pos{+ 3.2}&\cellcolor{lightbrown}\pos{+ 2.6} \\
 \cmidrule(lr){2-4}
  \multirow{3}{*}{\textbf{Deepseek-Math-7B}}
& GRPO    & 33.2              & 52.2	 \\
    & \cellcolor{lightbrown}\; + \textit{SALT}    & \cellcolor{lightbrown}37.2&\cellcolor{lightbrown}55.7 \\
    & \cellcolor{lightbrown}\; + $\Delta$    & \cellcolor{lightbrown}\pos{+ 4.0}&\cellcolor{lightbrown}\pos{+ 3.5} \\

     \bottomrule
    \end{tabular}
    }
    \vspace{-4em}
    \label{tab:cross}
\end{wraptable}

\section{Details of Experiment}
\subsection{Additional Experiments for RQ1}
\label{app:rq1}
In Table~\ref{tab:main_experiment}, we show that adding \textit{SALT} yields effective task-level gains. Figure~\ref{fig:rq2} further illustrates the geometric characteristics observed during training, demonstrating that \textit{SALT} can indeed bring meaningful improvements. In this section, we will focus on different model families and non-mathematical RLVR tasks to more intuitively show that this phenomenon is broadly prevalent in RLVR.

\paragraph{Cross-Models Generalization.} 
We evaluate whether the signed low-rank structure and the resulting gradient cancellation observed in Figure persist across model families, and whether \textit{SALT} consistently improves both gradient effectiveness ($PR$ and $n_{eff}$) and downstream accuracy. For all models, we keep the RLVR pipeline identical to the main experiments. We match the total sampling budget and the number of optimization steps between GRPO and adding \textit{SALT}~\citep{llama3modelcard,qwen3technicalreport,yang2024qwen25mathtechnicalreportmathematical,shao2024deepseekmathpushinglimitsmathematical}.

\begin{figure*}[!t]
    \centering
    \includegraphics[width=1\linewidth]{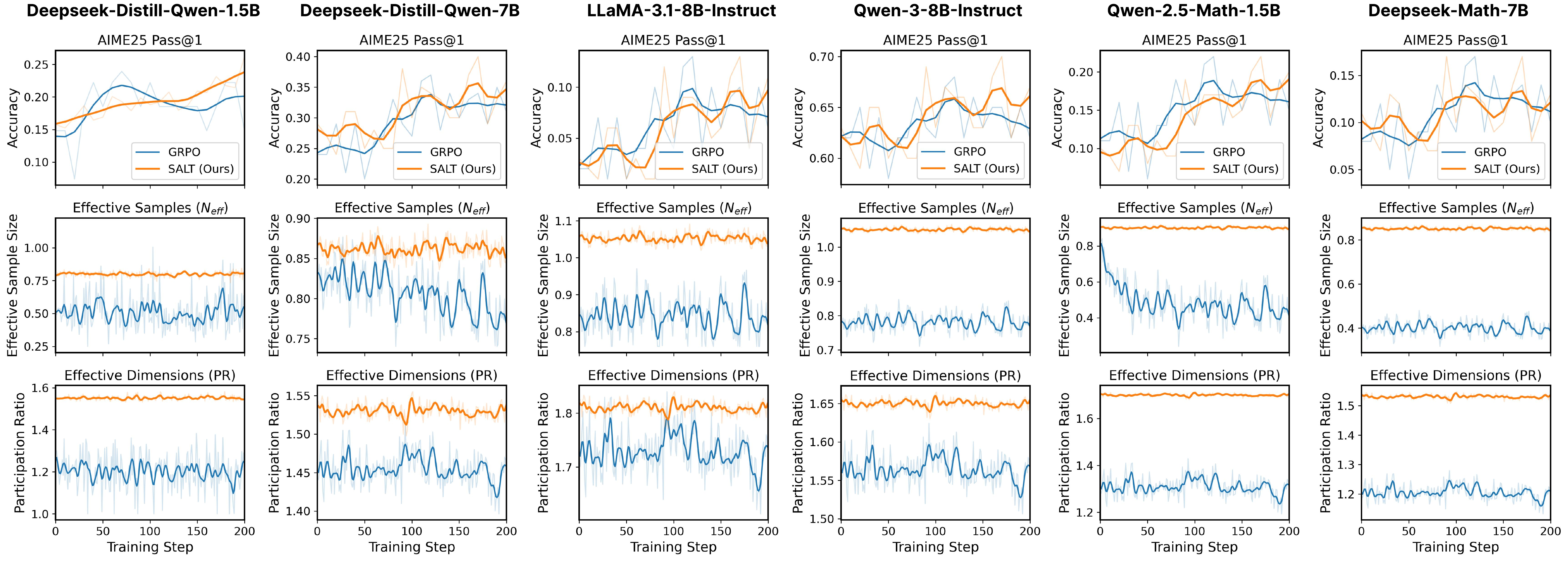}
\caption{\textbf{Signed low-rank geometry persists across model families, and \textit{SALT} mitigates gradient cancellation.}
Training dynamics on AIME25 across multiple model families under matched sampling and optimization budgets.
Top: Pass@1; Middle: effective sample size $n_{\mathrm{eff}}$; Bottom: participation ratio ($PR$).
Despite GRPO exhibiting low PR and low $n_{\mathrm{eff}}$ (strong cancellation), adding \textit{SALT} consistently increases both PR and $n_{\mathrm{eff}}$, translating into improved Pass@1.}

    \label{fig:cross_model}
\end{figure*}

Geometry persists across models. \textbf{We observe consistently low effective rank (small PR) and strong gradient cancellation (low $n_{eff}$)} in Figure~\ref{fig:cross_model}.
\textit{SALT} selectively fixes cancellation. Across models, \textit{SALT} increases both $PR$ and 
$n_{eff}$, indicating that rollouts are converted into effective update directions rather than being canceled by opposing alignments.

As summarized in Table~\ref{tab:cross}, on average across all models, adding \textit{SALT} improves ACC by \textbf{3.0} points and Pass@8 by \textbf{2.8} points compared to GRPO. \textbf{\textit{SALT} consistently improves average accuracy and Pass@8} across distilled base models, instruction-tuned models, and math-specialized models, suggesting that \textit{SALT}’s gains stem from mitigating gradient cancellation rather than merely encouraging randomness.

\begin{figure}[t]
    \centering
    \includegraphics[width=0.98\linewidth]{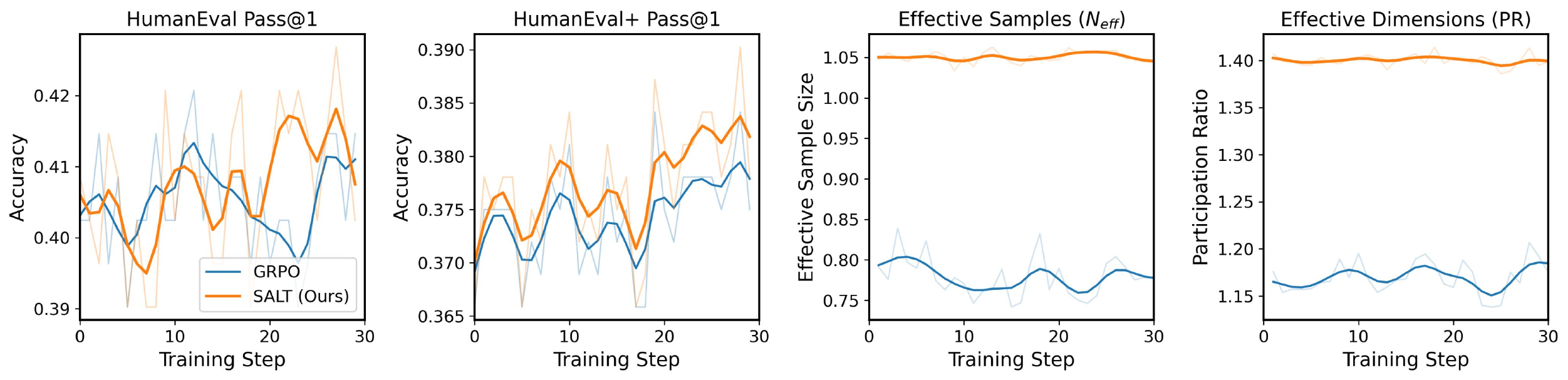}
\caption{\textbf{Code RLVR results on MBPP with DeepSeek-R1-Distill-Qwen-7B.}
GRPO exhibits strong cancellation (lower effective sample size $n_{\mathrm{eff}}$) and an effectively low-rank update geometry (lower $PR$), while \textit{SALT} consistently increases both $n_{\mathrm{eff}}$ and $PR$ with higher Pass@1 performance.}
    \label{fig:code}
\end{figure}

\paragraph{Code RLVR.}
We further evaluate \textit{SALT} on code generation with execution-based verification, where rewards are obtained by running unit tests. This setting serves as an orthogonal, non-math RLVR domain with a high-precision verifier, enabling us to examine whether the signed low-rank geometry and the effectiveness of \textit{SALT} generalize beyond mathematical reasoning. 

To better expose this phenomenon, we train DeepSeek-R1-Distill-Qwen-7B for five epochs on the MBPP training set~\citep{mbpp} with a batch size of 32. As shown in Figure~\ref{fig:code}, signed low-rank geometry persists during code RLVR training: compared with GRPO, \textit{SALT} yields modest but consistent Pass@1 gains on both HumanEval~\citep{chen2021evaluating} and HumanEval+~\citep{evalplus}, while also improving $n_{\mathrm{eff}}$ and PR, indicating less cancellation and more diverse effective updates.

\subsection{Detailed Results for RQ2}
\label{app:rq2}
This appendix provides the complete results and supplementary analyzes for RQ2 in Section~\ref{sec:main_result}.
\renewcommand{\arraystretch}{1.05}
\definecolor{lightgray}{gray}{0.9}
\colorlet{lightbrown}{brown!10}

\begin{table}[!htb]
\centering

\small
\caption{Per-benchmark performance comparison aggregated under GRPO and DAPO training. Results are averaged over $5$ runs. \textbf{Bold} numbers indicate best performance within each block. Overall, compared with entropy-control methods, \textbf{\textit{SALT} not only introduce more effective directions at the geometric level, but also achieve the best task performance.}}
    \resizebox{\textwidth}{!}{
\begin{tabular}{clcccccccccc}
\toprule
\multirow{2}{*}{\textbf{Training}} & \multirow{2}{*}{\textbf{Method}} 
& \multicolumn{2}{c}{\textbf{AIME24}} 
& \multicolumn{2}{c}{\textbf{AIME25}} 
& \multicolumn{2}{c}{\textbf{GSM8K}} 
& \multicolumn{2}{c}{\textbf{MATH-500}} 
& \multicolumn{2}{c}{\textbf{GPQA}} \\
\cmidrule(lr){3-4}\cmidrule(lr){5-6}\cmidrule(lr){7-8}\cmidrule(lr){9-10}\cmidrule(lr){11-12}
 & & ACC & Pass@8 & ACC & Pass@8 & ACC & Pass@8 & ACC & Pass@8 & ACC & Pass@8 \\
\midrule
\rowcolor{lightgray}\multicolumn{12}{c}{\textbf{Deepseek-Distill-Qwen-1.5B}} \\
\midrule

\multirow{10}{*}{\textbf{MATH-T}}
& GRPO
& 29.3 & 59.3 & 25.6 & 43.3 & 80.7 & 94.8 & 85.6 & 96.7 & 35.0 & 82.2 \\
& \; + Entropy Bonus
& 29.0 & 59.1 & 25.4 & 43.2 & 80.5 & 94.7 & 85.4 & 96.6 & 34.8 & 82.1 \\
& \; + Clip Cov
& 29.6 & \underline{60.1} & {25.9} & \underline{44.2} & 81.1 & \underline{95.3} & 86.0 & 97.0 & 35.2 & 82.4 \\
& \; + KL Cov
& \underline{29.8} & 60.0 & \underline{26.0} & 44.0 & \underline{81.3} & 95.1 & \underline{86.2} & \underline{97.1} & \underline{35.5} & \underline{82.6} \\
& \cellcolor{lightbrown}\; + \textit{SALT}
& \cellcolor{lightbrown}\textbf{32.1} & \cellcolor{lightbrown}\textbf{62.9}
& \cellcolor{lightbrown}\textbf{27.1} & \cellcolor{lightbrown}\textbf{47.2}
& \cellcolor{lightbrown}\textbf{83.4} & \cellcolor{lightbrown}\textbf{96.5}
& \cellcolor{lightbrown}\textbf{87.2} & \cellcolor{lightbrown}\textbf{98.4}
& \cellcolor{lightbrown}\textbf{38.4} & \cellcolor{lightbrown}\textbf{85.5} \\
\cmidrule(lr){2-12}
& DAPO
& \textbf{31.1} & 59.9 & 23.9 & 44.3 & 81.0 & 96.1 & 86.1 & 97.0 & 37.0 & 81.7 \\
& \; + Entropy Bonus
& 30.0 & 60.0 & 24.0 & 44.5 & 81.2 & 96.2 & 86.0 & 97.0 & 37.1 & 81.8 \\
& \; + Clip Cov
& {30.3} & 60.1 & {24.3} & 44.6 & 81.4 & 96.4 & {86.2} & \underline{97.1} & 37.3 & \underline{82.0} \\
& \; + KL Cov
& 30.4 & \underline{60.2} & \underline{24.4} & \underline{44.7} & \underline{81.5} & \underline{96.5} & \underline{86.3} & \textbf{97.2} & \underline{37.4} & \underline{82.0}\\
& \cellcolor{lightbrown}\; + \textit{SALT}
& \cellcolor{lightbrown}\underline{30.6} & \cellcolor{lightbrown}\textbf{60.6}
& \cellcolor{lightbrown}\textbf{25.6} & \cellcolor{lightbrown}\textbf{46.3}
& \cellcolor{lightbrown}\textbf{82.5} & \cellcolor{lightbrown}\textbf{97.1}
& \cellcolor{lightbrown}\textbf{87.1} & \cellcolor{lightbrown}97.0
& \cellcolor{lightbrown}\textbf{39.6} & \cellcolor{lightbrown}\textbf{83.5} \\
\midrule

\multirow{10}{*}{\textbf{DAPO-M}}
& GRPO
& 29.1 & 59.0 & 23.6 & 44.2 & 82.2 & 95.3 & 85.9 & 96.8 & 36.2 & 83.7 \\
& \; + Entropy Bonus
& 29.3 & 59.4 & 23.9 & 44.6 & 82.4 & 95.6 & 86.1 & 97.0 & 36.5 & 83.5 \\
& \; + Clip Cov
& 29.9 & 59.2 & 24.5 & 43.8 & {\underline{82.6}} & 95.0 & 86.5 & 96.6 & {36.9} & 83.5 \\
& \; + KL Cov
& \underline{30.0} & \underline{59.8} & \underline{24.6} & \underline{45.0} & \underline{82.6} & \underline{95.8} & \underline{86.6} & \underline{97.2} & \underline{37.0} & \underline{84.2} \\
& \cellcolor{lightbrown}\; + \textit{SALT}
& \cellcolor{lightbrown}\textbf{32.7} & \cellcolor{lightbrown}\textbf{62.8}
& \cellcolor{lightbrown}\textbf{27.8} & \cellcolor{lightbrown}\textbf{47.8}
& \cellcolor{lightbrown}\textbf{83.0} & \cellcolor{lightbrown}\textbf{96.5}
& \cellcolor{lightbrown}\textbf{88.0} & \cellcolor{lightbrown}\textbf{97.3}
& \cellcolor{lightbrown}\textbf{37.7} & \cellcolor{lightbrown}\textbf{85.0} \\
\cmidrule(lr){2-12}
& DAPO
& 29.0 & 61.3 & 26.0 & 45.0 & 82.2 & 95.0 & 84.8 & 96.8 & 36.0 & 80.9 \\
& \; + Entropy Bonus
& 29.2 & 61.5 & 26.2 & 45.4 & 82.4 & 95.4 & 85.0 & 97.0 & 36.3 & 81.3 \\
& \; + Clip Cov
& 29.8 & 61.2 & {26.8} & 44.9 & 82.7 & 95.2 & 85.3 & 96.7 & 36.7 & 81.0 \\
& \; + KL Cov
& \underline{30.0} & \underline{61.8} & \underline{27.0} & \underline{45.8} & \underline{82.9} & \underline{95.6} & \underline{85.6} & \underline{97.2} & \underline{37.0} & \underline{81.6} \\
& \cellcolor{lightbrown}\; + \textit{SALT}
& \cellcolor{lightbrown}\textbf{32.9} & \cellcolor{lightbrown}\textbf{63.8}
& \cellcolor{lightbrown}\textbf{28.0} & \cellcolor{lightbrown}\textbf{47.0}
& \cellcolor{lightbrown}\textbf{84.1} & \cellcolor{lightbrown}\textbf{96.8}
& \cellcolor{lightbrown}\textbf{88.0} & \cellcolor{lightbrown}\textbf{98.0}
& \cellcolor{lightbrown}\textbf{38.5} & \cellcolor{lightbrown}\textbf{82.9} \\\midrule
\rowcolor{lightgray}\multicolumn{12}{c}{\textbf{Deepseek-Distill-Qwen-7B}} \\
\midrule

\multirow{10}{*}{\textbf{MATH-T}}
& GRPO
& 60.7 & 77.3
& 42.0 & 71.3
& 89.9 & 97.7
& 94.9 & \underline{99.0}
& 46.8 & 83.8 \\
& \; + Entropy Bonus
& 60.2 & 77.1
& 41.6 & 71.2
& 90.0 & 97.7
& 94.8 & 98.9
& 46.5 & 83.7 \\
& \; + Clip Cov
& 61.1 & 77.9
& 42.3 & \underline{72.1}
& \underline{90.2} & \underline{97.8}
& 95.0 & \textbf{99.1}
& 47.1 & 84.3 \\
& \; + KL Cov
& \underline{61.5} & \underline{78.1}
& \underline{42.6} & 72.0
& 90.1 & {\underline{97.8}}
& \underline{95.1} & \underline{99.0}
& \underline{47.3} & \underline{84.4} \\
& \cellcolor{lightbrown}\;\; + \textit{SALT}
& \cellcolor{lightbrown}\textbf{64.7} & \cellcolor{lightbrown}\textbf{80.4}
& \cellcolor{lightbrown}\textbf{45.3} & \cellcolor{lightbrown}\textbf{74.5}
& \cellcolor{lightbrown}\textbf{91.9} & \cellcolor{lightbrown}\textbf{98.7}
& \cellcolor{lightbrown}\textbf{95.2} & \cellcolor{lightbrown}98.9
& \cellcolor{lightbrown}\textbf{51.2} & \cellcolor{lightbrown}\textbf{87.9}  \\
\cmidrule(lr){2-12}
& DAPO
& 54.0 & 79.3
& 42.0 & 62.7
& 90.1 & 97.5
& 94.2 & 99.0
& 48.9 & 83.7 \\
& \; + Entropy Bonus
& 54.3 & 79.6
& 42.2 & 63.4
& 90.1 & \underline{97.7}
& 94.2 & \textbf{99.2}
& 49.1 & 84.1 \\
& \; + Clip Cov
& 54.7 & 79.4
& 42.5 & 63.0
& \underline{90.3} & 97.6
& {\underline{94.3}} & 98.7
& 49.5 & 83.9 \\
& \; + KL Cov
& \underline{54.9} & \underline{79.8}
& \underline{42.7} & \underline{63.7}
& 90.2 & \underline{97.7}
& \underline{94.3} & {\underline{99.1}}
& \underline{49.6} & \underline{84.4} \\
& \cellcolor{lightbrown}\;\; + \textit{SALT}
& \cellcolor{lightbrown}\textbf{57.3} & \cellcolor{lightbrown}\textbf{82.2}
& \cellcolor{lightbrown}\textbf{44.8} & \cellcolor{lightbrown}\textbf{66.6}
& \cellcolor{lightbrown}\textbf{91.7} & \cellcolor{lightbrown}\textbf{99.0}
& \cellcolor{lightbrown}\textbf{95.2} & \cellcolor{lightbrown}\underline{99.1}
& \cellcolor{lightbrown}\textbf{52.2} & \cellcolor{lightbrown}\textbf{88.1}  \\
\midrule

\multirow{10}{*}{\textbf{DAPO-M}}
& GRPO
& 61.5 & 78.0
& 42.8 & 72.1
& 90.4 & 97.8
& 95.0 & \underline{99.1}
& 47.6 & 84.6 \\
& \; + Entropy Bonus
& 61.2 & 78.1
& 42.9 & 72.6
& 90.6 & 97.9
& 95.1 & 98.5
& \underline{48.8} & 85.1 \\
& \; + Clip Cov
& 61.9 & 78.6
& 43.3 & 72.5
& 90.7 & \underline{98.0}
& \underline{95.2} & 98.9
& 48.3 & 84.8 \\
& \; + KL Cov
& \underline{62.2} & \underline{78.8}
& \underline{43.4} & \underline{73.2}
& \underline{90.8} & \underline{98.0}
& \underline{95.2} & 98.8
& 48.5 & \underline{85.6} \\
& \cellcolor{lightbrown}\;\; + \textit{SALT}
& \cellcolor{lightbrown}\textbf{65.4} & \cellcolor{lightbrown}\textbf{81.2}
& \cellcolor{lightbrown}\textbf{46.5} & \cellcolor{lightbrown}\textbf{75.5}
& \cellcolor{lightbrown}\textbf{92.4} & \cellcolor{lightbrown}\textbf{99.3}
& \cellcolor{lightbrown}\textbf{96.4} & \cellcolor{lightbrown}\textbf{99.2}
& \cellcolor{lightbrown}\textbf{51.3} & \cellcolor{lightbrown}\textbf{89.2}  \\
\cmidrule(lr){2-12}
& DAPO
& 54.8 & 80.0
& 43.0 & 64.0
& 90.4 & 97.6
& 94.4 & \underline{99.1}
& 49.2 & 84.1 \\
& \; + Entropy Bonus
& 54.9 & 80.0
& 43.2 & 64.8
& 90.6 & 97.8
& 94.5 & \underline{99.1}
& 49.6 & 84.6 \\
& \; + Clip Cov
& \underline{55.2} & 79.7
& 43.8 & 64.3
& \underline{90.7} & 97.7
& \underline{94.6} & 98.7
& 49.9 & \underline{86.3} \\
& \; + KL Cov
& 55.0 & \underline{80.2}
& \underline{44.3} & \underline{65.4}
& \underline{90.7} & \underline{97.9}
& {\underline{94.6}} & \underline{99.1}
& \underline{50.4} & 84.8 \\
& \cellcolor{lightbrown}\;\; + \textit{SALT}
& \cellcolor{lightbrown}\textbf{56.6} & \cellcolor{lightbrown}\textbf{81.6}
& \cellcolor{lightbrown}\textbf{46.3} & \cellcolor{lightbrown}\textbf{68.3}
& \cellcolor{lightbrown}\textbf{92.1} & \cellcolor{lightbrown}\textbf{99.2}
& \cellcolor{lightbrown}\textbf{95.4} & \cellcolor{lightbrown}\textbf{99.3}
& \cellcolor{lightbrown}\textbf{53.1} & \cellcolor{lightbrown}\textbf{88.7} \\

\bottomrule
\end{tabular}
}
\label{tab:rq2_app}
\end{table}

\paragraph{Full Results.}
Table~\ref{tab:rq2} reports the averaged results for DeepSeek-Distill-Qwen-1.5B in the main paper.
Here we provide the complete results per-benchmark for both the 1.5B and 7B models in Table~\ref{tab:rq2_app}.

Two patterns are consistent across model scales and benchmarks.
First, \textbf{\textit{SALT} achieves the strongest overall performance}, with gains often more pronounced in Pass@8 than in Pass@1, indicating improved effectiveness under multi-sample evaluation.
Second, these gains are consistent with the mechanism emphasized throughout the paper: the update combines a stable main learning signal with an adaptive exploration component, which improves credit assignment and mitigates noisy or misaligned updates.
As a result, \textit{SALT} yields more reliable optimization under the same training budget, \textbf{translating into higher downstream accuracy rather than merely increasing or regulating response entropy}.

\subsection{Additional Ablations for RQ3}

\subsubsection{Component Necessity Ablations}
\label{app:ablation}

In Section~\ref{sec:ablation}, we conduct ablation studies on the individual components of our method \textit{SALT}, and present further details below (Table~\ref{tab:app_ablation}).
\paragraph{Ablation definitions.}
We ablate \textit{SALT} by selectively removing or modifying its key components while keeping the rest of the training pipeline identical.
Let $a$ denote the per-sample advantage vector, $P_k$ the estimated common-gradient subspace projector, and $P_k^\perp = I - P_k$ its orthogonal complement.
\textit{SALT} forms the modified advantages as:

    \noindent\textbullet \  \textbf{\textit{SALT} (full).} $a' = P_k a + \alpha_t\, P_k^\perp [a]_+$, where $\alpha_t$ is the cancellation-aware mixing weight computed from effective sample size.
    
    \noindent\textbullet \  \textbf{\textit{only-main}.} $a' = P_k a$. This variant keeps only the dominant spectral channel and removes the spectral-tail channel.
    
    \noindent\textbullet \  \textbf{\textit{only-exp}.} $a' = \alpha_t\, P_k^\perp [a]_+$. This variant removes the dominant spectral channel and updates purely along the spectral-tail directions.
    
    \noindent\textbullet \  \textbf{\textit{no-Positive}.} $a' = P_k a + \alpha_t\, P_k^\perp a$. This variant removes positive gating in the spectral-tail channel, allowing signed tail coefficients to be reweighted.
   
    \noindent\textbullet \  \textbf{\textit{no-proj}.} $a' = a + \alpha_t\, [a]_+$. This variant removes the subspace decomposition entirely (i.e., no projection), retaining only the mixing and gating mechanism.

   \noindent\textbullet \  \textbf{\textit{rand-proj}.} 
$a' = \widetilde{P}_k a + \alpha_t\, \widetilde{P}_k^\perp [a]_+$, where $\widetilde{P}_k = \widetilde{V}_k \widetilde{V}_k^\top$ is constructed from a random orthonormal basis $\widetilde{V}_k \in \mathbb{R}^{m \times k_t}$  and $\widetilde{P}_k^\perp = I - \widetilde{P}_k$.
This control removes data-driven subspace estimation while preserving the same two-channel structure.

To provide a more comprehensive assessment, we further evaluate alternative mixing schedules that control $\alpha_t$ using proxies other than $n_{eff}$ (e.g., PR-driven or entropy-driven rules), while keeping all other settings fixed. These controls test whether \textit{SALT}’s gains come from the cancellation-aware mixing, rather than from heuristic adaptivity.

\noindent\textbullet \  \textbf{\textit{fixed-$\alpha$}.}
$a' = P_k a + \alpha\, P_k^\perp [a]_+$ with a constant $\alpha \in [0,1]$ (shared across steps).
This control removes cancellation-aware adaptivity and tests whether a fixed mixing weight suffices.

\noindent\textbullet \  \textbf{\textit{PR-$\alpha$}.}
$a' = P_k a + \alpha_t\, P_k^\perp [a]_+$ with $\alpha_t = \mathrm{clip}\!\left(\frac{PR_{\mathrm{total}}}{m - 1},\, 0,\, 1\right)$.
This control drives mixing using effective dimensionality (PR) rather than cancellation ($n_{\mathrm{eff}}$).

\noindent\textbullet \  \textbf{\textit{Entropy-$\alpha$}.}
$a' = P_k a + \alpha_t\, P_k^\perp [a]_+$ with $\alpha_t = \mathrm{clip}\!\left(\frac{H_t - H_{\min}}{H_{\max} - H_{\min}},\, 0,\, 1\right)$, where $H_t$ is the policy entropy at step $t$ and $H_{\min}, H_{\max}$ are computed over the same training run.
This control uses exploration level as a proxy to set $\alpha_t$ instead of $n_{\mathrm{eff}}$.

\renewcommand{\arraystretch}{1.05}
    \definecolor{lightgray}{gray}{0.9} 
\colorlet{lightbrown}{blue!10}
\colorlet{lightbrown}{brown!10}
    
\begin{table}[t]
    \centering
    \caption{\textit{SALT} component ablations and alternative mixing schedules under GRPO (Deepseek-Distill-Qwen-1.5B; MATH-Train). Results are reported as Pass@1/Pass@8 on AIME24/25, GSM8K, MATH-500, and GPQA. \textbf{Bold} indicates the best performance. Overall, \textbf{the full \textit{SALT} design achieves the strongest average results, and removing key components consistently degrades performance.}}
    \small
          \resizebox{\textwidth}{!}{
\begin{tabular}{lcccccccccccc}
\toprule
\multirow{2}{*}{\textbf{Method}} & \multicolumn{2}{c}{\textbf{AIME24}} & \multicolumn{2}{c}{\textbf{AIME25}} & \multicolumn{2}{c}{\textbf{GSM8K}} & \multicolumn{2}{c}{\textbf{MATH-500}} & \multicolumn{2}{c}{\textbf{GPQA}} & \multicolumn{2}{c}{\textbf{Average}} \\
\cmidrule(lr){2-3}\cmidrule(lr){4-5}\cmidrule(lr){6-7}\cmidrule(lr){8-9}\cmidrule(lr){10-11}\cmidrule(lr){12-13}
    &    ACC       & Pass@8      & ACC       & Pass@8      & ACC     & Pass@8     & ACC      & Pass@8      & ACC       & Pass@8   & ACC       & Pass@8         \\ \midrule
Vanilla & 29.1 & 59.3 & 22.0 & 42.6 & 80.3 & 95.0 & 85.5 & 96.4 & 34.5 & 82.4 & 50.3 & 75.1 \\ \midrule
GRPO & 29.3 & 59.3 & \underline{25.6} & 43.3 & 80.7 & 94.8 & 85.6 & 96.7 & 35.0 & 82.2 & 51.2 & 75.3 \\
\;\; + \textit{only-main} & 30.1 & 60.5 & 25.3 & 44.6 & 81.9 & 95.7 & 86.0 & 97.1 & 36.2 & 83.2 & 51.9 & 76.2 \\
\;\; + \textit{only-exp}  & 29.8 & 60.0 & 25.0 & 44.0 & 81.5 & 95.4 & 85.9 & 97.0 & 35.8 & 82.9 & 51.6 & 75.9 \\
\;\; + \textit{no-Pos}   & 29.6 & 59.7 & 24.7 & 43.8 & 81.1 & 95.2 & 85.8 & 96.8 & 35.4 & 82.5 & 51.3 & 75.6 \\
\;\; + \textit{no-proj}   & 29.4 & 59.4 & 24.4 & 43.4 & 80.9 & 95.0 & 85.7 & 96.7 & 35.2 & 82.3 & 51.1 & 75.4 \\
\;\; + \textit{rand-proj} & 29.5 & 59.5 & 24.6 & 43.6 & 81.0 & 95.1 & 85.8 & 96.8 & 35.3 & 82.4 & 51.2 & 75.5 \\
\;\; + \textit{PR-$\alpha_t$}    & 30.3 & \underline{61.0} & 25.4 & \underline{45.0} & 82.0 & \underline{95.9} & \underline{86.1} & \underline{97.3} & \underline{36.5} & \underline{83.7} & \underline{52.1} & \underline{76.6} \\
\;\; + \textit{Ent-$\alpha_t$}   & 30.0 & 60.6 & 25.2 & 44.5 & 81.8 & 95.6 & 86.0 & 97.2 & 36.1 & 83.3 & 51.8 & 76.2 \\
\;\; + \textit{Fixed-$\alpha_t$} & \underline{30.7} & 59.9 & 24.9 & 44.1 & \underline{82.4} & 95.4 & 85.9 & 97.0 & 35.7 & 82.8 & 51.9 & 75.8 \\ 
 \cellcolor{lightbrown}\;\; + \textit{SALT}
& \cellcolor{lightbrown}\textbf{32.1} & \cellcolor{lightbrown}\textbf{62.9}
& \cellcolor{lightbrown}\textbf{27.1} & \cellcolor{lightbrown}\textbf{47.2}
& \cellcolor{lightbrown}\textbf{83.4} & \cellcolor{lightbrown}\textbf{96.5}
& \cellcolor{lightbrown}\textbf{87.2} & \cellcolor{lightbrown}\textbf{98.4}
& \cellcolor{lightbrown}\textbf{38.4} & \cellcolor{lightbrown}\textbf{85.5} 
& \cellcolor{lightbrown}\textbf{53.6} & \cellcolor{lightbrown}\textbf{78.1}\\

     \bottomrule
    \end{tabular}
}
    \label{tab:app_ablation}
\end{table}

All variants use the same $k_t$ and $\alpha_t$ computation when applicable, and differ only in how $a$ is transformed into $a'$.

\paragraph{Benchmark performance.}
While the main text focuses on the PR--$n_{\mathrm{eff}}$ geometry in Figure~\ref{fig:rq3} (a), we also report the \emph{\textit{final benchmark performance}} of \textit{SALT} and all ablations under identical training budgets.
Table~\ref{tab:app_ablation} summarizes results across all benchmarks.
Overall, the full \textit{SALT} achieves the best performance and the most consistent gains.
In contrast, removing either channel (\textit{only-main}/\textit{only-exp}), breaking the learned subspace structure (\textit{no-proj}/\textit{rand-proj}), or replacing cancellation-aware mixing with alternative schedules (\textit{fixed-$\alpha$}/\textit{PR-$\alpha$}/\textit{Entropy-$\alpha$}) leads to degraded performance, corroborating that \textbf{both the dual-channel design and cancellation-aware mixing are necessary}.

\begin{figure}[t]
    \centering
    \includegraphics[width=1.0\linewidth]{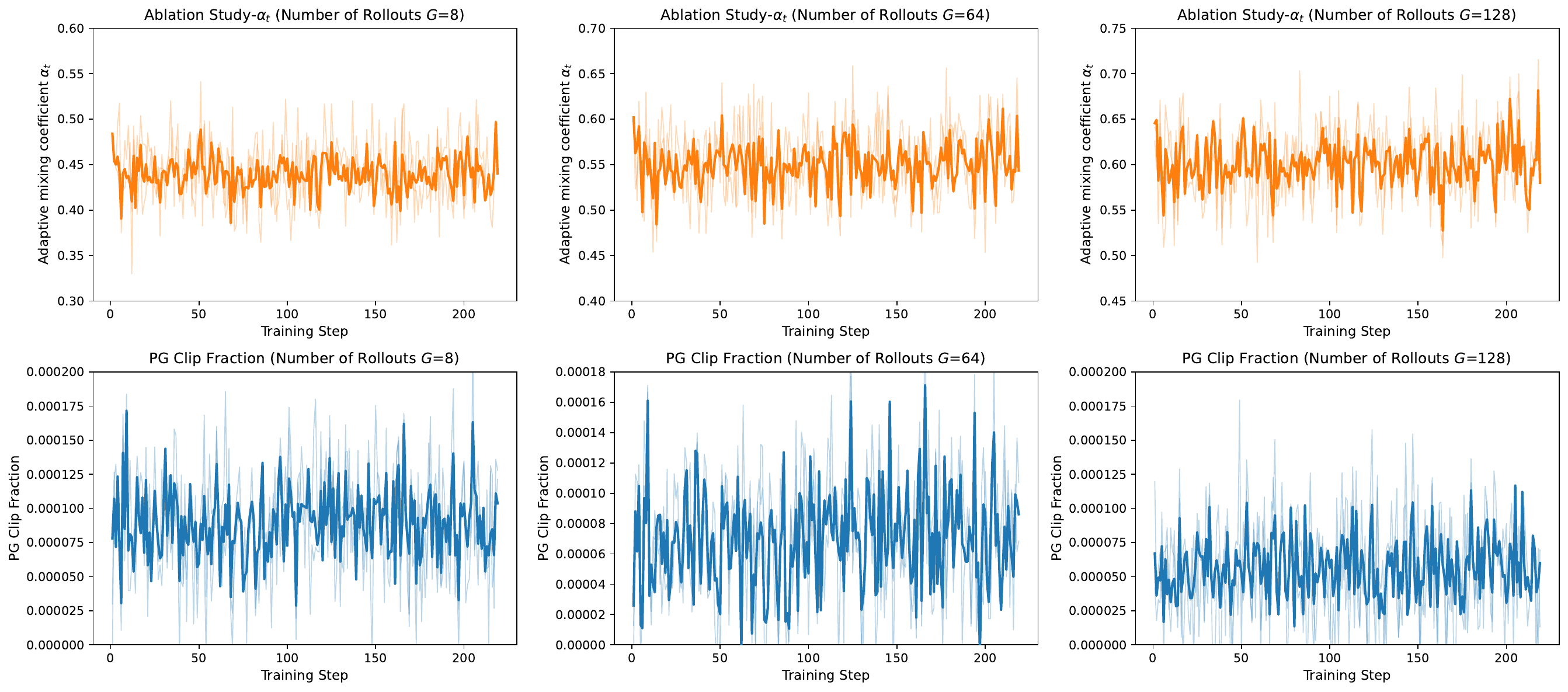}
\caption{
Adaptive mixing coefficient and policy-gradient clipping fraction under different rollout group sizes.
As the rollout group size increases from $G=8$ to $G=128$, the adaptive mixing coefficient $\alpha_t$ increases, indicating that \textbf{\textit{SALT} assigns more weight to the spectral-tail channel when larger rollout groups induce stronger redundancy and cancellation}.
Meanwhile, the policy-gradient clipping fraction remains at the $10^{-4}$ level and stays below $2\times10^{-4}$ throughout training, suggesting that the \textbf{observed gains are unlikely to be driven by PPO clipping artifacts}.
}
\vspace{-1em}
\label{fig:alpha_clip}
    \label{fig:visual}
\end{figure}

\paragraph{Visualization of $\alpha_t$ dynamics and clipping diagnostics.}
We further visualize the temporal evolution of the adaptive mixing coefficient $\alpha_t$ under different rollout group sizes.
As shown in Figure~\ref{fig:visual}, larger rollout groups consistently lead to higher $\alpha_t$, indicating that \textit{SALT} adaptively places more weight on the residual/exploration channel when redundancy and signed cancellation become stronger.
This behavior is consistent with the design of the cancellation-aware mixing rule.

We also report the policy-gradient clipping fraction in the same figure to rule out clipping-induced artifacts.
Across all rollout group sizes, the clipping fraction remains at the $10^{-4}$ level and stays below $2\times 10^{-4}$ throughout training.
Therefore, the effect of \textit{SALT} is unlikely to be driven by PPO clipping nonlinearities; instead, the diagnostic supports our interpretation that \textit{SALT} improves update aggregation through geometry-aware residual reweighting.

\subsubsection{Matched-Coefficient Subspace Intervention}
\label{app:matched_intervention}
The component ablations above verify the necessity of individual \textit{SALT} components. However, modifying a component may also change the marginal distribution, scale, and positive mass of the reweighted coefficients, making it difficult to fully isolate whether the improvement comes from geometry-aware subspace alignment or from coefficient reshaping alone. To address this, we introduce a matched-coefficient subspace intervention that \textbf{preserves the coefficient distribution of \textit{SALT} while breaking its alignment with the true mini-batch gradient geometry}.

\paragraph{Protocol.}
For each checkpoint and mini-batch, we fix the prompts, sampled rollouts, rewards, original group-normalized advantages, and the adaptive mixing coefficient. We first compute the standard \textit{SALT} coefficients $a'_{\mathrm{SALT}}$ and record their sorted values. We then construct matched null variants by reassigning the same sorted coefficient values according to alternative sample orderings. This keeps the marginal distribution, norm, and positive mass of the coefficients identical to \textit{SALT}, while disrupting their sample-wise alignment with the data-dependent gradient geometry.

\paragraph{Intervention variants.}
We consider two matched null variants. \textbf{Permuted-\textit{SALT}} randomly permutes the assignment between $a'_{\mathrm{SALT}}$ and samples. \textbf{Matched-RandSubspace} first samples a random orthogonal subspace with the same dimension as the \textit{SALT} dominant subspace, uses the induced residual scores to rank samples, and then assigns the sorted values of $a'_{\mathrm{SALT}}$ according to this random-subspace ranking. In both variants, the coefficient distribution is exactly matched to \textit{SALT}; only the alignment between coefficients and the true mini-batch gradient geometry is broken.

\paragraph{Results and interpretation.}
As shown in Table~\ref{tab:matched_subspace_intervention}, both matched null variants degrade the effective update geometry compared with \textit{SALT}. Since these variants preserve the coefficient distribution, this degradation cannot be explained by changes in coefficient scale, sparsity, or positive mass. Instead, the result indicates that the gains of \textit{SALT} rely on assigning reweighted coefficients to geometry-aligned samples, rather than merely changing the distribution of coefficient values. Together with the clipping diagnostic in Figure~\ref{fig:visual}, where the policy-gradient clipping fraction remains below $2\times10^{-4}$, these results further suggest that the observed gains are unlikely to arise from PPO clipping artifacts.
\renewcommand{\arraystretch}{1.05}
    \definecolor{lightgray}{gray}{0.9} 
\colorlet{lightbrown}{blue!10}
\colorlet{lightbrown}{brown!10}
    
\begin{table}[t]
\centering
\caption{
\textbf{Matched-coefficient subspace intervention} on DeepSeek-Distill-Qwen-7B / MATH-TRAIN.
\textbf{Coef. Dist.} indicates whether the marginal distribution of reweighted coefficients is matched to the corresponding standard \textit{SALT} variant under the same backbone;
\textbf{Geo. Align.} indicates whether the coefficient-to-sample assignment is aligned with the true data-dependent mini-batch gradient geometry.
Matched null variants preserve the coefficient distribution of \textit{SALT} while breaking this geometry alignment.
Their degradation in effective update geometry indicates that \textbf{\textit{SALT}'s gains come from geometry-aligned coefficient assignment rather than coefficient redistribution alone.}
}
\label{tab:matched_subspace_intervention}
\resizebox{0.98\linewidth}{!}{
\begin{tabular}{llcccccc}
\toprule
\textbf{Backbone}
& \textbf{Method} 
& \textbf{Coef. Dist.} 
& \textbf{Geo. Align.} 
& $D_a \downarrow$ 
& $D_g \downarrow$ 
& $n_{\mathrm{eff}} \uparrow$ 
& $U_{\perp}$ \textbf{Gain} $\uparrow$ \\
\midrule
\textbf{GRPO} 
& Backbone 
& -- 
& -- 
& $0.000$ 
& $0.000$ 
& $0.31 \pm 0.05$ 
& $1.00\times$ \\

\textbf{GRPO} 
& RandSubspace-\textit{SALT} 
& No 
& No 
& $0.22 \pm 0.04$ 
& $0.18 \pm 0.03$ 
& $0.50 \pm 0.05$ 
& $1.13\times$ \\

\textbf{GRPO} 
& Permuted-\textit{SALT} 
& Yes 
& No 
& $0.24 \pm 0.04$ 
& $0.19 \pm 0.03$ 
& $0.43 \pm 0.05$ 
& $1.08\times$ \\

\textbf{GRPO} 
& Matched-RandSubspace 
& Yes 
& No 
& $0.24 \pm 0.04$ 
& $0.18 \pm 0.03$ 
& $0.55 \pm 0.04$ 
& $1.18\times$ \\

\rowcolor{lightbrown}
\textbf{GRPO} 
& + \textit{SALT} 
& Yes 
& Yes 
& $0.24 \pm 0.04$ 
& $0.16 \pm 0.03$ 
& $0.80 \pm 0.04$ 
& $1.47\times$ \\

\midrule
\textbf{DAPO} 
& Backbone 
& -- 
& -- 
& $0.000$ 
& $0.000$ 
& $0.82 \pm 0.03$ 
& $1.00\times$ \\

\textbf{DAPO} 
& RandSubspace-\textit{SALT} 
& No 
& No 
& $0.16 \pm 0.03$ 
& $0.13 \pm 0.02$ 
& $0.86 \pm 0.04$ 
& $1.07\times$ \\

\textbf{DAPO} 
& Permuted-\textit{SALT} 
& Yes 
& No 
& $0.17 \pm 0.03$ 
& $0.13 \pm 0.02$ 
& $0.84 \pm 0.04$ 
& $1.04\times$ \\

\textbf{DAPO} 
& Matched-RandSubspace 
& Yes 
& No 
& $0.17 \pm 0.03$ 
& $0.12 \pm 0.02$ 
& $0.89 \pm 0.03$ 
& $1.10\times$ \\

\rowcolor{lightbrown}
\textbf{DAPO} 
& + \textit{SALT} 
& Yes 
& Yes 
& $0.17 \pm 0.03$ 
& $0.11 \pm 0.02$ 
& $1.00 \pm 0.03$ 
& $1.24\times$ \\
\bottomrule
\end{tabular}
}
\end{table}

\subsection{Additional evidence for RQ4}
\label{app:rq4}
In Section~\ref{sec:rollout}, we show that larger rollout groups $G$ amplify both update diversity and gradient cancellation, and that \textit{SALT} can better translate large-$G$ rollouts into effective learning.

\begin{figure}[ht]
    \centering
    \includegraphics[width=1.0\linewidth]{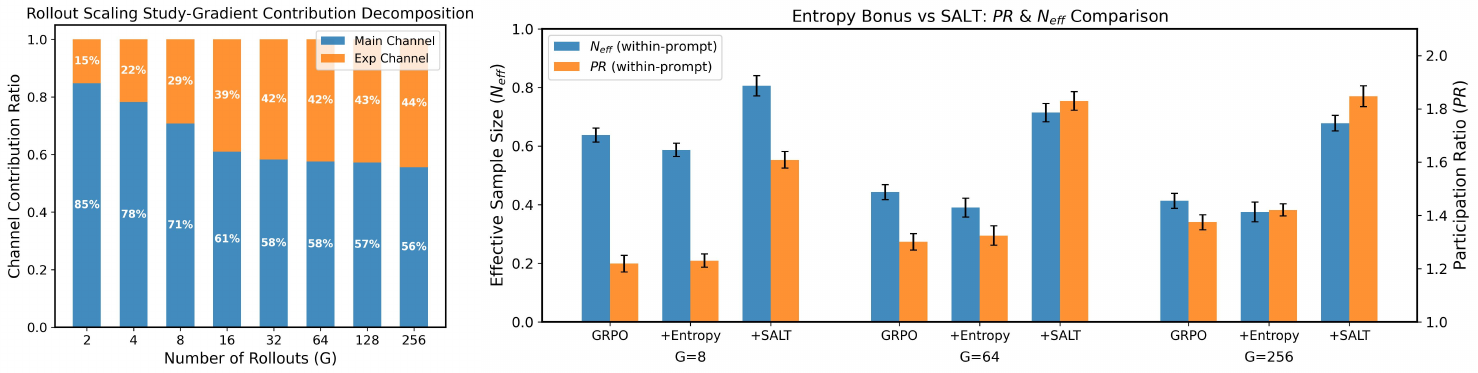}
    \vspace{-1em}

\caption{\textbf{Rollout scaling diagnostics:}
(a) Decomposition of \textit{SALT}’s gradient/update contributions into the main and exploration channels as the rollout group size $G$ increases; the exploration share (and thus mixing) grows with $G$, \textbf{indicating on-demand exploration under stronger cancellation.}
(b) Comparison with an entropy-only baseline at $G\!=\!8$, $G\!=\!64$ and $G\!=\!256$: \textbf{entropy increases randomness (higher $PR$) but does not recover $N_{\mathrm{eff}}$} as effectively as \textit{SALT}, leading to weaker or less reliable gains.}

    \label{fig:app_rq4}
\end{figure}

Here we provide additional analyses to connect these geometric trends to the underlying mechanism and to rule out simpler alternatives.
First, we examine how \textit{SALT} allocates updates between its main and exploration channels (Figure~\ref{fig:rq3}).
Consistently, the exploration channel’s contribution and the mixing coefficient $\alpha_t$ increase with $G$, indicating on-demand exploration under severe cancellation.
Second, we compare against an entropy-only variant: while an entropy bonus improves randomness, it does not recover $N_{\mathrm{eff}}$ as effectively, leading to weaker or less reliable gains.
Together, \textbf{these results suggest that increasing entropy alone is insufficient; cancellation-aware mixing is key.}


\subsection{Bias--Cancellation Trade-off Diagnostics}
\label{app:bias_cancellation}

Section~\ref{sec:surrogate} shows that \textit{SALT} intentionally departs from the original GRPO-style estimator by replacing the group-normalized coefficients $a$ with geometry-reweighted coefficients $a'$. This appendix quantifies whether this biased surrogate is both \emph{controlled} and \emph{useful}: controlled if the update deviation from GRPO remains moderate, and useful if the deviation increases the effective residual update and improves downstream accuracy.

\paragraph{Metrics.}
Let $S=[s_1,\ldots,s_m]$ denote the score-gradient feature matrix and $K=\frac{1}{m}S^\top S$ be the mini-batch Gram matrix. For any coefficient vector $c$, define the induced update
\begin{equation}
\hat g(c)=\frac{1}{m}Sc,
\qquad
\|\hat g(c)\|_2
=
\sqrt{\frac{1}{m}c^\top Kc}.
\end{equation}
We measure the coefficient distortion and update deviation introduced by \textit{SALT} as
\begin{equation}
D_a
=
\frac{\|a'-a\|_2}{\|a\|_2+\epsilon},
\qquad
D_g
=
\frac{\|\hat g(a')-\hat g(a)\|_2}{\|\hat g(a)\|_2+\epsilon}.
\end{equation}
To measure whether the deviation is useful, we report the cancellation-adjusted effective update size
\begin{equation}
n_{\mathrm{eff}}(c)
=
\frac{c^\top Kc}{\sum_{i=1}^m c_i^2 K_{ii}+\epsilon},
\end{equation}
and the spectral-tail update strength
\begin{equation}
U_{\perp}(c)
=
\left\|
\frac{1}{m}S P_k^\perp c
\right\|_2
=
\sqrt{
\frac{1}{m}
(P_k^\perp c)^\top K(P_k^\perp c)
}.
\end{equation}
All quantities are computed from the same Gram statistics already used by \textit{SALT}, without materializing full-model gradients.

\paragraph{Results.}
Table~\ref{tab:bias_cancellation} reports averages over training checkpoints and seeds. The original GRPO/DAPO updates have zero coefficient distortion by definition. \textit{SALT} introduces a nonzero but bounded deviation from the backbone update, while consistently increasing $n_{\mathrm{eff}}$, spectral-tail update strength, and downstream accuracy. In the measured settings, this indicates that reducing signed cancellation outweighs the coefficient distortion; it is not a claim of true-policy-gradient unbiasedness or a universal improvement guarantee.
\renewcommand{\arraystretch}{1.05}
\definecolor{lightgray}{gray}{0.9} 
\colorlet{lightbrown}{blue!10}
\colorlet{lightbrown}{brown!10}


\begin{table}[t]
\centering
\caption{
Bias--cancellation trade-off diagnostics averaged over the last 20\% training checkpoints and 5 seeds on DeepSeek-Distill-Qwen-7B / MATH-TRAIN.
$D_a$ measures coefficient distortion, $D_g$ measures update deviation from the corresponding backbone, $n_{\mathrm{eff}}$ measures cancellation-adjusted effective update size, and $U_\perp$ measures spectral-tail update strength.
\textbf{\textit{SALT} introduces a moderate deviation while increasing effective update strength and downstream accuracy}.
}

\label{tab:bias_cancellation}
\resizebox{0.90\linewidth}{!}{
\begin{tabular}{llccccc}
\toprule
\textbf{Backbone} & \textbf{Method} 
& $D_a \downarrow$ 
& $D_g \downarrow$ 
& $n_{\mathrm{eff}} \uparrow$ 
& $U_\perp$ \textbf{Gain} $\uparrow$ 
& \textbf{Avg. ACC} $\uparrow$ \\
\midrule
\textbf{GRPO} & Backbone 
& $0.000$ 
& $0.000$ 
& $0.31 \pm 0.05$ 
& $1.00\times$ 
& $66.9$ \\
\rowcolor{lightbrown}\textbf{GRPO} & + \textit{SALT} 
& $0.24 \pm 0.04$ 
& $0.16 \pm 0.03$ 
& $0.80 \pm 0.04$ 
& $1.47\times$ 
& $69.7$ \\
\midrule
\textbf{DAPO} & Backbone 
& $0.000$ 
& $0.000$ 
& $0.82 \pm 0.03$ 
& $1.00\times$ 
& $65.8$ \\
\rowcolor{lightbrown}\textbf{DAPO} & + \textit{SALT} 
& $0.17 \pm 0.03$ 
& $0.11 \pm 0.02$ 
& $1.00 \pm 0.03$ 
& $1.24\times$ 
& $68.2$ \\
\bottomrule
\end{tabular}
}
\end{table}

\section{Computational Cost}
\label{app:time_cost}
In this section, we will report the the complexity analysis  and the wall-clock time cost.

\subsection{Computational Complexity}
\label{app:complexity}

Let $B$ be the number of prompts in a mini-batch, $G$ the number of rollouts per prompt, and
$m \triangleq BG$ the total number of sampled responses in the batch.
Our dual-channel reweighting introduces a \emph{geometry module} on top of the base GRPO/DAPO~\citep{grpo,yu2025dapoopensourcellmreinforcement} update.
Its cost depends on $m$ (batch rollout count), rather than the full parameter size of the language model.

\renewcommand{\arraystretch}{1.05}
\definecolor{lightgray}{gray}{0.9} 
\colorlet{lightbrown}{blue!10}
\colorlet{lightbrown}{brown!10}


\begin{wraptable}{r}{0.48\textwidth}
\vspace{-0.7em}
    \centering
\caption{End-to-end wall-clock training time (Hour:Min) for GRPO/DAPO and variants, with and without \textit{SALT}, on MATH-TRAIN/DAPO-MATH. All runs use the same training configuration; times include the full training loop.}
\vspace{1em}

    \small
        \resizebox{0.47\textwidth}{!}{
\begin{tabular}{lcc}
\toprule
\multirow{2}{*}{\textbf{Method}} & \multicolumn{2}{c}{\textbf{Time Cost (Hour:Min)}} \\
 \cmidrule(lr){2-3}                    & \textbf{MATH-TRAIN}  & \textbf{DAPO-MATH}  \\ \midrule
\rowcolor{lightgray}\multicolumn{3}{l}{\textbf{Deepseek-Distill-Qwen-1.5B}} \\
\midrule
GRPO                    & 07:21       & 15:41      \\
\rowcolor{lightbrown} \; \; + \textit{SALT}                   & 07:58       & 17:08      \\
\; \; + Entropy Bonus          & 08:48       & 17:14      \\
\; \; + Clip Cov               & 07:53       & 16:03      \\
\; \; + KL Cov                 & 07:34       & 16:01      \\
DAPO                    & 07:17       & 16:01      \\
\rowcolor{lightbrown} \; \; + \textit{SALT}                   & 08:05       & 17:58      \\
\midrule
\rowcolor{lightgray}\multicolumn{3}{l}{\textbf{Deepseek-Distill-Qwen-7B}} \\
\midrule
GRPO                    & 24:07       & 61:26      \\
\rowcolor{lightbrown}\; \; + \textit{SALT}                   & 26:22       & 66:08      \\
\; \; + Entropy Bonus          & 25:52       & 17:14      \\
\; \; + Clip Cov               & 25:31       & 63:30      \\
\; \; + KL Cov                 & 25:32       & 63:35      \\
DAPO                    & 24:31       & 62:23      \\
\rowcolor{lightbrown} \; \; + \textit{SALT}                   & 25:58       & 66:04     \\
\bottomrule
    \end{tabular}
    }
    \label{tab:time}
    \vspace{-3em}
\end{wraptable}

\paragraph{(1) Per-sample gradient feature extraction.}
For each response $(b,i)$, we compute a last-layer gradient embedding proxy $g_{b,i}$ based on the
output projection gradients (Eq.~\eqref{eq:back_logit}).
This step reuses quantities already produced in backprop (token-level $\nabla_{z}\ell$ and hidden states $h$),
and aggregates over tokens to obtain a response-level feature.
Denoting the average response length by $T$ and the proxy feature dimension by $d_g$.
In practice this overhead is typically small compared to the full forward+backward pass through the transformer stack~\citep{badge,tracin,trak}.

\paragraph{(2) Forming the Gram matrix.}
Stack all features into $G_{\text{all}}\in\mathbb{R}^{d_g\times m}$ and form
\begin{equation}
K_{\text{total}}=\frac{1}{m}G_{\text{all}}^\top G_{\text{all}}\in\mathbb{R}^{m\times m}.
\end{equation}
The memory to store $K_{\text{total}}$ is $O(m^2)$.
The time to compute $G_{\text{all}}^\top G_{\text{all}}$ is $O(d_g m^2)$; since $d_g$ is fixed for a given proxy choice,
the dominant scaling with $m$ is $O(m^2)$ up to the constant factor $d_g$.

\paragraph{(3) Computing the dominant subspace (top-$k_t$ eigenvectors).}
We only need the leading $k_t$ eigenvectors to build the projector $P_k$.
Using partial eigensolvers, the time complexity scales as
\begin{equation}
O(m^2 k_t),
\end{equation}
and the additional memory beyond storing $K_{\text{total}}$ is typically $O(m k_t)$ to hold the eigenvectors
$V_k\in\mathbb{R}^{m\times k_t}$.
The dominant dimension is set adaptively by
$k_t=\mathrm{clip}(\lfloor \mathrm{PR}_{\text{total}}\rceil,1,BG)$,
so $k_t\ll m$ in our settings.

\paragraph{(4) Advantage projection and reweighting.}
Given $V_k$, the main/exploration channels are
\begin{equation}
a_{\text{main}}=P_k a,\qquad a_{\text{exp}}=P_k^\perp [a]_+,
\end{equation}
where $P_k=V_kV_k^\top$ and $P_k^\perp=I-P_k$.
We never need to materialize $P_k\in\mathbb{R}^{m\times m}$.
Compute $P_k a$ as $V_k(V_k^\top a)$, which costs $O(mk_t)$; similarly for
$P_k^\perp [a]_+$.
Thus, projection/reweighting is linear in $m$ up to $k_t$ and is negligible compared with the Gram/eigensolver steps.

\paragraph{Overall overhead.}
The geometry module is dominated by (i) storing the Gram matrix ($O(m^2)$ memory) and
(ii) extracting the top-$k_t$ subspace ($O(m^2k_t)$ time).
Crucially, this overhead depends on $m=BG$ and is independent of the total parameter count of the backbone model,
since we use last-layer gradient embeddings as a scalable proxy.
\subsection{Time Cost}
Table~\ref{tab:time} reports the end-to-end wall-clock training time for each method under the same hardware and training setup.
All runs use identical hyperparameters and differ only in the reweighting/regularization module.
\textit{SALT} adds a lightweight geometry module on top of GRPO/DAPO; consistent with the analysis in Appendix~\ref{app:complexity}, its overhead is small relative to the full forward/backward pass and mainly comes from forming the $m\times m$ Gram statistics.
\section{Statistical significance}

For each setting, we run \textit{SALT} and the corresponding baseline (GRPO/DAPO) with $N=5$ matched seeds $\{2025,2026,2027,2028,2029\}$ under identical training and evaluation configurations.
For each seed $s$, we define the paired improvement as $d_s \triangleq \text{score}^{\text{SALT}}_s-\text{score}^{\text{base}}_s$, and report the mean improvement $\Delta \triangleq \frac{1}{N}\sum_{s=1}^N d_s$.

We compute 95\% confidence intervals via paired bootstrap: we resample $\{d_s\}_{s=1}^N$ with replacement for $B$ times and take the 2.5/97.5 percentiles of the bootstrap means as $(\Delta_{\mathrm{low}},\Delta_{\mathrm{high}})$.
We report results in the compact form $\Delta^{+u}_{-l}$ where $u=\Delta_{\mathrm{high}}-\Delta$ and $l=\Delta-\Delta_{\mathrm{low}}$ in Table~\ref{tab:sign}.
\renewcommand{\arraystretch}{1.05}
\definecolor{lightgray}{gray}{0.9}
\colorlet{lightbrown}{brown!10}
\newcommand{\deltaci}[3]{\ensuremath{#1^{#2}_{#3}}}

\begin{table}[t]
    \centering
    \caption{Paired improvements of \textit{SALT} over the corresponding baseline, reported as $\Delta^{+u}_{-l}$ (mean paired gain with 95\% bootstrap CI over $N=5$ matched seeds).}
    \small
        \resizebox{\textwidth}{!}{

\begin{tabular}{clc@{\hspace{4pt}}c@{\hspace{5pt}} c@{\hspace{4pt}}c@{\hspace{5pt}}c@{\hspace{4pt}}c@{\hspace{5pt}}c@{\hspace{4pt}}c@{\hspace{5pt}}c@{\hspace{4pt}}c}
\toprule
\multirow{2}{*}{\textbf{Training}} &  & \multicolumn{2}{c}{\textbf{AIME24}} & \multicolumn{2}{c}{\textbf{AIME25}} & \multicolumn{2}{c}{\textbf{GSM8K}} & \multicolumn{2}{c}{\textbf{MATH-500}} & \multicolumn{2}{c}{\textbf{GPQA}} \\
\cmidrule(lr){3-4}\cmidrule(lr){5-6}\cmidrule(lr){7-8}\cmidrule(lr){9-10}\cmidrule(lr){11-12}
  &      &    ACC       & Pass@8      & ACC       & Pass@8      & ACC     & Pass@8     & ACC      & Pass@8      & ACC       & Pass@8          \\ \midrule

\rowcolor{lightgray} \multicolumn{12}{c}{\textbf{Deepseek-Distill-Qwen-1.5B}}     \\ \midrule
\multirow{2}{*}{\textbf{MATH-T}}
& GRPO &
\deltaci{+2.8}{+0.7}{-0.6} &
\deltaci{+3.6}{+1.0}{-0.8} &
\deltaci{+1.5}{+0.4}{-0.3} &
\deltaci{+3.9}{+0.9}{-0.8} &
\deltaci{+2.7}{+0.8}{-0.6} &
\deltaci{+1.7}{+0.6}{-0.5} &
\deltaci{+1.6}{+0.6}{-0.4} &
\deltaci{+1.7}{+0.5}{-0.5} &
\deltaci{+3.4}{+0.9}{-0.8} &
\deltaci{+3.3}{+0.8}{-0.7} \\
\cmidrule(lr){2-12}
& DAPO
& \deltaci{-0.5}{+0.4}{-0.3} &
\deltaci{+0.7}{+0.6}{-0.5} &
\deltaci{+1.7}{+0.5}{-0.4} &
\deltaci{+2.0}{+0.7}{-0.6} &
\deltaci{+1.5}{+0.6}{-0.5} &
\deltaci{+1.0}{+0.3}{-0.2} &
\deltaci{+1.0}{+0.4}{-0.3} &
\deltaci{+0.0}{+0.9}{-0.8} &
\deltaci{+2.6}{+0.9}{-0.7} &
\deltaci{+1.8}{+0.6}{-0.5} \\
\midrule

\multirow{2}{*}{\textbf{DAPO-M}}
& GRPO
& \deltaci{+3.6}{+1.1}{-0.9} &
\deltaci{+3.8}{+1.0}{-0.8} &
\deltaci{+4.2}{+1.1}{-0.9} &
\deltaci{+3.6}{+1.0}{-0.8} &
\deltaci{+0.8}{+0.6}{-0.8} &
\deltaci{+1.2}{+0.5}{-0.4} &
\deltaci{+2.1}{+0.8}{-0.6} &
\deltaci{+0.5}{+0.6}{-0.5} &
\deltaci{+1.5}{+0.5}{-0.4} &
\deltaci{+1.3}{+0.7}{-0.8} \\
\cmidrule(lr){2-12}
& DAPO
& \deltaci{+3.9}{+1.1}{-0.9} &
\deltaci{+2.5}{+0.9}{-0.8} &
\deltaci{+2.0}{+0.7}{-0.6} &
\deltaci{+2.0}{+0.8}{-0.7} &
\deltaci{+1.9}{+0.7}{-0.6} &
\deltaci{+1.8}{+0.7}{-0.6} &
\deltaci{+3.2}{+0.9}{-0.8} &
\deltaci{+1.2}{+0.5}{-0.5} &
\deltaci{+2.5}{+0.9}{-0.7} &
\deltaci{+2.0}{+0.7}{-0.8} \\
\midrule

\rowcolor{lightgray} \multicolumn{12}{c}{\textbf{Deepseek-Distill-Qwen-7B}}     \\ \midrule

\multirow{2}{*}{\textbf{MATH-T}}
& GRPO
&
\deltaci{+4.0}{+1.0}{-0.8} &
\deltaci{+3.1}{+0.9}{-0.7} &
\deltaci{+3.3}{+0.9}{-0.8} &
\deltaci{+3.2}{+1.0}{-0.8} &
\deltaci{+2.0}{+0.7}{-0.6} &
\deltaci{+1.0}{+0.3}{-0.2} &
\deltaci{+0.3}{+0.4}{-0.3} &
\deltaci{-0.1}{+0.3}{-0.2} &
\deltaci{+4.4}{+1.1}{-0.9} &
\deltaci{+4.1}{+1.1}{-0.9} \\
\cmidrule(lr){2-12}
& DAPO
& \deltaci{+3.3}{+1.0}{-0.8} &
\deltaci{+2.9}{+0.9}{-0.7} &
\deltaci{+2.8}{+1.0}{-0.8} &
\deltaci{+3.9}{+1.1}{-0.9} &
\deltaci{+1.6}{+0.7}{-0.6} &
\deltaci{+1.5}{+0.6}{-0.5} &
\deltaci{+1.0}{+0.3}{-0.2} &
\deltaci{+0.1}{+0.3}{-0.2} &
\deltaci{+3.3}{+1.0}{-0.8} &
\deltaci{+4.4}{+1.2}{-0.9} \\
\midrule

\multirow{2}{*}{\textbf{DAPO-M}}
& GRPO
& \deltaci{+3.9}{+1.0}{-0.9} &
\deltaci{+3.2}{+0.9}{-0.7} &
\deltaci{+3.7}{+1.0}{-0.8} &
\deltaci{+3.4}{+1.0}{-0.8} &
\deltaci{+2.0}{+0.7}{-0.6} &
\deltaci{+1.5}{+0.6}{-0.5} &
\deltaci{+1.4}{+0.4}{-0.3} &
\deltaci{+0.1}{+0.4}{-0.3} &
\deltaci{+3.7}{+1.1}{-0.9} &
\deltaci{+4.6}{+1.2}{-0.9} \\
\cmidrule(lr){2-12}
& DAPO
& \deltaci{+1.8}{+0.6}{-0.6} &
\deltaci{+1.6}{+0.6}{-0.7} &
\deltaci{+3.3}{+1.0}{-0.9} &
\deltaci{+4.3}{+1.2}{-0.9} &
\deltaci{+1.7}{+0.7}{-0.6} &
\deltaci{+1.6}{+0.6}{-0.5} &
\deltaci{+1.0}{+0.3}{-0.2} &
\deltaci{+0.2}{+0.4}{-0.3} &
\deltaci{+3.9}{+1.1}{-0.9} &
\deltaci{+4.6}{+1.2}{-0.9} \\
\bottomrule
\end{tabular}
}
\label{tab:sign}
\end{table}

\section{Formulations of GRPO-style Variants}
\label{appendix:formulations}
In Section~\ref{sec:pre}, we briefly introduced the formulations of RL and GRPO. In this section, we will proceed from PPO to GRPO, and then present the formal formulations for several optimized variants~\citep{shao2024deepseekmathpushinglimitsmathematical}.

\paragraph{Proximal Policy Optimization (PPO)}
Proximal Policy Optimization (PPO)~\citep{ppo} is a widely adopted actor--critic reinforcement learning algorithm, and it is frequently employed in the RL-based fine-tuning of large language models (LLMs). Specifically, PPO updates the LLM by maximizing the surrogate objective defined as follows:
\begin{equation}
\begin{aligned}
\mathcal{J}_{\text{PPO}}(\theta)
&= \mathbb{E}_{q \sim P(Q),\, o \sim \pi_{\theta_{\text{old}}}(O \mid q)}
\frac{1}{|o|}\sum_{t=1}^{|o|}
 \min\!\Bigg(
\frac{\pi_{\theta}\!\left(o_t \mid q, o_{<t}\right)}{\pi_{\theta_{\text{old}}}\!\left(o_t \mid q, o_{<t}\right)} \hat{A_t},\;\\ 
& \mathrm{clip}\!\left(
\frac{\pi_{\theta}\!\left(o_t \mid q, o_{<t}\right)}{\pi_{\theta_{\text{old}}} \!\left(o_t \mid q, o_{<t}\right)},
1-\epsilon,\,1+\epsilon
\right) \hat{A_t}
\Bigg)
\Bigg].
\label{eq:ppo}
\end{aligned}
\end{equation}
In this objective, \(q \sim P(Q)\) denotes a query sampled from the data distribution,
\(o=(o_1,\ldots,o_{|o|})\) is a response generated by the behavior policy
\(\pi_{\theta_{\text{old}}}\) and $\hat{A_t}$ is an estimator of the advantage at time step $t$. Given the value function $V$ and the  reward function $R$, $\hat{A_t}$ is computed using the Generalized Advantage Estimation (GAE)~\citep{schulman2016gae}.

\paragraph{Group Relative Policy Optimization (GRPO)}
In PPO, the value function is often realized as an additional model comparable in size to the policy, leading to substantial memory and computation overhead.
In the LLM setting, rewards are typically provided only at the sequence level, which further complicates learning token-level value estimates.
To address this, \citet{shao2024deepseekmathpushinglimitsmathematical} propose \emph{Group Relative Policy Optimization} (GRPO), which eliminates the value function and instead uses the average reward of multiple samples for the same question as a baseline.
Concretely, for each question \(q\), GRPO draws \(\{o_1,\ldots,o_G\}\sim \pi_{\theta_{\text{old}}}(\cdot\mid q)\) and maximizes the following objective (Eq.~\eqref{equation:grpo}):
\begin{equation}
\label{equation:grpo_app}
    \begin{aligned}
\mathcal{J}_{\mathrm{GRPO}}(\theta)
&=
\mathbb{E}_{q\sim P(Q),\,\{o_i\}_{i=1}^{G}\sim \pi_{\theta_{\mathrm{old}}}(O\mid q)}
\Bigg[
\frac{1}{G}\sum_{i=1}^{G}\frac{1}{|o_i|}\sum_{t=1}^{|o_i|}
\Bigg(
\min\Bigg(
\frac{\pi_{\theta}(o_{i,t}\mid q,o_{i,<t})}{\pi_{\theta_{\mathrm{old}}}(o_{i,t}\mid q,o_{i,<t})}\,\hat{A}_{i,t},
\\
&
\operatorname{clip}\Bigg(
\frac{\pi_{\theta}(o_{i,t}\mid q,o_{i,<t})}{\pi_{\theta_{\mathrm{old}}}(o_{i,t}\mid q,o_{i,<t})},
1-\epsilon,\;1+\epsilon
\Bigg)\hat{A}_{i,t}
\Bigg)
-\beta\,\mathbb{D}_{\mathrm{KL}}\!\left[\pi_{\theta}\,\|\,\pi_{\mathrm{ref}}\right]
\Bigg)
\Bigg]. \end{aligned}\end{equation}
Rewards within the group are normalized to obtain a response-level advantage (Eq.~\eqref{eq:advantage})
\begin{equation}
    \hat{A}_{i}= \frac{r_i-\text{mean}(r_1,r_2,\cdots,r_G)}{\text{std}(r_1,r_2,\cdots,r_G)}.
\end{equation}
And different from the KL penalty term used in Eq.~\eqref{eq:ppo}, the KL divergence is computed with the following unbiased estimator
\begin{equation}
\begin{aligned}
\mathbb{D}_{\mathrm{KL}}\! \left[\pi_{\theta}\,\|\,\pi_{\mathrm{ref}}\right]
&=
\frac{\pi_{\mathrm{ref}}\!\left(o_{i,t}\mid q,\,o_{i,<t}\right)}{\pi_{\theta}\!\left(o_{i,t}\mid q,\,o_{i,<t}\right)}
-\log\!\left(
\frac{\pi_{\mathrm{ref}}\!\left(o_{i,t}\mid q,\,o_{i,<t}\right)}{\pi_{\theta}\!\left(o_{i,t}\mid q,\,o_{i,<t}\right)}
\right)
-1 .
\end{aligned}
\end{equation}
which is guaranteed to be positive~\citep{schulman_kl_approx}.

\paragraph{Decoupled Clip and Dynamic Sampling Policy Optimization (DAPO)} \citet{yu2025dapoopensourcellmreinforcement}
propose the Decouple Clip and Dynamic Sampling Policy Optimization (DAPO) algorithm. DAPO samples a group of outputs $\{o_i\}_{i=1}^G$, and optimizes the policy via the following objective:
\begin{equation}
\begin{aligned}
\mathcal{J}_{\mathrm{DAPO}}(\theta)
&=
\mathbb{E}_{(q,a)\sim\mathcal{D},\ \{o_i\}_{i=1}^{G}\sim\pi_{\theta_{\mathrm{old}}}(\cdot\mid q)}
\Bigg[
\frac{1}{\sum_{i=1}^{G}|o_i|}
\sum_{i=1}^{G} \sum_{t=1}^{|o_i|}
\min\!\Bigg(
\frac{\pi_{\theta}(o_{i,t}\mid q,o_{i,<t})}{\pi_{\theta_{\mathrm{old}}}(o_{i,t}\mid q,o_{i,<t})}\,\hat{A}_{i,t},\; \\ 
&\mathrm{clip} \! \ \Bigg(\frac{\pi_{\theta}(o_{i,t}\mid q,o_{i,<t})}{\pi_{\theta_{\mathrm{old}}}(o_{i,t}\mid q,o_{i,<t})},\,1-\epsilon_{\mathrm{low}},\, 1+\epsilon_{\mathrm{high}}\Bigg)\,\hat{A}_{i,t}
\Bigg)
\Bigg],
\end{aligned}
\end{equation}
\[\text{s.t.}\ \ 0<\bigl|\{\,o_i\mid \mathrm{is\_equivalent}(a,o_i)\,\}\bigr|<G .
\]
where the reward $r_{i,t}(\theta)$ and advantage $\hat{A}_{i,t}$ are the same with Eq.~\eqref{equation:grpo} in GRPO.

\paragraph{Entropy Bonus.}
We consider a standard entropy regularization baseline by augmenting the GRPO objective with a token-level policy entropy term.
For a state $s_{i,t} \!=\! (q, o_{i,<t})$, define~\citep{shen2025entropycontrolllmrlalgorithms,mnih2016asynchronousmethodsdeepreinforcement}
\begin{equation}
\mathcal{H}_\theta(s_{i,t})
\triangleq
- \sum_{a \in \mathcal{V}} \pi_\theta(a \mid s_{i,t}) \log \pi_\theta(a \mid s_{i,t}),
\end{equation}
where $\mathcal{V}$ is the vocabulary.
The entropy-bonus variant optimizes with entropy coefficient $\lambda_{\mathrm{ENT}} \ge 0$.

\begin{equation}
\begin{aligned}
\mathcal{J}_{\mathrm{ENT}}(\theta)
&=
\mathcal{J}_{\mathrm{GRPO}}(\theta)
+ 
\lambda_{\mathrm{ENT}}\; \cdot \;
\mathbb{E}_{q,\{o_i\}_{i=1}^G}
\Bigg[ 
\frac{1}{G}\sum_{i=1}^G \frac{1}{|o_i|}\sum_{t=1}^{|o_i|}
\mathcal{H}_\theta(q,o_{i,<t})
\Bigg],
\end{aligned}
\end{equation}




\newpage

\section{Algorithm}
\label{app_algo}

Algorithm~\ref{alg:salt_full} provides the full per-step implementation of the \textit{SALT} advantage reweighting procedure described in Section~\ref{sec:method}.

\begin{algorithm*}[!h]
\caption{\textbf{\textit{SALT}}: Dual-Channel Advantage Reweighting for GRPO-style RLVR}
\label{alg:salt_full}
\begin{algorithmic}[1]
\REQUIRE Policy $\pi_\theta$; prompts $\{q_b\}_{b=1}^B$; rollouts per prompt $G$; small numerical constant $\epsilon$.
\ENSURE Updated parameters $\theta$.

\STATE \textbf{Rollout and reward.}
For each prompt $q_b$, sample $G$ responses $\{o_{b,i}\}_{i=1}^G \sim \pi_\theta(\cdot|q_b)$ and compute rewards $\{r_{b,i}\}_{i=1}^G$.

\STATE \textbf{Group-normalized advantages.}
Compute GRPO-style normalized advantages $\hat A_{b,i}$ within each prompt group and stack them into
\[
a \leftarrow \mathrm{vec}(\hat A_{b,i}) \in \mathbb{R}^{m},
\qquad m=BG.
\]

\STATE \textbf{Output-projection gradient features.}
For each response $(b,i)$, compute the response-level output-projection gradient
\[
\nabla_{W_{\mathrm{out}}}\ell_{b,i}
=
\frac{1}{|o_{b,i}|}
\sum_{t=1}^{|o_{b,i}|}
\big(\nabla_{z_{b,i,t}}\ell_{b,i,t}\big)h_{b,i,t}^{\top},
\]
and flatten it into a feature vector $g_{b,i}$.

\STATE \textbf{Batch Gram geometry.}
Stack all features into
\[
\mathbf{F} \leftarrow [g_{b,i}]_{(b,i)} \in \mathbb{R}^{d\times m},
\qquad
K \leftarrow \frac{1}{m}\mathbf{F}^{\top}\mathbf{F}\in\mathbb{R}^{m\times m}.
\]

\STATE \textbf{Dominant subspace.}
Compute $K=V\Lambda V^\top$ and the regularized participation score
\[
\mathrm{PR}(K)
\leftarrow
\frac{(\sum_{j=1}^{m}\lambda_j)^2}
{\sum_{j=1}^{m}\lambda_j^2}.
\]
Set
\[
k_t \leftarrow \mathrm{clip}\big(\lfloor \mathrm{PR}(K)\rceil,\,1,\,m\big),
\qquad
V_k \leftarrow [v_1,\ldots,v_{k_t}].
\]

\STATE \textbf{Projectors.}
Construct
\[
P_k \leftarrow V_kV_k^\top,
\qquad
P_k^\perp \leftarrow I-P_k.
\]

\STATE \textbf{Dual-channel advantage decomposition.}
\[
a_{\mathrm{main}} \leftarrow P_k a,
\qquad
a_{\mathrm{exp}} \leftarrow P_k^\perp\,[a]_+.
\]

\STATE \textbf{Signed-cancellation estimate.}
Construct signed per-sample update contributions
\[
u_{b,i} \propto \hat A_{b,i}\nabla_\theta \log \pi_\theta(o_{b,i}|q_b),
\]
and compute
\[
n_{\mathrm{eff}}
\leftarrow
\frac{
\left\|\sum_{(b,i)}u_{b,i}\right\|_2^2
}{
\sum_{(b,i)}\|u_{b,i}\|_2^2+\epsilon
}.
\]

\STATE \textbf{Adaptive mixing and policy update.}
Set
\[
\alpha_t \leftarrow \max(0,1-n_{\mathrm{eff}}),
\qquad
a' \leftarrow a_{\mathrm{main}}+\alpha_t a_{\mathrm{exp}}
= P_k a+\alpha_t P_k^\perp[a]_+.
\]
Update $\theta$ with the base GRPO-style objective, replacing each $\hat A_{b,i}$ by $a'_{b,i}$.

\end{algorithmic}
\end{algorithm*}



\newpage
\section*{NeurIPS Paper Checklist}

\begin{enumerate}

\item {\bf Claims}
    \item[] Question: Do the main claims made in the abstract and introduction accurately reflect the paper's contributions and scope?
    \item[] Answer: \answerYes{} 
    \item[] Justification: The paper clearly states its main claims and supports them with theory, diagnostics, and experiments.

\item {\bf Limitations}
    \item[] Question: Does the paper discuss the limitations of the work performed by the authors?
    \item[] Answer: \answerYes{} 
    \item[] Justification: The limitation of the work is described in the conclusion section.

\item {\bf Theory assumptions and proofs}
    \item[] Question: For each theoretical result, does the paper provide the full set of assumptions and a complete (and correct) proof?
    \item[] Answer: \answerYes{} 
    \item[] Justification: Theoretical assumptions are stated, and the main residual-gradient bound is proved in the appendix.

    \item {\bf Experimental result reproducibility}
    \item[] Question: Does the paper fully disclose all the information needed to reproduce the main experimental results of the paper to the extent that it affects the main claims and/or conclusions of the paper (regardless of whether the code and data are provided or not)?
    \item[] Answer: \answerYes{} 
    \item[] Justification: The paper provides datasets, models, training settings, hyperparameters, and the SALT algorithm.

\item {\bf Open access to data and code}
    \item[] Question: Does the paper provide open access to the data and code, with sufficient instructions to faithfully reproduce the main experimental results, as described in supplemental material?
    \item[] Answer: \answerYes{}{} 
    \item[] Justification: The work uses public datasets and provides public code, training scripts, and reproduction instructions.

\item {\bf Experimental setting/details}
    \item[] Question: Does the paper specify all the training and test details (e.g., data splits, hyperparameters, how they were chosen, type of optimizer) necessary to understand the results?
    \item[] Answer: \answerYes{} 
    \item[] Justification: The experimental setup, evaluation metrics, sampling settings, and implementation details are described.

\item {\bf Experiment statistical significance}
    \item[] Question: Does the paper report error bars suitably and correctly defined or other appropriate information about the statistical significance of the experiments?
    \item[] Answer: \answerYes{} 
    \item[] Justification: Results are averaged over five runs and include paired confidence intervals.

\item {\bf Experiments compute resources}
    \item[] Question: For each experiment, does the paper provide sufficient information on the computer resources (type of compute workers, memory, time of execution) needed to reproduce the experiments?
    \item[] Answer: \answerYes{} 
    \item[] Justification: The appendix reports hardware, wall-clock time, and SALT overhead.
    
\item {\bf Code of ethics}
    \item[] Question: Does the research conducted in the paper conform, in every respect, with the NeurIPS Code of Ethics \url{https://neurips.cc/public/EthicsGuidelines}?
    \item[] Answer: \answerYes{} 
    \item[] Justification: The work uses public assets and automated evaluation without human-subject data.

\item {\bf Broader impacts}
    \item[] Question: Does the paper discuss both potential positive societal impacts and negative societal impacts of the work performed?
    \item[] Answer: \answerNo{} 
    \item[] Justification:
This is a methodological RLVR optimization paper with no new datasets, human-subject experiments, deployment scenarios, or application-specific systems; any societal impact is indirect and tied to downstream model use.
    
    
\item {\bf Safeguards}
    \item[] Question: Does the paper describe safeguards that have been put in place for responsible release of data or models that have a high risk for misuse (e.g., pre-trained language models, image generators, or scraped datasets)?
    \item[] Answer: \answerNA{} 
    \item[] Justification: No new high-risk dataset or deployed model system is released.

\item {\bf Licenses for existing assets}
    \item[] Question: Are the creators or original owners of assets (e.g., code, data, models), used in the paper, properly credited and are the license and terms of use explicitly mentioned and properly respected?
    \item[] Answer: \answerNo{} 
    \item[] Justification: The draft does not yet explicitly list licenses for all used datasets, models, and tools.

\item {\bf New assets}
    \item[] Question: Are new assets introduced in the paper well documented and is the documentation provided alongside the assets?
    \item[] Answer: \answerYes{} 
    \item[] Justification: The submission provides public code for SALT and the experimental pipeline.

\item {\bf Crowdsourcing and research with human subjects}
    \item[] Question: For crowdsourcing experiments and research with human subjects, does the paper include the full text of instructions given to participants and screenshots, if applicable, as well as details about compensation (if any)? 
    \item[] Answer: \answerNA{} 
    \item[] Justification: The work does not involve crowdsourcing or human-subject research.

\item {\bf Institutional review board (IRB) approvals or equivalent for research with human subjects}
    \item[] Question: Does the paper describe potential risks incurred by study participants, whether such risks were disclosed to the subjects, and whether Institutional Review Board (IRB) approvals (or an equivalent approval/review based on the requirements of your country or institution) were obtained?
    \item[] Answer: \answerNA{} 
    \item[] Justification: No human-subject research is conducted.

\item {\bf Declaration of LLM usage}
    \item[] Question: Does the paper describe the usage of LLMs if it is an important, original, or non-standard component of the core methods in this research? Note that if the LLM is used only for writing, editing, or formatting purposes and does \emph{not} impact the core methodology, scientific rigor, or originality of the research, declaration is not required.
    \item[] Answer: \answerYes{} 
    \item[] Justification: LLMs are the main experimental subject and are described in the paper.

\end{enumerate}

\end{document}